\documentclass[journal]{IEEEtran}
\usepackage{amsmath,amsthm}
\usepackage{amsmath}
\usepackage{amssymb}
\usepackage{amsfonts}
\usepackage{graphicx}
\graphicspath{{graphics/}}
\usepackage{tikz,tikz-3dplot}
\usepackage{xcolor}
\usepackage{bm}
\usepackage{dsfont}
\usepackage{stfloats}
\usepackage[flushleft]{threeparttable}
\usepackage[numbers,sort&compress]{natbib}

\usepackage[switch]{lineno}

\usepackage[colorlinks]{hyperref}
\hypersetup{colorlinks,breaklinks,linkcolor=blue,urlcolor=blue,anchorcolor=blue,citecolor=blue}
\usepackage{booktabs}
\usepackage{tabularx}
\usepackage{multirow}
\usepackage{lscape}
\usepackage{epstopdf}

\usepackage{booktabs}

\usepackage{enumitem}

\usepackage{booktabs}
\usepackage{multirow}
\usepackage{array}

\usepackage{caption}
\usepackage{graphicx}
\usepackage{float}

\usepackage{subcaption}

\usepackage{microtype}
\usepackage{lineno}
\usepackage{placeins}
\usepackage{xcolor}

\usepackage{algorithm}
\usepackage{algorithmic}

\usepackage{xcolor}
\usepackage{pifont}       
\usepackage{bbding}       

\newcommand{\A}{\mathcal{A}}
\newcommand{\B}{\mathcal{B}}
\newcommand{\C}{\mathcal{C}}

\newcommand{\G}{\mathcal{G}}

\newcommand{\K}{\mathcal{K}}

\newcommand{\M}{\mathcal{M}}
\newcommand{\N}{\mathcal{N}}

\newcommand{\T}{\mathcal{T}}
\newcommand{\U}{\mathcal{U}}
\newcommand{\V}{\mathcal{V}}

\newcommand{\X}{\mathcal{X}}
\newcommand{\Y}{\mathcal{Y}}
\newcommand{\Z}{\mathcal{Z}}

\newcommand{\tensor}[1]{\mathcal{#1}}

\newtheorem{definition}{Definition}[section]
\newtheorem{theorem}{Theorem}[section]
\newtheorem{proposition}{Proposition}[section]
\newtheorem{lemma}{Lemma}[section]

\newcommand{\norm}[1]{\lVert#1\rVert}

\newcommand{\normlarge}[1]{\left\lVert#1\right\rVert}

\newcommand{\0}{\mathcal{O}}

\newcommand{\Pomega}{\mathcal{P}_{\Omega}}

\newcommand{\Pomegac}{\mathcal{P}_{{\Omega}^{\perp}}}

\newcommand{\Pomegat}{\mathcal{P}_{{\Omega}_{\mathcal{T}}}}

\usepackage{amsmath}
\renewcommand{\algorithmicrequire}{\textbf{Input:}}  
\renewcommand{\algorithmicensure}{\textbf{Output:}} 

\usepackage{threeparttable}
\usepackage{dcolumn}

\newcolumntype{d}[1]{D{.}{.}{#1}}

\ifCLASSINFOpdf
\else
\fi

\usepackage{algorithmic}
\usepackage{algorithm}
\renewcommand{\algorithmicrequire}{\textbf{Input:}}  
\renewcommand{\algorithmicensure}{\textbf{Output:}} 

\usepackage{xcolor}
\definecolor{darkblue}{rgb}{0.0,0.5,0.5}

\begin{document}
%

\title{Guaranteed Multidimensional Time Series Prediction via Deterministic Tensor Completion Theory}

\author{\IEEEauthorblockN{Hao Shu, Jicheng Li, Yu Jin,}
}

\author{
Hao~Shu,~Jicheng~Li,~Yu~Jin,~Hailin Wang 
\thanks{The work was supported by National Natural Science Foundation of China (12171384). (\textit{Corresponding author: Jicheng Li})} \thanks{H. Shu, J. Li, Y. Jin  and H. Wang  are with the School of Mathematics and Statistics, Xi'an Jiaotong University, Xi'an 710049, Shanxi, China (email: haoshu812@gmail.com,
 jcli@mail.xjtu.edu.cn, jinyu1491240@163.com, wanghailin97@163.com).}}

\maketitle
\begin{abstract}
In recent years, the prediction of multidimensional time series data has become increasingly important due to its wide-ranging applications. Tensor-based prediction methods have gained attention for their ability to preserve the inherent structure of such data. However, existing approaches, such as tensor autoregression and tensor decomposition,  often have consistently failed to provide clear assertions regarding the number of samples that can be exactly predicted. While matrix-based methods using nuclear norms address this limitation, their reliance on matrices limits accuracy and increases computational costs when handling multidimensional data. To overcome these challenges, we reformulate multidimensional time series prediction as a deterministic tensor completion problem and propose a novel theoretical framework. Specifically, we develop a deterministic tensor completion theory and introduce the \textit{Temporal Convolutional Tensor Nuclear Norm} (TCTNN) model.
By convolving the multidimensional time series along the temporal dimension and applying the tensor nuclear norm, our approach identifies the maximum forecast horizon for exact predictions. Additionally, TCTNN achieves superior performance in prediction accuracy and computational efficiency compared to existing methods across diverse real-world datasets, including climate temperature, network flow, and traffic ride data. Our implementation is publicly available at \url{https://github.com/HaoShu2000/TCTNN}.

\end{abstract}



%
\IEEEpeerreviewmaketitle

\section{Introduction}\label{sec:Introduction}
\IEEEPARstart{M}ultidimensional time series, characterized by data with two or more dimensions at each time point, are often generated in a wide range of real-world scenarios, such as  regional climate data \cite{chen2021bayesian}, network flow  data \cite{ling2021t}, video data \cite{liu2012tensor}, international relations data \cite{schein2016bayesian}, and social network data \cite{chen2022factor}. Accurate forecasting of these temporal datasets is critical for numerous applications \cite{liu2020fast} \cite{meynard2021efficient} \cite{isufi2019forecasting}, such as optimizing route planning and traffic signal management in intelligent transportation systems \cite{li2018brief} \cite{shu2024low}. 
Affected by the acquisition equipment and other practical factors, the length of the observation data may be limited.  For instance, high-throughput biological data, such as gene expression datasets, typically contain numerous features but only a few time samples \cite{ma2018randomly}. Therefore, developing reliable few-sample multidimensional time series prediction methods has been a long-standing basic research challenge.

Traditional time series models, such as vector autoregression \cite{hyndman2018forecasting} and its variants \cite{athanasopoulos2008varma} \cite{wang2022high} \cite{basu2015regularized}, are not well-suited for handling multidimensional data. 
Recent studies represent multidimensional time series as tensors \cite{chen2022factor}, where the first dimension corresponds to time and the remaining dimensions capture features or locations.
For example, temperature data across different geographical locations can be structured as a third-order tensor (time × longitude × latitude), inherently preserving interactions between dimensions \cite{chen2023discovering}. This tensor representation has enabled the development of methods such as tensor factor models 
\cite{chen2022factor} \cite{lam2012factor}
and tensor autoregressive models \cite{chen2021autoregressive} 
\cite{wang2024high} \cite{shi2020block} \cite{jing2018high}.

\begin{figure*}[t]
\centering
\includegraphics[width=0.98\linewidth]{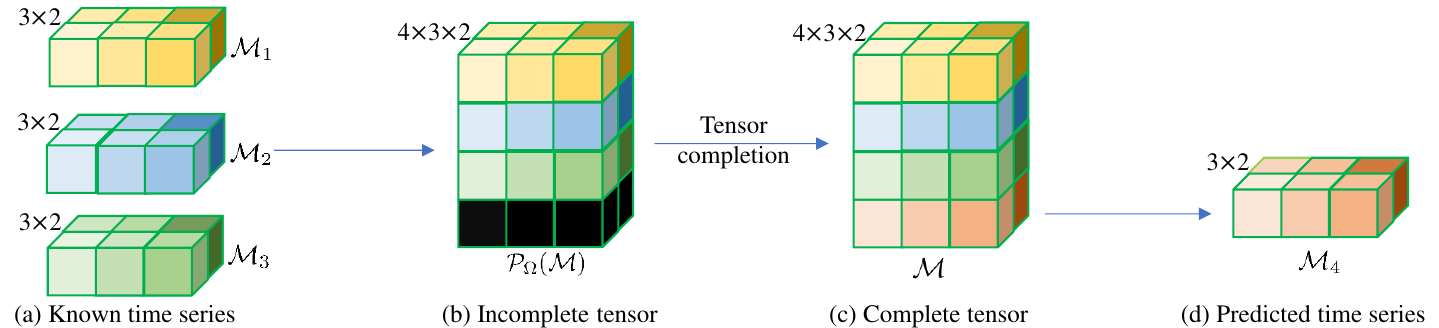}
\vspace{-0.2cm}
\caption{Illustration of tensor completion for multidimensional time series prediction, where black cubes represent unsampled entries, and other cubes represent sampled entries.}\label{fig:prediction---completion}
\vspace{-0.6cm}
\end{figure*}

Another promising approach is to frame multidimensional time series prediction as a tensor completion task, where the regions of interest are treated as missing values. As illustrated in Fig.\ref{fig:prediction---completion}, this process involves three steps: concatenating observed and predicted time series into an incomplete tensor, completing the tensor, and extracting the forecasted values. Within this framework, the low-rankness of multidimensional time series has been extensively utilized, leading to decomposition-based forecasting methods  \cite{chen2021bayesian} 
 \cite{takeuchi2017autoregressive} \cite{yu2016temporal}. Nevertheless, existing tensor factorization, autoregressive, and decomposition models fail to  offer
a definitive conclusion about the number of samples that can be predicted exactly.

Low-rank tensor completion methods based on nuclear norm minimization, supported by recovery theory, offer a promising alternative. Some researchers have converted multidimensional time series into matrices and applied deterministic matrix nuclear norm minimization theories for predictive modeling \cite{liu2022time} \cite{liu2022recovery} \cite{gillard2018structured} \cite{butcher2017simple}. While these approaches provide theoretical support, they suffer from two key limitations: (1) the conversion from tensors to matrices loses critical structural information, reducing prediction accuracy; and (2) the computation of large-scale matrix nuclear norms imposes significant computational and memory costs. Despite advances in deterministic matrix recovery theory, an equivalent theory for tensor nuclear norm minimization remains absent, necessitating such matrix-based approaches \cite{kiraly2012combinatorial,singer2010uniqueness,harris2021deterministic,tsakiris2023low,foucart2020weighted,shapiro2018matrix,chatterjee2020deterministic,chen2015completing,liu2019matrix,lee2013matrix}.

To address this gap, we propose a deterministic tensor completion theory based on tensor nuclear norm minimization for non-random missing data. Unlike direct application of tensor nuclear norm minimization to forecasting, our method introduces a Temporal Convolution Tensor Nuclear Norm (TCTNN) model, which transforms multidimensional time series via temporal convolution and applies tensor nuclear norm minimization to the transformed data. This approach establishes an exact prediction theory while achieving higher accuracy and lower computational costs compared to existing methods. Our main contribution can be summarized as follows:

\begin{itemize}
\item  Within the t-SVD framework, a tensor completion theory suitable for any deterministic missing data has been established by introducing the minimum slice sampling rate as a new sampling metric.
\item The TCTNN model is proposed by encoding the low-rankness of the temporal convolution tensor, and its corresponding exact prediction theory is also provided. Furthermore, the impact of the smoothness and periodicity of the time dimension on the temporal convolution low-rankness is analyzed.
\item The proposed TCTNN model is efficiently solved based on ADMM framework and well tested on several real-world multidimensional time series datasets.  The results indicate that the proposed method achieves better performance in prediction accuracy and computation time over many state-of-the-art algorithms, especially in the few-shot scenario.

\end{itemize}

The remainder of this paper is organized as follows. In Section~\ref{sec:rel}, we review a series of related works. 
In Section~\ref{sec:Deterministic Tensor Completion Theory},   A deterministic tensor completion theory is establish.
Section~\ref{sec:Method and Model} gives the proposed TCTNN method and its corresponding prediction theory, and Section~\ref{sec:Algorithm} introduces the optimization algorithm. In Section~\ref{sec:experiments}, we conduct extensive experiments on some multidimensional time series datasets. Finally, the conclusion is given in Section~\ref{sec:conclusion}. It should be noted that all proof details are given in supplementary materials.

\section{Related Works}\label{sec:rel}

\subsection{Deterministic completion theory}

Recovery theory for matrix/tensor completion in random cases has been extensively studied, yielding significant advancements in many fields \cite{candes2009exact} \cite{candes2010power}  \cite{wang2023guaranteed} \cite{peng2022exact}. In contrast, deterministic observation patterns have garnered considerably less attention.
However, the pattern of missing entries depends on the specific problem context in practice, which can lead to highly non-random occurrences. A case in point is the presence of missing entries in images, which is often attributed to occlusions or shadows present in the scene. 
To investigate the recovery theory for matrix/tensor completion under non-random missing conditions, previous researchers have conducted extensive exploration using various tools \cite{kiraly2012combinatorial,singer2010uniqueness,harris2021deterministic,tsakiris2023low,foucart2020weighted,shapiro2018matrix,chatterjee2020deterministic,chen2015completing,liu2019matrix,lee2013matrix}.
For instance, Singer et al. \cite{singer2010uniqueness} apply the basic ideas and tools of rigidity theory to determine the uniqueness of low-rank matrix completion. 
Based on  Plücker coordinates, Tsakiris \cite{tsakiris2023low} provides three families of patterns of arbitrary rank and arbitrary matrix size that allow unique or finite number of completions.
Besides, Liu  et al. \cite{liu2019matrix} propose isomeric condition and relative well-conditionedness to ensure that any matrix can be recovered from sampling of matrix terms. Despite these advancements, most of the current theoretical work focuses on deterministic matrix completion, while the tensor case is not explored enough. Therefore, the theoretical framework for deterministic tensor completion still requires significant enhancement.

\subsection{Time series prediction via matrix/tensor completion}

Matrix and tensor completion methods have garnered significant attention for time series forecasting \cite{yu2016temporal} 
\cite{gillard2018structured} \cite{tan2016short}. These approaches generally fall into two categories: decomposition-based methods and nuclear norm-based methods.
Matrix/tensor decomposition, widely used for collaborative filtering, provide a natural solution for handling missing data in time series forecasting \cite{tan2016short}  \cite{dunlavy2011temporal} \cite{chen2019bayesian}. These models approximate the incomplete time series using bilinear or multilinear decompositions with predefined rank parameters.
To capture temporal dynamics, recent studies have introduced temporal constraint functions, such as autoregressive regularization terms, to regularize the temporal factor matrix \cite{yang2021real} \cite{ yu2016temporal} \cite{ chen2019bayesian}.
Based on this, Takeuchi et al. \cite{takeuchi2017autoregressive}  model spatiotemporal tensor data by introducing spatial autoregressive regularizer,  providing additional predictive power in the spatial dimension.

As opposed to decomposition-based methods, nuclear norm-based approaches do not require predefined rank parameters \cite{candes2009exact}  \cite{wang2023guaranteed} \cite{peng2022exact}. However, directly applying nuclear norms to time series often yields suboptimal predictions \cite{liu2020fast}. A practical enhancement involves applying structural transformations to time series data to create structured matrices, enabling the application of nuclear norms for forecasting. For example, Gillard et al. 
 \cite{gillard2018structured} and Butcher et al. \cite{butcher2017simple} used the Hankel transform to convert univariate time series into Hankel matrices and applied nuclear norm minimization for prediction. Similarly, Liu et al. \cite{liu2022time} \cite{liu2022recovery}  utilized convolution matrices for encoding nuclear norms, providing insights into the predictive domain that can be exactly forecasted.
Although these methods offer predictive insights, they involve converting time series into large matrices, resulting in the loss of inherent tensor structure and imposing high computational costs.
To address these limitations, we propose the  \textit{Temporal Convolution Tensor Nulclear Norm} (TCTNN) model, which preserves the tensor structure, offers theoretical guarantees for predictions, and achieves superior performance in terms of both accuracy and computational efficiency.

\section{Deterministic Tensor Completion} \label{sec:Deterministic Tensor Completion Theory}
In contrast to the typical scenario of tensor completion with random missing values, the prediction problem in the context of multidimensional time series prediction corresponds to a deterministic tensor completion problem. To address this, we first introduce a new theory of deterministic tensor completion in this section. 
\subsection{T-SVD Framework}
We begin by providing a brief overview of the key concepts within the \textit{tensor Singular Value Decomposition} (t-SVD) framework  \cite{kilmer2011factorization} \cite{kilmer2021tensor}  \cite{lu2019tensor} \cite{martin2013order} \cite{qin2022low}. This includes the tensor-tensor product, tensor nuclear norm, and the conditions for tensor incoherence. Following this, we propose a novel deterministic sampling metric designed to establish an exact recovery theory for deterministic tensor completion.

For consistency, we use lowercase, boldface lowercase, capital, and Euler script letters to represent scalars, vectors, matrices, and tensors, respectively, e.g., $x \in \mathbb{R}$, $\boldsymbol{x}\in \mathbb{R}^{m}$, $X\in \mathbb{R}^{m_1 \times m_2}$, and $ \tensor{X}\in \mathbb{R}^{m_1\times m_2\times\cdots\times m_d}$.
For an order-$d$ tensor $\mathcal{X}$ of dimensions $m_1\times m_2\times\cdots\times m_d$, its elements, slices, and sub-tensors are represented as $\X_{(...)}$, such as $\X_{(i_1,i_2,\cdots,i_d)}$ represents the $(i_1,i_2,\cdots,i_d)$-th element and $\mathcal{X}_{(:,:,i_3,\cdots,i_d)}$      represents the $(i_3,\cdots,i_d)$-th face slice.  To construct the t-SVD framework, some operators on  $ \tensor{X}\in \mathbb{R}^{m_1\times m_2\times\cdots\times m_d}$ are summarized in  Table \ref{tab:operators}.

\begin{table}[t]
  \caption{Summary of operators for t-SVD} 
  \centering  
  \begin{tabular}{cc}  
    \toprule
    \midrule 
    Notations & Descriptions \\ 
    \midrule

    \multirow{2}{*}{$\operatorname{circ}(\mathcal{X})$} &  $
    \left[\begin{array}{cccc}
    \X_{(:,\cdots,:,1)} & \X_{(:,\cdots,:,n_d)} & \cdots & \X_{(:,\cdots,:,2)} \\
    \X_{(:,\cdots,:,2)} & \X_{(:,\cdots,:,1)} & \cdots & \X_{(:,\cdots,:,3)} \\
    \vdots & \vdots & \vdots & \vdots \\
    \X_{(:,\cdots,:,n_d)} & \X_{(:,\cdots,:,n_d-1)} & \cdots & \X_{(:,\cdots,:,1)}
    \end{array}\right]$, \\
                       & size  $m_1 m_d \times m_2m_d \times m_3 \cdots \times m_{d-1}$   \\ 
    \midrule 
    \multirow{2}{*}{$\operatorname{bcirc}(\X)$} & $\operatorname{bcirc}(\X)=\operatorname{circ}^{d-2}(\X)$,  \\
                       & size  $m_1 \prod_{j=3}^d m_j \times m_2 \prod_{j=3}^d m_j$ \\ 
    \midrule  
    \multirow{2}{*}{$\operatorname{unfold}(\mathcal{X})$ } 
    &  $\left[\begin{array}{c}
    \X_{(:,\cdots,:,1)}  \\
    \X_{(:,\cdots,:,2)} \\
    \vdots  \\
    \X_{(:,\cdots,:,n_d)} \\ 
    \end{array}\right]$,\\ 
    & size  $m_1 m_d \times m_2 \times \cdots \times m_{d-1}$\\
    \midrule                    
    \multirow{2}{*}{$\operatorname{bunfold}(\mathcal{X})$} &   $\operatorname{bunfold}(\X)=\operatorname{unfold}^{d-2}(\X)$,\\
                       & size  $m_1 \prod_{j=3}^d m_j \times m_2$\\
    \midrule                    
    $\operatorname{bfold}(\mathcal{X})$ & the inverse operation of $\operatorname{bunfold}(\mathcal{X})$\\                  
    \midrule 
    \multirow{2}{*}{$\operatorname{bdiag}(\mathcal{X})$} &  $
    \left[\begin{array}{cccc}
    \X_{(:,:,1,\cdots,1)} &  &  &  \\
     & \X_{(:,:,2,\cdots,1)} &  &  \\
     &  & \ddots &  \\
     &  &  & \X_{(:,:,i_3,\cdots,i_d)}
    \end{array}\right]$, \\
                       & size  $m_1\prod_{j=3}^d m_j\times m_2\prod_{j=3}^d m_j$   \\ 
    \midrule

    \bottomrule
  \end{tabular}
  \label{tab:operators}
  \vspace{-0.3cm}
\end{table}

\begin{definition}[T-product \cite{martin2013order}]
Let order-$d$ tensors $\mathcal{A}\in\mathbb{R}^{m_1\times a\times m_3\times\cdots\times m_d}$ and $\mathcal{B}\in\mathbb{R}^{a\times m_2\times m_3\times\cdots\times m_d}$, then the t-product $\mathcal{A}*\mathcal{B}$  is defined to be a tensor
of size $m_1\times m_2\times m_3\times\cdots\times m_d$,
\begin{equation}\label{tprod-bfold}
\mathcal{A}*\mathcal{B}=\operatorname{bfold}(\operatorname{bcirc}(\A)\cdot \operatorname{bunfold}(\B)).
\end{equation}
\end{definition}

\begin{definition}[Circular convolution \cite{fahmy2012new}] For any tensors $\X \in \mathbb{R}^{m_1\times m_2\times\cdots\times m_d}$ and $\K \in \mathbb{R}^{\nu_1 \times \nu_2 \times\cdots\times \nu_d}$ with $\nu_s \leq m_s, s=1,2,\cdots,d$, the circular convolution of two tensors is
\begin{equation}
\Z= \X \star\K \in \mathbb{R}^{m_1\times m_2\times\cdots\times m_d}
\end{equation}
or element-wise,
\begin{equation}
\Z_{(i_1,..,i_d)}=\sum_{j_1=1}^{\nu_1} ...\sum_{j_d=1}^{\nu_d}  \X_{(i_1-j_1+1,..., i_d-j_d+1)} \K_{(j_1,..,j_d)}.
\end{equation}
where $\star$ denotes the circular convolution  operator.
The circulant boundary condition are satisfied by assuming $\X_{(i_1-j_1+1+\delta_1 m_1,..., i_d-j_d+1+\delta_d m_d)}=\X_{(i_1-j_1+1,..., i_d-j_d+1)}$ with $\delta_s=1$ if $i_s < j_s-1$ and otherwise, $\delta_s=0$. 
\label{circular convolution}           
\end{definition}

Since the multiplication of $\operatorname{bcirc}(\A)$ and 
$\operatorname{bunfold}(\B)$ is in the form of circular convolution, t-product can also be defined using circular convolution:
\begin{equation}\label{tprod-conv}
[\mathcal{A}*\mathcal{B}]_{(i_1,i_2,:,\cdots,:)}=\sum_{j=1}^{m_2}\A_{(i_1,j,:,\cdots,:)} \star \B_{(j,i_2,:,\cdots,:)}
\end{equation}
where $\star$ denotes the operator of circular convolution.
Evidently, the t-product of tensors is an operation analogous to matrix multiplication. As a result, many properties of matrix multiplication can be extended to the t-product \cite{lu2019tensor} \cite{qin2022low}.

In addition, the discrete Fourier transform (DFT) can convert the circular convolution operation into an element-wise product.
Therefore, the t-product $\mathcal{A}*\mathcal{B}$ can  be obtained through DFT.
For a given tensor $\X$, its Fourier-transformed in last d-2 dimensions is expressed as
\begin{equation}
\bar{\mathcal{X}}:=\mathcal{F}_{d-2}(\mathcal{X})=\mathcal{X}\times_3 \mathrm{F}_{m_3}\times_4\cdots\times_d \mathrm{F}_{m_d},
\end{equation}
where $\times_j$ denotes the j-mode product \cite{kolda2009tensor} and $\mathrm{F}_{m_j}$ represents the DFT 
matrices of size $m_j \times m_j$ for $j=3,\cdots,d$. Then
\begin{equation}
\mathcal{A}*\mathcal{B}=\mathcal{F}_{d-2}^{-1}(\mathcal{F}_{d-2}(\mathcal{A})\Delta \mathcal{F}_{d-2}(\mathcal{B})),
\end{equation}
where $\Delta$ denotes the face-wise product ($\mathcal{Z}=\mathcal{X}\Delta\mathcal{Y}\Leftrightarrow \mathcal{Z}_{(:,:,i_3,\cdots,i_d)}=\mathcal{X}_{(:,:,i_3,\cdots,i_d)}\mathcal{Y}_{(:,:,i_3,\cdots,i_d)}$ for all face slices).

\begin{definition}[Identity tensor \cite{martin2013order}]
The identity tensor $\mathcal{I}\in\mathbb{R}^{m\times m\times m_3\times\cdots\times  m_d}$
is the tensor with its first face slice $\mathcal{X}_{(:,:,1,\cdots,1)}$ being the identity matrix $m \times m$ and the other face slices being all zeros.
\end{definition}

\begin{definition}[Transpose \cite{martin2013order}]
For an order-$d$   tensor $\mathcal{X}\in\mathbb{R}^{m_1\times m_2\times\cdots\times m_d}$, its transpose $\mathcal{X}^\mathrm{T}\in\mathbb{R}^{m_2\times m_1\times m_3\times\cdots\times m_d}$ is the obtained by tensor transposing each $\mathcal{X}_{(:,\cdots,:,i)}$ for $i=1, \ldots, n_p$ and then reversing the order of the $\mathcal{X}_{(:,\cdots,:,i)}$ $2$ through $n_p$.
\end{definition}

\begin{definition}[Orthogonal tensor \cite{martin2013order}]
An order-$d$ tensor $\mathcal{U}\in\mathbb{R}^{m\times m\times m_3\times\cdots\times  m_d}$ is orthogonal if $\mathcal{U}^\mathrm{T}*\mathcal{U}
=\mathcal{U}*\mathcal{U}^\mathrm{T}=\mathcal{I}$, where $\mathcal{I} \in\mathbb{R}^{m\times m\times m_3\times\cdots\times  m_d}$  is  an order-$d$ identity tensor.
\end{definition}

\begin{definition}[F-diagonal tensor \cite{martin2013order}]
A tensor is said to be f-diagonal if all of its face slices are diagonal.  
\end{definition}

\begin{lemma}[T-SVD \cite{martin2013order}]
 Let tensor $\mathcal{X}\in\mathbb{R}^{m_1\times m_2\times\cdots\times m_d}$, then it can be
factorized as
\begin{equation}\label{eq.6}
\mathcal{X} = \mathcal{U}*\mathcal{S}*\mathcal{V}^\mathrm{T},
\end{equation}
where $\mathcal{U}\in\mathbb{R}^{m_1\times m_1\times\cdots\times m_d}$, $\mathcal{V}\in\mathbb{R}^{m_2\times m_2\times\cdots\times m_d}$ are orthogonal tensors, and $\mathcal{S}\in\mathbb{R}^{m_1\times m_2\times\cdots\times m_d}$ is a f-diagonal tensor.
\end{lemma}

\begin{definition}[Tensor tubal rank) \cite{lu2019tensor} \cite{ qin2022low}]
For $\mathcal{X}\in\mathbb{R}^{m_1\times m_2\times\cdots\times m_d}$ with t-SVD $\mathcal{X} = \mathcal{U} *\mathcal{S} *\mathcal{V}^\mathrm{T}$, its tubal rank is defined as
$$
\begin{aligned}
\operatorname{rank}_t(\mathcal{X}):&=\sharp\{i: \mathcal{S}_{(i, i,:, \cdots,:)} \neq \textbf{0}\},
\end{aligned}
$$
where $\sharp$ denotes the cardinality of a set.
\end{definition}

\begin{definition}[Tensor multi-rank \cite{lu2019tensor} \cite{ qin2022low}]
The multi-rank of a tensor  $\mathcal{X}\in\mathbb{R}^{m_1\times m_2\times\cdots\times m_d}$  is represented by a vector $\mathbf{r}\in\mathbb{R}^{m_3 m_4\cdots m_d}$. The 
$i$-th element of $\mathbf{r}$ corresponds to the rank of the 
$i$-th  block of $\operatorname{bdiag}(\bar{\mathcal{X}})$.
 We denote the tensor multi-rank sum  as $r_s$, which means $r_s=\sum_{i=1}^{m_3 m_4\cdots m_d}\mathbf{r}^{(i)}$.
\end{definition}

\begin{definition}[Tensor spectral norm \cite{lu2019tensor} \cite{ qin2022low}]
For order-$d$ tensor $\mathcal{X}\in\mathbb{R}^{m_1\times m_2\times\cdots\times m_d}$, its tensor spectral norm is defined as
$$
\begin{aligned}
\|\mathcal{X}\| :=
\|\operatorname{bdiag}(\bar{\mathcal{X}})\|,
\end{aligned}
$$
where $\|\cdot\|$ denotes the spectral norm of a  matrix or tensor.
\end{definition}

\begin{definition}[Tensor nuclear norm \cite{lu2019tensor} \cite{ qin2022low}]
For order-$d$ tensor $\mathcal{X}\in\mathbb{R}^{m_1\times m_2\times\cdots\times m_d}$ under the t-SVD framework, its tensor nuclear norm (TNN) is defined as
$$
\begin{aligned}
\|\mathcal{X}\|_{\circledast}:= \frac{1}{m} \norm{\operatorname{bdiag}(\bar{\mathcal{X}})}_*
\end{aligned}
$$
where $m=m_3 m_4\cdots m_d$ and $\|\cdot\|_*$ denotes the nuclear norm of a matrix. Note that the tensor nuclear norm  is the dual norm of its tensor spectral norm, which is consistent with the matrix case \cite{candes2009exact} \cite{lu2019tensor}.
\end{definition}
  
\begin{lemma}[T-SVT \cite{lu2019tensor} \cite{ qin2022low}]\label{th.2}
Given $\mathcal{X}\in\mathbb{R}^{m_1\times m_2\times\cdots\times m_d}$ with t-SVD  $\mathcal{X} = \mathcal{U}*\mathcal{S}*\mathcal{V}^\mathrm{T}$, its tensor singular value thresholding (t-SVT) is defined by $\operatorname{t-SVT}_\tau(\mathcal{X}):=\mathcal{U}*\mathcal{S}_\tau *\mathcal{V}^\mathrm{T}$, where $\mathcal{S}_\tau = \mathcal{F}_{d-2}^{-1}((\bar{\mathcal{S}}-\tau)_+)$, $a_+=\max(0,a)$, which obeys
\begin{equation}\label{eq.9}
\operatorname{t-SVT}_\tau(\mathcal{X})=\arg\min_{\mathcal{Y}} \tau\|\mathcal{Y}\|_{\circledast}+\frac{1}{2}\|\mathcal{Y}-\mathcal{X}\|_\mathrm{F}^2.
\end{equation}
\end{lemma}

\subsection{Deterministic Tensor Completion}
Now we consider deterministic tensor completion problem, a broader research challenge. Let $\M \in \mathbb{R}^{m_1\times\cdots\times m_d}$ represent an unknow target tensor with tubal rank $r$ and skinny t-SVD $\M=\U *\mathcal{S} *\V^\mathrm{T}$. 
Suppose that we observe only  the entries of $\M$ over a deterministic sampling set $\Omega \subseteq \left[m_1\right] \otimes\left[m_2\right] \otimes  \cdots \otimes \left[m_d\right] $.
The corresponding mask tensor is denoted by $\bar{\Omega}$, with the following entry definition: 
$$
[\bar{\Omega}]_{(i_1, \cdots,i_d)}= \begin{cases}1 & \text { if }(i_1, \cdots,i_d) \in \Omega, \\ 0 & \text { otherwise.}\end{cases}
$$
The sampling operator $\Pomega$  is then  defined as 
$\Pomega(\M)=\bar{\Omega} \circ \M,$
where $\circ$ denotes the Hadamard product.

\begin{figure}[t]
\centering
\vspace{-0.2cm}
\includegraphics[width=0.98\linewidth]{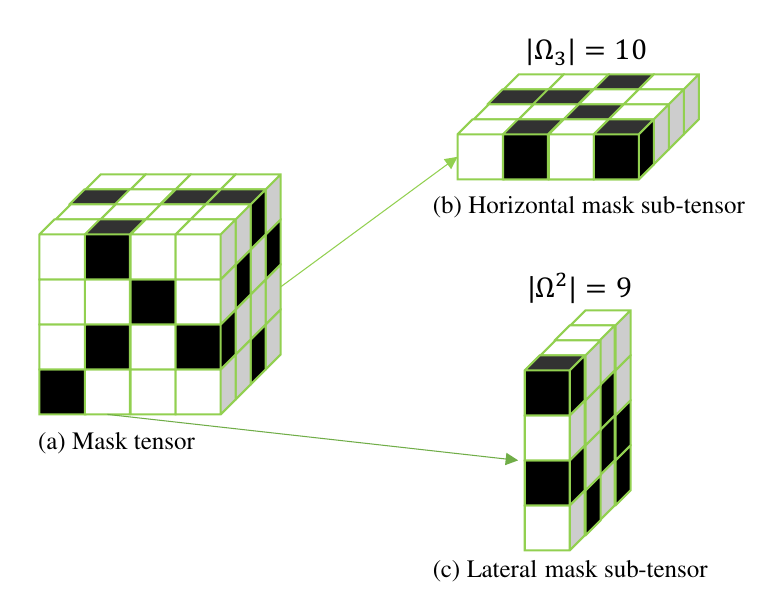}
\vspace{-0.2cm}
\caption{Illustrations of horizontal/lateral mask sub-tensor sampling number in the three-dimensional case.  (a): Arrangement of the mask tensor $\bar{\Omega}$, where the white cubes represent the value 1 (sampled entries) and the black cubes represent the value 0 (unsampled entries); (b):$\left|\bar{\Omega}_{3}\right|=10$ indicates that there are 10 sampled entries in the third horizontal mask sub-tensor; (c): $\left|\bar{\Omega}^{2}\right|=9$ indicates that there are 9 sampled entries in the second lateral mask sub-tensor.}\label{fig:Sampling}
\vspace{-0.5cm}
\end{figure}

To address the deterministic tensor completion problem, we employ a standard constrained  \textit{Tensor Nuclear Norm} (TNN) model:
\begin{equation}\label{LRTC}
\min_{\tensor{X}\in \mathbb{R}^{m_1\times m_2\times\cdots\times m_d}} ~\norm{\X}_{\circledast},\ \text{s.t.} \ \Pomega(\X)= \Pomega(\M),
\end{equation}
where $\Omega $ is a deterministic sampling set.
In the context of tensor completion under random sampling, it is often assumed that the data sampling follows a specific distribution, such as the Bernoulli distribution, then the sampling probability $p$ can naturally serve as a measure of sampling \cite{lu2019tensor}. However, this approach is clearly ineffective for deterministic sampling.  
Inspired by the minimum row/column sampling ratio for matrix deterministic sampling \cite{liu2019matrix}, we introduce a tensor deterministic sampling metric $\rho$, termed the minimum horizontal/lateral sub-tensor sampling ratio. Here, the horizontal/lateral sub-tensors of a tensor are analogous to the rows/columns of a matrix, respectively.
As mentioned above, the tensor t-product, spectral/nuclear norm within the t-SVD framework are defined in a matrix-like manner, providing a solid basis for defining tensor deterministic sampling metric by following matrix case.
\begin{definition}[Horizontal/lateral mask sub-tensor sampling number] 
For any fixed sampling set $\Omega \subseteq \left[m_1\right] \otimes\left[m_2\right] \otimes  \cdots \otimes \left[m_d\right] $
with corresponding mask tensor $\bar{\Omega} \in 
\mathbb{R}^{m_1\times\cdots\times m_d} $, 
the $i_1$-th horizontal mask sub-tensor, i,e., the $i_1$-th horizontal sub-tensor of mask tensor, is given as $\bar{\Omega}_{i_1}:=[\bar{\Omega}]_{(i_1,:,\cdots,:)}$.
The $i_1$-th  horizontal mask sub-tensor sampling number is then defined as
$$
\left|\bar{\Omega}_{i_1}\right|:=\sharp\{(i_2,i_3,\cdots,i_d)|[\bar{\Omega}_{i_1}]_{(i_2,i_3, \cdots,i_d)}=1\}.
$$
Similarly, following $\bar{\Omega}^{i_2} :=\bar{\Omega}_{(:,i_2,\cdots,:)}$, the $i_2$-th  lateral mask sub-tensor sampling number is  defined as
$$
\left|\bar{\Omega}^{i_2}\right|:=\sharp\{(i_1,i_3,\cdots,i_d)|[\bar{\Omega}^{i_2}]_{(i_1,i_3, \cdots,i_d)}=1\}.
$$
where $\sharp$ denotes the cardinality of a set.
\end{definition}

For an intuitive understanding, we provide a schematic diagram illustrating  horizontal/lateral mask sub-tensor sampling number in the three-dimensional case, as shown in Fig.\ref{fig:Sampling}.

\begin{definition}[Minimum horizontal/lateral sub-tensor sampling ratio] For any fixed sampling set $\Omega \subseteq \left[m_1\right] \otimes\left[m_2\right] \otimes  \cdots \otimes \left[m_d\right] $, its minimum horizontal/lateral sub-tensor sampling ratio is defined as the smallest fraction of sampled entries in each horizontal and lateral mask sub-tensor; namely,
\begin{equation}\label{samplingrate}
\rho (\Omega) =\min \left(\min _{1 \leq i_1 \leq m_1} \frac{\left|\bar{\Omega}_{i_1}\right|}{m_2 \times m}, \min _{1 \leq i_2 \leq m_2} \frac{\left|\bar{\Omega}^{i_2}\right|}{m_1 \times m}\right),
\end{equation}
where $m =m_3 m_4  \cdots  m_d$.
\end{definition}

Tensor incoherence is a crucial theoretical tool in low-rank tensor recovery \cite{candes2009exact} \cite{candes2010power} \cite{qin2022low} \cite{zhang2016exact}. It enforces or constrains the low-rank structure of the underlying tensor to avoid ill-posed problems, such as when most elements of $\M$ are zero, yet it is still low-rank.

\begin{definition}[Tensor incoherence conditions \cite{qin2022low}]
For $\M \in\mathbb{R}^{m_1\times\cdots\times m_d}$ with tubal rank $r$ and it has  the  skinny t-SVD $\M=\U *\mathcal{S} *\V^\mathrm{T}$, and then $\M$ is said to satisfy the tensor incoherence conditions with parameter $\mu>0$ if
\begin{equation}\label{incoherence1}
\max_{i_1 = 1,\cdots,m_1} \|\U^\mathrm{T}*\mathring{\mathfrak{e}}_1^{(i_1)}\|_\mathrm{F}\leq
\sqrt{\frac{\mu r}{m_1 m}},
\end{equation}
\begin{equation}\label{incoherence2}
\max_{i_2 = 1,\cdots,m_2} \|\V^\mathrm{T}*\mathring{\mathfrak{e}}_2^{(i_2)}\|_\mathrm{F}\leq
\sqrt{\frac{\mu r}{m_2 m}},
\end{equation}
where  $m =m_3 m_4  \cdots  m_d$, $\mathring{\mathfrak{e}}_1^{(i_1)}$ is the order-$d$ tensor mode-1 basis sized $m_1\times1\times m_3\times\cdots\times m_d$, whose $(i_1,1,i_3,\cdots,i_d)$-th entry equals 1 and the rest equal 0, and $\mathring{\mathfrak{e}}_2^{(i_2)}:=(\mathring{\mathfrak{e}}_1^{(i_2)})^\mathrm{T}$ is the mode-2 basis.
\end{definition}

\subsection{Exact Recovery Guarantee}
We briefly show the exact recovery guarantee of  deterministic low-rank tensor completion problem. 
\begin{theorem}\label{thm:exact tensor completion}
 Suppose that $\M \in \mathbb{R}^{m_1\times\cdots\times m_d}$  obeys the standard  tensor incoherence conditions (\ref{incoherence1})-(\ref{incoherence2}), $\Omega \subseteq \left[m_1\right] \otimes\left[m_2\right] \otimes  \cdots \otimes \left[m_d\right] $ and $\rho (\Omega)$ is its minimum horizontal/lateral sub-tensor sampling ratio.
if 
\begin{equation}\label{sampling}
\rho(\Omega) > 1-\frac{1}{ 2 \mu r (r_s+1)},
\end{equation}
where $r$ is the tubal rank of tensor $\M$, $r_s$ is the multi-rank sum and  $\mu$ is the parameter of  tensor incoherence conditions  , then  $\M$ is the unique solution to  the TNN  model (\ref{LRTC}).
\end{theorem}
The above result demonstrates that minimizing the tensor nuclear norm can achieve exact deterministic tensor completion, provided that condition (\ref{sampling}) is satisfied. The feasibility of exact recovery hinges upon the interplay between the sampling set and the tensor rank. As the low-rank structure of the tensor strengthens, the minimum slice sampling ratio required for exact deterministic tensor completion decreases correspondingly. Moreover, the proposed deterministic exact recovery theory can be extended to the random sampling case. The following corollary reveals that when the sampling set satisfies $\Omega\sim\operatorname{Ber}(p)$ and $p\geq  1-1/{ 2 \mu r (r_s+1)}$, exact recovery holds with high probability.

\begin{proposition}\label{Ber}
Suppose that $\M \in \mathbb{R}^{m_1\times\cdots\times m_d}$ obeys the standard  tensor incoherence conditions (\ref{incoherence1})-(\ref{incoherence2}) and $\Omega\sim\operatorname{Ber}(p)$.  if
$p\geq  1-1/{ 2 \mu r (r_s+1)}$, then $\M$ is the unique solution to the TNN  model (\ref{LRTC}) with probability at least $1-e^{-4a^2 m_0}$, 
where  $a=p-1+1/{ 2 \mu r (r_s+1)}, m_0=m_1m_2\cdots m_d$.
\end{proposition}

\section{Multidimensional Time Series Prediction}\label{sec:Method and Model}
Let us denote the multidimensional time series as $\left\{\M_i\right\}_{i=1}^{t}$, where $\M_i \in \mathbb{R}^{ n_1\times\cdots\times n_p} $. The objective is to predict the next $h$ unseen samples $\{\M_i \}_{i=t-h+1}^{t}$ given the historical portion $\{\M_i \}_{i=1}^{t-h}$, where $t$ represents the number of samples in the time dimension and $h$ denotes the number of samples to be predicted, referred to as the forecast horizon. 

\subsection{Time series prediction via TNN}
Multidimensional time series prediction, as a specific instance of deterministic tensor completion, can be addressed using the aforementioned tensor nuclear norm minimization, which results in the following model:
\begin{equation}\label{pred-to-comp}
\min_{\tensor{X}\in \mathbb{R}^{t\times n_1\times\cdots\times n_p}} ~\norm{\X}_{\circledast},\ \text{s.t.} \ \Pomega(\X)= \Pomega(\M),
\end{equation}
where  $\Omega =\left[t-h\right] \otimes \left[n_1\right]  \otimes  \cdots \otimes \left[n_p\right]$ represents the deterministic  sampling set corresponding to the historical portion sample, $\Pomega(\cdot)$ is the projection operator. $\M\in \mathbb{R}^{t\times n_1\times n_2\times\cdots\times n_p}$ denotes the multidimensional time series to be completed.

Deterministic tensor completion theory can be applied to prediction tasks where the sampling set 
$\Omega$ is chosen as the historical data region in the prediction problem. However, in such cases, the minimum horizontal/lateral sub-tensor sampling ratio
\begin{equation}
\rho(\Omega) = 0,
\end{equation}
which fails to meet the recovery conditions (\ref{sampling}). This implies that tensor nuclear norm minimization cannot be directly applied to prediction tasks, as it would result in prediction outcomes that are entirely zero. The core issue lies in the concentrated nature of missing data in time series forecasting, which leads to insufficient sampling coverage. A natural solution to this problem is to implement a structural transformation that disperses the missing data, thereby increasing the minimum horizontal/lateral sub-tensor sampling
ratio above zero. To address this, we introduce the temporal convolution tensor for multidimensional time series. This approach ensures that each horizontal/lateral sub-tensor contains a sufficient number of sampled elements, as illustrated in 
Fig.\ref{fig:temporal convolution}.

\subsection{Temporal Convolution Tensor}

In this study, circular convolution (\ref{circular convolution}) is applied only along the time dimension of the multidimensional time series, a process referred to as temporal circular convolution. Following this, we obtain the temporal convolution tensor of the multidimensional time series.
\begin{definition} [Temporal circular convolution operator]
\label{def:Temporal Circular Convolution Operator}
 The temporal circular convolution procedure of converting $\M \in \mathbb{R}^{t \times n_1 \times\cdots\times n_p}$ and $\mathbf{k} \in \mathbb{R}^k(k \leq t)$ into $\M \star_t \mathbf{k} \in \mathbb{R}^{t \times n_1 \times\cdots\times n_p}$ is expressed as follows:
\begin{equation}
[\M \star_t \mathbf{k}]_{(i,i_1,...,i_p)}=\sum_{j=1}^k\M_{(i-j+1,i_1,...,i_p)}\mathbf{k}_{(j)},
\end{equation}
where $\star_t$ denotes the temporal circular convolution  operator, $\mathbf{k}$ is the kernel vector, and it is assumed
that $\M_{(i-j+1,i_1,...,i_p)}=\M_{(i-j+1+t,i_1,...,i_p)}$ for $i+1 \leq j$,  which is the temporal circulant boundary condition.   
\end{definition}

\begin{definition} [Temporal Convolution Tensor]
\label{def:Temporal Convolution Tensor}
The temporal convolution tensor $\mathcal{T}_k(\M)  \in \mathbb{R}^{t \times k\times n_1 \times\cdots\times n_p}$ of 
multidimensional time 
series $\M \in \mathbb{R}^{t \times n_1 \times\cdots\times n_p}$  
can be obtained via temporal circular convolution:
\begin{equation}
[\M \star_t \mathbf{k}]_{(:,i_1,...,i_p)}=[\mathcal{T}_k(\M)]_{(:,:,i_1,...,i_p)} \mathbf{k},
\end{equation}
where $\star_t$ denotes the temporal circular convolution operator, $\mathcal{T}_k(\cdot)$ is the temporal convolution transform, 
and the temporal convolution tensor $\mathcal{T}_k(\M)$ is given by:
$$
\mathcal{T}_k(\M)=\left[\begin{array}{cccc}
\M^1 & \M^t & \cdots & \M^{t-k+2} \\
\M^2 & \M^1 & \cdots & \M^{t-k+3} \\
\vdots & \vdots & \vdots & \vdots \\
\M^t & \M^{t-1} & \cdots & \M^{t-k+1}
\end{array}\right],
$$
where $\M^i \in \mathbb{R}^{ 1 \times 1 \times n_1 \times\cdots\times n_p}$ represents  the information of multidimensional time series $\M$ at the $i$-th time sampling point, in other words, $\M^i=reshape(\M_i, 1, 1, n_1, \cdots, n_p)$.
\end{definition}

\begin{figure}[t]
\centering
\vspace{-0.2cm}
\includegraphics[width=0.98\linewidth]{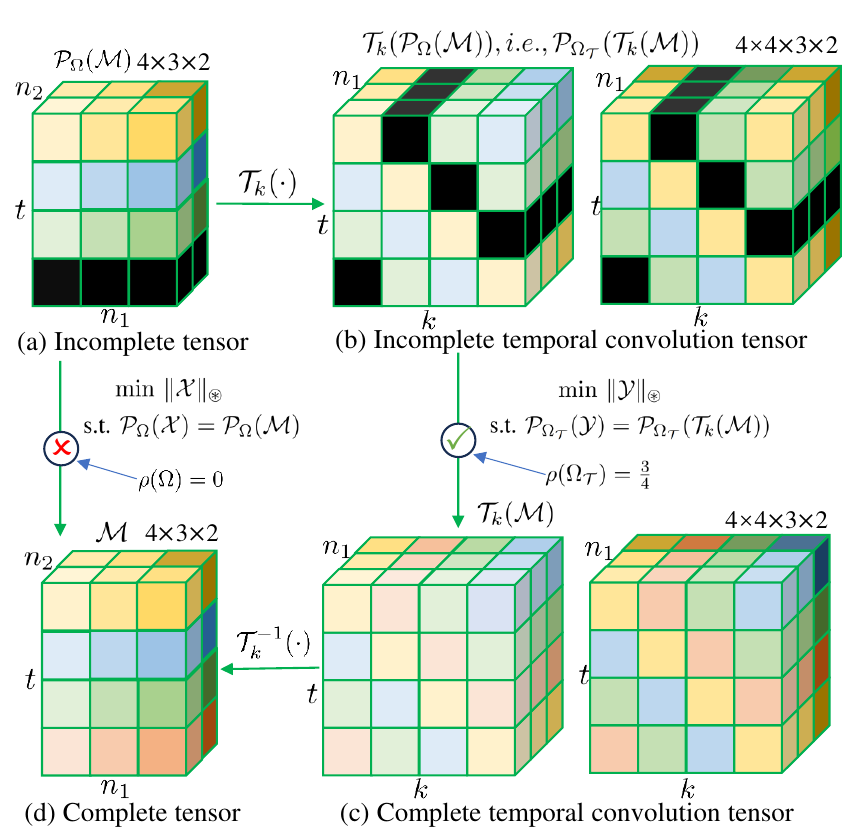}
\vspace{-0.2cm}
\caption{Illustrations of temporal convolution tensor nulclear norm
for multidimensional time series prediction. (a): Incomplete tensor, where black squares represent 0 values (not observed) and other color blocks represent non-zero values (observed); (b): incomplete temporal convolution tensor; (c): complete temporal convolution tensor; (d): complete tensor. }\label{fig:temporal convolution}
\vspace{-0.5cm}
\end{figure}

\begin{definition}[Temporal convolution sampling set]\label{Temporal Convolution Sampling Set} 
For a sampling set of prediction problem $\Omega =\left[t-h\right] \otimes \left[n_1\right]  \otimes  \cdots \otimes \left[n_p\right] $, its temporal convolution sampling set is denoted by $\Omega_{\mathcal{T}}$ and given by
\begin{equation}
\bar{\Omega}_{\mathcal{T}}=\mathcal{T}_k\left(\bar{\Omega}\right) \text { and } \Omega_{\mathcal{T}}=\operatorname{supp}\left(\bar{\Omega}_{\mathcal{T}}\right),
\end{equation}
where $\bar{\Omega} \in \mathbb{R}^{t \times n_1 \times \cdots \times n_p}$  is the mask tensor
of $\Omega$ and $\bar{\Omega}_{\mathcal{T}} \in \mathbb{R}^{t \times k\times n_1 \times \cdots \times n_p}$ is the mask tensor of $\Omega_{\mathcal{T}}$.
Note that the consistency between sampling after transformation and transformation after sampling:
$\mathcal{T}_k (\mathcal{P}_{\Omega}(\M))=\Pomegat(\mathcal{T}_k(\M))$, where $\Pomegat(\Y)=\bar{\Omega}_{\mathcal{T}} \circ \Y$, $\forall \Y\in \mathbb{R}^{t \times k\times n_1 \times\cdots\times n_p}.$ 
\end{definition}
To provide an intuitive understanding, we present a schematic diagram illustrating the conversion from a multidimensional time series (incomplete tensor) of size 
 $4 \times 3 \times 2$ to the corresponding  temporal convolution tensor of size $4 \times 4 \times 3\times 2$, as shown in Fig.\ref{fig:temporal convolution}(a,b).  
Notably, the last horizontal sub-tensor of the incomplete tensor $\Pomega(\M)$  consists entirely of zeros. However, after applying the temporal convolution transform $\mathcal{T}_k(\cdot)$, each horizontal and lateral slice of the incomplete temporal convolution tensor $\mathcal{T}_k(\Pomega(\M))$ contains a sufficient number of sampled entries. As depicted in Fig.\ref{fig:temporal convolution}(b), the temporal convolution sampling set achieves a minimum horizontal/lateral sub-tensor sampling ratio
\begin{equation}
\rho(\Omega_{\T}) =3/4.
\end{equation}
This significantly disperses the unobserved regions, enhancing the potential for axact predictions.

\subsection{Temporal Convolution Tensor Low-rankness  from Smoothness and  Periodicity}
\begin{figure*}[t]
\centering
\vspace{-0.1cm}
\includegraphics[width=0.98\linewidth]{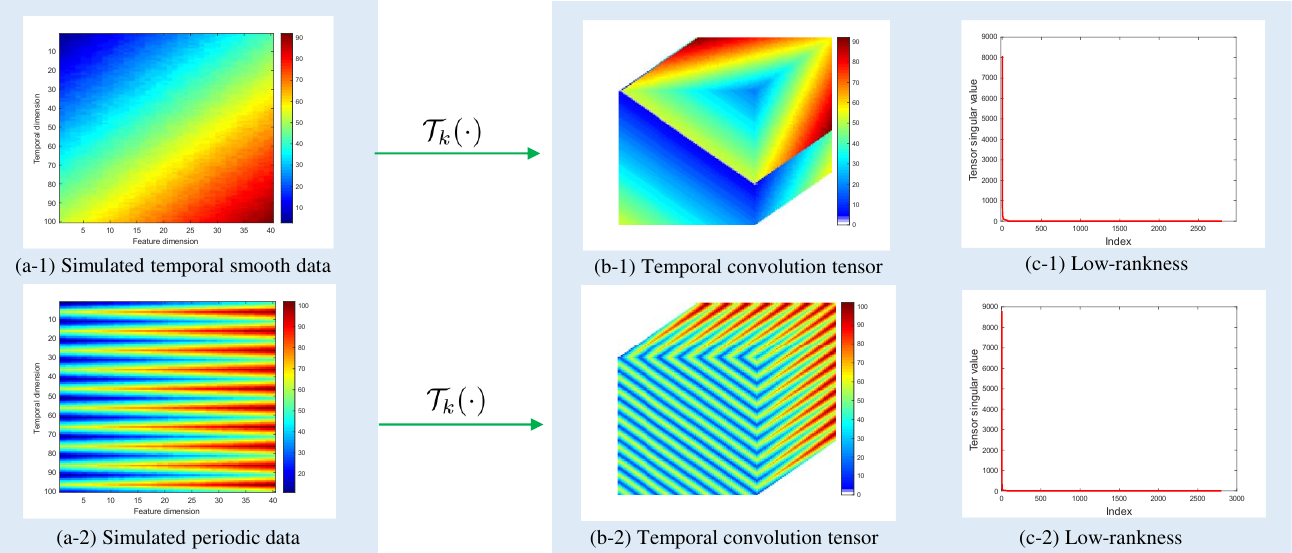}
\vspace{-0.2cm}
\caption{Illustrations of temporal convolution low-rankness from smoothness and periodicity using simulated data. }\label{fig:PS-TCPS}
\vspace{-0.5cm}
\end{figure*}
By applying transform $\mathcal{T}_k(\cdot)$ to the incomplete  multidimensional time series $\Pomega(\M)$, we obtain the incomplete temporal convolution tensor $\Pomegat(\mathcal{T}_k(\M))$  with a relatively dispersed distribution of missing values.  In many cases, the completion of a tensor relies on its low-rankness, and this holds true for the temporal convolution tensor $\mathcal{T}_k(\M)$ as well.
Fortunately, The low-rankness of the temporal convolution tensor $\mathcal{T}_k(\M)$ can often be satisfied in various scenarios, as it can be derived from the smoothness and periodicity of the time series $\M$.
To reach a rigorous conclusion, we consider the rank-r approximation error of temporal convolution tensor $\mathcal{T}_k(\M) \in \mathbb{R}^{t \times k \times n_1 \times\cdots\times n_p} $, which is denoted by $\varepsilon_r(\mathcal{T}_k(\M))$ and defined as
$$
\varepsilon_r(\mathcal{T}_k(\M))=\min _{\Y}\|\mathcal{T}_k(\M)-\Y\|_F, \text { s.t. } \operatorname{rank}_{t}(\Y) \leq r,
$$
where $\Y \in \mathbb{R}^{t \times k \times n_1 \times\cdots\times n_p}$.

\begin{lemma}\label{smoothness to lowrankness}
  We can find an indicator $\eta(\M)$ to represent the strength of the smoothness of tensor  $\M$ along the temporal dimension, which can be defined as 
$$ 
\begin{aligned}
\eta(\M)= \left(\sum_{i=1}^{t-1} ~\norm{\M_{i+1}-\M_i}_F^2\right)^{\frac{1}{2}},
\end{aligned}
$$
then 
$$
\varepsilon_r\left(\mathcal{T}_k(\M)\right)  \leq \sqrt{\frac{t(k+r)}{3}}\left\lceil\frac{k}{r}\right\rceil\eta(\M).
$$
\end{lemma}

\begin{lemma}\label{periodicity to lowrankness}
If the approximate period is $\tau$,  we can find an indicator $\beta_{\tau}(\M)$ to represent the strength of the periodicity of tensor  $\M$ along the time dimension, which can be defined as 
$$ \beta_{\tau}(\M)= \max_{i\in 1,...,t} \norm{\M_i-\M_{i+\tau}}_F,$$ where $\M_i=\M_{i-t}$ for $i>t$
then $$
\varepsilon_{\tau}\left(\mathcal{T}_k(\M)\right) \leq \tau t(\left\lceil\frac{k}{\tau}\right\rceil-1) \beta_{\tau}(\M).
$$
\end{lemma}
The notation $\left\lceil \cdot \right\rceil$ in aboves lemmas denotes the ceiling operation.
Lemma \ref{smoothness to lowrankness}  and Lemma \ref{periodicity to lowrankness} indicate that as the periodicity and smoothness of the original data matrix/tensor $\M$ along the time dimension increase, the low-rankness of its temporal convolution tensor $\mathcal{T}_k(\M)$ becomes more pronounced.
To illustrate this concept intuitively, we generated two multivariate time series: one exhibiting strong periodicity and the other characterized by temporal smoothness, as shown in Fig.\ref{fig:PS-TCPS}(a-1,a-2), respectively.  Temporal convolution transform were applied to these time series matrices, and the resulting temporal convolution tensors are depicted in Fig.\ref{fig:PS-TCPS}(b-1,b-2). As shown in Fig.\ref{fig:PS-TCPS}(c-1,c-2), the tensor singular values of the temporal convolution tensors decrease rapidly. This trend highlights the enhanced low-rankness of the temporal convolution tensors, further substantiating the theoretical findings.

\subsection{Temporal Convolution Tensor Nulclear Norm}

Given that the smoothness and periodicity of the original time series tensor $\M$ ensure the low-rank property of the temporal convolution tensor $\mathcal{T}_k(\M)$, it is natural to leverage the convex relaxation of tensor rank—the tensor nuclear norm—to encode this low-rankness. This leads to the formulation of the following completion model:
\begin{equation}\label{DTC}
    \begin{aligned}
    \min_{\tensor{Y}\in \mathbb{R}^{t \times k \times n_1 \times\cdots\times n_p}} ~\norm{\Y}_{\circledast},
    ~\text { s.t.} ~ \Pomegat(\Y)= \Pomegat(\mathcal{T}_k(\M)). \\
    \end{aligned}
\end{equation}
to obtain $\hat{\Y}$ as the estimator of $\T_k(\M)$. After that, we perform the inverse temporal convolution on $\hat{\Y}$, thereby obtaining $\T_k^{-1}(\hat\Y)$ as the final preditior of $\M$.

This logical flow is depicted in Fig.\ref{fig:temporal convolution}, outlining the steps to obtain the complete tensor 
 from the incomplete tensor $\Pomega(\M)$ in the prediction problem. The process consists of three main steps:
1.Transformation: Apply the temporal convolution transform $\mathcal{T}_k(\cdot)$ to the incomplete tensor $\Pomega(\M)$, resulting in the incomplete temporal convolution tensor P$\Pomegat(\mathcal{T}_k(\M))$;
2. Completion: Solve model (\ref{DTC}) to derive the complete temporal convolution tensor $\mathcal{T}_k(\M)$ from the incomplete tensor $\Pomegat(\mathcal{T}_k(\M))$;
3. Inverse Transformation: Retrieve the complete tensor $\M$  by applying the inverse transform $\mathcal{T}_k^{-1}(\cdot)$.
By consolidating these steps into an equivalent representation, we formulate the \textit{Temporal Convolution Tensor Nuclear Norm} (TCTNN) model for multidimensional time series prediction, i.e., 
\begin{equation}\label{TSP}
\min_{\tensor{X}\in \mathbb{R}^{t \times n_1 \times\cdots\times n_p}} ~\norm{\mathcal{T}_k(\X)}_{\circledast},\ \text{s.t.} \ \Pomega(\X)= \Pomega(\M).
\end{equation}
where $\Omega =\left[t-h\right] \otimes \left[n_1\right]  \otimes  \cdots \otimes \left[n_p\right] $ and $h$ is the forecast horizon.

\subsection{Exact Prediction Guarantee}
We now develop the exact prediction gaurantee for time series data prediction by applying the deterministic tensor completion theory established in the preceding section.  We first define the temporal convolution tensor incoherence conditions as follows:
\begin{definition} [Temporal convolution tensor incoherence]
For $\M \in\mathbb{R}^{ t\times n_1\times\cdots\times n_p}$, assume that the temporal convolution tensor $\mathcal{T}_k(\M)\in\mathbb{R}^{ t \times k\times n_1\times\cdots\times n_p}$   with tubal rank $r_{\T}$ and it has  the skinny t-SVD $\mathcal{T}_k(\M)=\U * \mathcal{S} * \V^\mathrm{T}$, and then $\M$ is said to satisfy the temporal convolution tensor incoherence conditions with kernel size $k$ and parameter $\mu_{\T}>0$ if
\begin{equation}\label{incoherence1_convolution}
\max_{i_t = 1,\cdots,t} \|\U^\mathrm{T}*\mathring{\mathfrak{e}}_t^{(i_t)}\|_\mathrm{F}\leq
\sqrt{\frac{\mu_{\T} r_{\T}}{t n}},
\end{equation}
\begin{equation}\label{incoherence2_convolution}
\max_{i_k = 1,\cdots,k} \|\V^\mathrm{T}*\mathring{\mathfrak{e}}_k^{(i_k)}\|_\mathrm{F}\leq
\sqrt{\frac{\mu_{\T} r_{\T}}{k n}},
\end{equation}
where $n =n_1 \times  \cdots \times n_p$,$\mathring{\mathfrak{e}}_t^{(i_t)}$ is the order-$(p+2)$ tensor sized $t\times1\times n_1\times\cdots\times n_p$, whose $(i_t,1,i_1,\cdots,i_p)$-th entry equals 1 and the rest equal 0, and $\mathring{\mathfrak{e}}_k^{(i_k)}:=(\mathring{\mathfrak{e}}_t^{(i_t)})^\mathrm{T}$ .
\end{definition}

Based on the temporal convolution tensor incoherence, we derive the exact prediction theory for the TCTNN model (\ref{TSP}) from the deterministic tensor completion recovery theory. 
\begin{theorem}\label{thm: prediction exact}
 Suppose that  $\M \in \mathbb{R}^{t\times n_1\times\cdots\times n_p}$  obeys the  temporal convolution tensor incoherence conditions   (\ref{incoherence1_convolution})-(\ref{incoherence2_convolution}) with kernel size k. If
\begin{equation}\label{sampling-pr}
h < \frac{ k}{ 2 \mu_{\T} r_{\T} ((r_s)_{\T}+1)},
\end{equation}
where $r_{\T}$ is the tubal rank of tensor $\mathcal{T}_k(\M)$ , $(r_s)_{\T}$ is the multi-rank sum of tensor $\mathcal{T}_k(\M)$  and  $\mu_{\T}$ is the parameter of  temporal convolution tensor incoherence conditions, then  $\M$ is the unique solution to  the TCTNN model (\ref{TSP}).
\end{theorem}
Theorem \ref{thm: prediction exact} states that exact prediction   is promising when the temporal convolution low-rankness is satisfied.  As the low-rankness of the temporal convolution tensor increases and the prediction horizon $h$ decreases, achieving exact predictions becomes increasingly feasible. According to Lemmas \ref{periodicity to lowrankness} and \ref{smoothness to lowrankness}, stronger periodicity and smoothness in the temporal dimension contribute to greater low-rankness in the temporal convolution tensor, thereby improving the likelihood of exact prediction.

\subsection{Comparison with the CNNM model}
In very recent, Liu et al. \cite{liu2022recovery} propose the \textit{convolution matrix nulclear norm minimization} (CNNM) model 
 which can be formulated as
\begin{equation}
\min_{\tensor{X}\in \mathbb{R}^{t \times n_1 \times\cdots\times n_p}} ~\norm{\mathcal{C}_{\mathbf{s}}(\X)}_*,\ \text{s.t.} \ \Pomega(\X)= \Pomega(\M),
\end{equation}
where $\mathbf{s}=[s_t,s_1,\cdots,s_p]$ is kernel size, $\M$ represents the time series to be completed,  $\|\cdot\|_*$ denotes the nuclear norm of a matrix and 
$\mathcal{C}_{\mathbf{s}}(\cdot)$ denotes the transformation that converts the time series into the convolution matrix.

 In contrast to the CNNM model, the proposed TCTNN model does not convert the original data tensor into convolution matrices along each dimension. Instead, we perform tensor convolution along the temporal dimension of the time series data. Our approach offers advantages in three aspects.  1. Tensor structure information: It preserves the tensor structure of multi-dimensional time series and avoids the information loss caused by converting tensor data into matrices. As demonstrated by the fact that $[\mathcal{T}_k(\X)](:,1,:,:)=\X$, our temporal convolution tensors maintain the structure information of the multidimensional time series, unlike the convolution matrices, which satisfy $[\mathcal{C}_{\mathbf{s}}(\X)](:,1)=vec(\X)$.
 2. Computational complexity: It avoids the computationally expensive task of processing the nuclear norm of very large matrices, significantly reducing the computational cost. For example, to predict a $(p+1)$-order multidimensional time series $\tensor{M}\in \mathbb{R}^{n\times\cdots\times n}$, the computational complexity of each iteration of the CNNM model is $O(n^{3p+3})$, while the TCTNN model requires only $O(n^{p+3})$ per iteration. Clearly, the computational complexity of the former is nearly prohibitive.
 3. Feature utilization: By performing convolution exclusively along the time dimension, our method is better suited for time series data. Many time series exhibit strong characteristics such as periodicity and smoothness only along the time dimension. Convolution along all dimensions could lead to improper feature capture, thereby reducing prediction accuracy.

\section{Optimization Algorithm}\label{sec:Algorithm}
This section gives the optimization algorithm of the proposed TCTNN model (\ref{TSP}) for the multidimensional  time series prediction task.
\subsection{Optimization to TCTNN}
We adopt the famous optimization framework \textit{alternating direction method of multipliers} (ADMM) \cite{boyd2011distributed,bai2018generalized,bai2022inexact} to solve the model (\ref{TSP}). First of all, we introduce the auxiliary variable  $\Y=\mathcal{T}_k(\X)$ to separate the temporal convolution
transform $\mathcal{T}_k(\cdot)$,
then the TCTNN model (\ref{TSP}) can be rewritten as below:
\begin{equation}
    \begin{aligned}
    &\min_{\X,\Y} \   \norm{\Y}_{\circledast} ,\\
    ~\text { s.t.} ~\Y=&\mathcal{T}_k(\X), \Pomega(\X)= \Pomega(\M). \\
    \end{aligned}
    \label{tctnn_au}
\end{equation}
For model \eqref{tctnn_au}, its  partial augmented Lagrangian function  is
\begin{equation} \label{tctnn_lf}
    \begin{aligned}
	L(\X,\Y,\N)  = &\norm{\Y}_{\circledast}+ \langle \N,\Y-\mathcal{T}_k(\X) \rangle +
 \frac{\mu_{\ell}}{2}\normlarge{\Y-\mathcal{T}_k(\X)}_F^2\\
  =&\norm{\Y}_{\circledast}+ 
 \frac{\mu_{\ell}}{2}\normlarge{\Y-\mathcal{T}_k(\X)+{\N}/{\mu_{\ell}}}_F^2+\C\\
    \end{aligned}  
\end{equation}
where  $\N$ are Lagrange multipliers, $\mu_{\ell}>0$ is a positive scalar, and $\C$ is only the multipliers dependent squared items.
According to the  ADMM optimization framework,   the minimization problem  of \eqref{tctnn_lf} can be transformed into the following subproblems for each variable in an iterative manner, i.e.,
\begin{align}
\Y^{{\ell}+1}&=\operatorname{arg}\min_{\Y} ~L(\X^{\ell},\Y,\N^{\ell}),\label{subb1}
\end{align}
\begin{equation}\label{subb2}
\begin{aligned}
\X^{{\ell}+1}&=\operatorname{arg}\min_{\X} ~L(\X,\Y^{{\ell}+1},\N^{\ell}),\\
\text{s.t.}~&\Pomega(\X^{{\ell}+1})= \Pomega(\M),
\end{aligned}
\end{equation}
and the multipliers is updated by
\begin{align}
\N^{{\ell}+1}&=\N^{{\ell}}+\mu_{\ell}(\Y^{{\ell}+1}-\mathcal{T}_k(\X^{{\ell}+1})),\label{subb4} 
\end{align}
where ${\ell}$ denotes the count of iteration in the ADMM. In the following, we deduce the solutions for \eqref{subb1} and \eqref{subb2} respectively, each of which has the closed-form solution.

\textit{1) Updating $\Y^{{\ell}+1} $}: Fixed other variables in \eqref{subb1}, then it degenerates into the following  tensor nuclear norm minimization with respect to $\Y$, i.e.,
\begin{equation}
    \begin{aligned}
   \Y^{{\ell}+1}=\operatorname{arg}\min_{\Y} \frac{1}{\mu_{\ell}}\norm{\Y}_{\circledast}+
 \frac{1}{2}\normlarge{\Y-(\mathcal{T}_k(\X)-{\N^{\ell}}/{\mu_{\ell}})}_F^2,
    \end{aligned}\label{TNN-PRO}
\end{equation}
the close-form solution of this sub-problem is given as
\begin{align}\label{GJ1}
\mathcal{Y}^{{\ell}+1} = \operatorname{t-SVT}_{1/\mu_{\ell}}(\mathcal{T}_k(\X^{{\ell}})-{\N^{\ell}}/{\mu_{\ell}})
\end{align}
via the order-$(p+2)$ t-SVT as stated in Lemma \ref{th.2}.

\textit{2) Updating $\X^{{\ell}+1}$}:
By deriving simply the KKT conditions, the closed-form solution of \eqref{subb2} is given by
\begin{equation}\label{XJ1}
   \X^{{\ell}+1} =\Pomega(\M)+\Pomegac(\mathcal{T}_k^{-1}(\Y^{{\ell}+1}+\N^{\ell}/\mu_{\ell})),
\end{equation}
where  $\Omega^{\perp}$  is the complement of $\Omega$ and  $\mathcal{T}_k^{-1}(\cdot)$ is the inverse operator of $\mathcal{T}_k(\cdot)$.  The whole optimization procedure for solving  proposed TCTNN  model \eqref{TSP} is summarized in Algorithm \ref{alg1}.

\begin{algorithm}[tbp]\vspace{-1mm}
\renewcommand{\algorithmicrequire}{ \textbf{Input}:}
\renewcommand{\algorithmicensure}{ \textbf{Output}:}
\caption{ADMM for solving the TCTNN  model \eqref{TSP}}\label{alg1}
\begin{algorithmic}[1]
\REQUIRE observed tensor $\Pomega(\M)$, $k$.
\STATE Initialize $\X^0=\Pomega(\M)$, $\Y^0=\mathcal{T}_k(\X)$, $\N^0=\0$, $\mu_0=1e-5$.
\STATE \textbf{while} not converge \textbf{do}
\STATE \quad Update $\Y^{{\ell}+1}$ by \eqref{GJ1};
\STATE \quad Update $\X^{{\ell}+1}$ by \eqref{XJ1};
\STATE \quad Update multipliers $\N^{{\ell}+1}$  by \eqref{subb4} ;
\STATE \quad Let $\mu_{{\ell}+1}=1.1\mu_{\ell}$; ${\ell} = {\ell} +1$.
\STATE \textbf{end while}
\ENSURE imputed tensor ${\X}^{*}=\X^{{\ell}+1}$.
\end{algorithmic}
\end{algorithm}
\subsection{Computational Complexity Analysis}

For Algorithm 1, the computational complexity in each iteration contains three parts, i.e., steps $3\sim5$. First, the time complexity for order-$(p+2)$ t-SVT in step 3 is $O(tk(n_1\cdots n_p)(n_1+\cdots+n_p)+tk^2n_1\cdots n_p)$, corresponding to the linear transform  and the matrix SVD, respectively \cite{qin2022low}. The steps 4 and 5 have the same complexity $O(tkn_1\cdots n_p)$ with only element-wise computation. In all, the pre-iteration computational complexity of Algorithm 1 is $O(tk(n_1\cdots n_p)(n_1+\cdots+n_p)+tk^2n_1\cdots n_p)$.

\subsection{Convergence Analysis}

Using optimality conditions and the saddle point theory of the Lagrangian function, we derive the following convergence theory for Algorithm 1.
\begin{theorem}\label{Convergence1}
The sequence $(\X^{{\ell}},\G^{{\ell}},\N^{\ell})$ generated by Algorithm 1 converges to a feasible solution of the TCTNN model (\ref{TSP}), and the corresponding objective function converges to the optimal value 
$ p^* = \norm{\Y^{*}}_{\circledast} $, where $(\X^{*},\Y^{*},\N^*)$ is an optimal solution of model (\ref{TSP}).
\end{theorem}

\section{Experimental results}
\label{sec:experiments}

In this section, we present a comprehensive experimental evaluation of the proposed \textit{temporal convolution tensor nulclear norm} (TCTNN) approach using several real-world multidimensional time series datasets. 

\subsection{Multidimensional Time Series Datasets}
Throughout the experiments, we adopt the following three  real-world multidimensional time series  datasets which are widely used in peer works. 
\begin{enumerate}

     \item \textbf{Pacific}: The Pacific surface temperature dataset\footnote{\url{http://iridl.ldeo.columbia.edu/SOURCES/.CAC/}}. This dataset is sourced from sea surface temperature on the Pacific over 50 consecutive months from January 1970 to February 1974.
     The spatial locations are represented by a grid system with a resolution of 2 × 2 latitude-longitude. The grid comprises 30 × 84 cells, resulting in a three-dimensional temperature data tensor with dimensions of 50 × 30 × 84, where 50 corresponds to the temporal dimension, while 30 and 84 represent the latitudinal and longitudinal grid counts, respectively.

    \item \textbf{Abilene}: The Abilene network flow dataset\footnote{\url{http://abilene.internet2.edu/observatory/data-collections.html}}.  
     Regarding the Abilene dataset, it encompasses 12 routers. Network traffic data for each Origin-Destination (OD) pair is recorded from 12:00 AM to 5:00 AM on March 1, 2004, with a temporal resolution of 5 minutes. We structure this data into a tensor with dimensions 60 × 12 × 12, where the first mode represents 60 time intervals, the second mode denotes 12 source routers, and the third mode signifies 12 destination routers.   
     
    \item \textbf{NYC taxi}: The New York taxi ride information Dataset\footnote{\url{https://www1.nyc.gov/site/tlc/about/tlc-trip-record-data.page}}. For the experimental phase, we selected trip data collected from 12:00 AM on May 1st to 02:00 AM on May 3rd, 2018. The raw data is subsequently organized into a third-order tensor, structured as (pick-up zone × drop-off zone × time slot). We delineate a total of 30 distinct pick-up/drop-off zones, and adopt a temporal resolution of 1 hour for trip aggregation. Consequently, the resulting spatiotemporal tensor exhibits dimensions of 50 × 30 × 30.

\end{enumerate}

\subsection{Baseline Methods}
To evaluate the proposed TCTNN method, we compare it with the following baselines:
\begin{enumerate}

\item SNN: The Sum of Nuclear Norms \cite{liu2012tensor}. This \textit{Low-Rank Tensor Completion} (LRTC) method employs the minimization of the sum of nuclear norms of unfolded matrices of tensors to accurately estimate unobserved or missing entries in tensor data.

\item TNN: Tensor Nuclear Norm \cite{zhang2016exact}. This method achieves low-rank tensor completion by minimizing the tensor nuclear norm, thereby facilitating the reconstruction of missing data.

\item BTTF: Bayesian Temporal Tensor Factorization \cite{chen2021bayesian}. This method employs tensor factorization and incorporates a Gaussian Vector Autoregressive (VAR) process to characterize the temporal factor matrix.

\item BTRTF: Bayesian Temporal Regularized Tensor Factorization \cite{yu2016temporal}. The \textit{temporal regularized matrix factorization} (TRMF)  method discussed in the literature is applicable only to multivariate time series forecasting and is not suitable for multidimensional time series. Therefore, we employ its tensor variant method.

\item CNNM:  Convolution  Nulclear Norm Minimization  \cite{liu2022recovery}. This method convolves the multidimensional time series tensor along each dimension into matrices and minimizes the matrix nuclear norm of the convolution matrix to achieve temporal data prediction.
\end{enumerate}

\subsection{Experiment Setup}
For all the parameters in the aforementioned baselines, we set them according to their released code and perform further fine-tuning to achieve optimal performance. For the proposed  
TCTNN method, we use the parameter $k=t/2$ in all datasets. In the evaluation criteria setting,
We use the MAE and RMSE to assess the prediction performance, where lower values of MAE and RMSE indicate more precise predictions.

\subsection{Experimental Results}

\begin{table*}[!ht]
\renewcommand{\arraystretch}{1}
\centering
\fontsize{10pt}{12pt}\selectfont
\caption{Performance comparison (in MAE/RMSE) ) of TCTNN and other baseline models for multidimensional time series prediction across various scenarios. The forecast horizon (FH) column indicates the respective forecast horizons.}\label{tab:TCTNN}
\label{MAE/RMSE}
\begin{tabularx}{\textwidth}{XX|XX|XXXX}
\toprule
Dataset&FH&SNN&TNN&BTTF&BTRTF&CNNM&TCTNN\\
\midrule

\multirow{5}{*}{\textbf{Pacific}}
& h=2  & 25.55/25.69& 25.55/25.69& \underline{1.52}/\underline{1.89}& 1.57/1.95 &  1.67/2.10   &\textbf{0.63}/\textbf{0.83} \\
& h=4  &  25.55/25.69 & 25.55/25.69 & 1.88/2.39 & 2.51/2.94& \underline{1.70}/\underline{2.21} &  \textbf{0.84}/\textbf{1.13} \\
& h=6  &  25.46/25.65 &25.46/25.65& 
2.70/3.25 &  2.55/2.99& \underline{1.95}/\underline{2.46}  & \textbf{1.07}/\textbf{1.39} \\
& h=8& 25.39/25.59  &25.39/25.59 &
2.84/3.69 & 4.45/5.04 & \underline{2.23}/\underline{2.79}  & \textbf{1.28}/\textbf{1.65}  \\
& h=10  & 25.43/25.62 &25.43/25.62&
3.43/4.21& 5.55/6.82& \underline{2.29}/\underline{2.90}  & \textbf{1.34}/\textbf{1.70} \\

\midrule

\multirow{5}{*}{\textbf{Abilene}}
& h=2 & 2.77/4.78&2.77/4.78&  0.49/0.78& 0.48/0.80 & \underline{0.40}/\underline{0.71}   & \textbf{0.40}/\textbf{0.65} \\
& h=4 & 2.78/4.87& 2.78/4.87&  0.56/1.01& 0.53/0.88 & \underline{0.48}/\underline{0.91}   & \textbf{0.48}/\textbf{0.83}  \\
&  h=6  & 2.77/4.85  & 2.77/4.85&  0.58/1.03& 0.59/1.06&\textbf{0.49}/\underline{1.00} & \underline{0.52}/\textbf{0.96} \\
&  h=8& 2.79/4.85  & 2.79/4.85&  0.62/1.07& 0.73/1.31&\textbf{0.52}/\underline{1.01}   & \underline{0.55}/\textbf{0.98}\\
&  h=10 & 2.78/4.84  & 2.78/4.84&  0.60/1.02& 0.65/1.12&\textbf{0.51}/\underline{1.00}  & \underline{0.57}/\textbf{0.99}\\

\midrule

\multirow{5}{*}{\textbf{NYC taxi}}
& h=2   & 4.49/7.88& 4.49/7.88& 2.89/3.98&  3.30/4.69 &\underline{2.73}/\underline{4.04}    & \textbf{2.54}/\textbf{3.48}\\
& h=4  & 7.37/12.62 &7.37/12.62 &  3.40/\underline{4.99}&
3.93/6.54 & \underline{3.23}/5.09  & 
 \textbf {3.05}/\textbf{4.43} \\
& h=6  & 8.89/15.08  & 8.89/15.08  &  \underline{3.38}/\underline{5.21}& 4.44/ 6.83& 3.96/6.59   & \textbf{3.24}/\textbf{4.78} \\
& h=8  & 10.01/17.42 & 10.01/17.42&  \textbf{3.47}/\textbf{5.31}&  6.28/10.11& 5.30/9.28   & \underline{3.55}/\underline{5.38} \\
& h=10& 10.17/17.96 & 10.17/17.96& \underline{3.67}/\underline{5.72}&  7.73/13.39& 5.89/10.85   & \textbf{3.57}/\textbf{5.55} \\

\bottomrule
\multicolumn{8}{l}{{The best results are highlighted with \textbf{bold} and the second best are highlighted with \underline{underline}}.}
\end{tabularx}
\vspace{-0.5cm}
\end{table*}

In assessing the proposed TCTNN model under various conditions, forecast horizons are established at 2, 4, 6, 8, and 10 for all the multidimensional time series datasets.
 We summarize the predictive results of the TCTNN model alongside other baseline
 methods in Table \ref{tab:TCTNN}. The first two models are low-rank tensor completion models, while the latter four are tensor-based prediction models. From Table  \ref{tab:TCTNN}, we observe that our proposed TCTNN model outperforms the other baseline methods in most testing scenarios, with the CNNM and BTTF models being the next best performers. Notably, the poor prediction results of the SNN and TNN models confirm that low-rank tensor completion models are not suitable for prediction problems.
The running times of different methods across various datasets, with the prediction domain set to 10, are summarized in Table \ref{tab:time}. From Table \ref{tab:time}, it is evident that our proposed TCTNN model is more time-efficient than the CNNM model, which is consistent with the previous complexity analyses. 

\begin{table}[!htbp]
	\centering
\renewcommand{\arraystretch}{1}
\setlength{\tabcolsep}{3pt}
\fontsize{9pt}{10pt}\selectfont
	\caption{Summary of the running time (in seconds) of different methods for prediction tasks across various datasets. Time less than 1 second is counted as 1 second. }\label{tab:time}
	\begin{tabular}{c|c c c c c }
		\hline
 Method & Abilene & NYC taxi &Pacific \\
\hline 
SNN &1 & 1 & 2 \\
TNN & 1& 1& 2 \\
BTTF  & 17& 72& 114 \\
BTRTF& 9& 67&119\\
CNNM  & 28& 418 & 18192 \\
TCTNN  & 7& 25& 104 \\
\hline
 \end{tabular}\\
 \vspace{-0.4cm}
\end{table}

\begin{figure}[!htbp]
\renewcommand{\arraystretch}{0.5}
\setlength\tabcolsep{0.5pt}
\centering
\begin{tabular}{ccccccc}
\centering

\scriptsize \textbf{May 2nd, 11:00 PM}  & \scriptsize \textbf{May 3rd, 2:00 AM}\\

\includegraphics[width=39.3mm, height = 35.3mm]{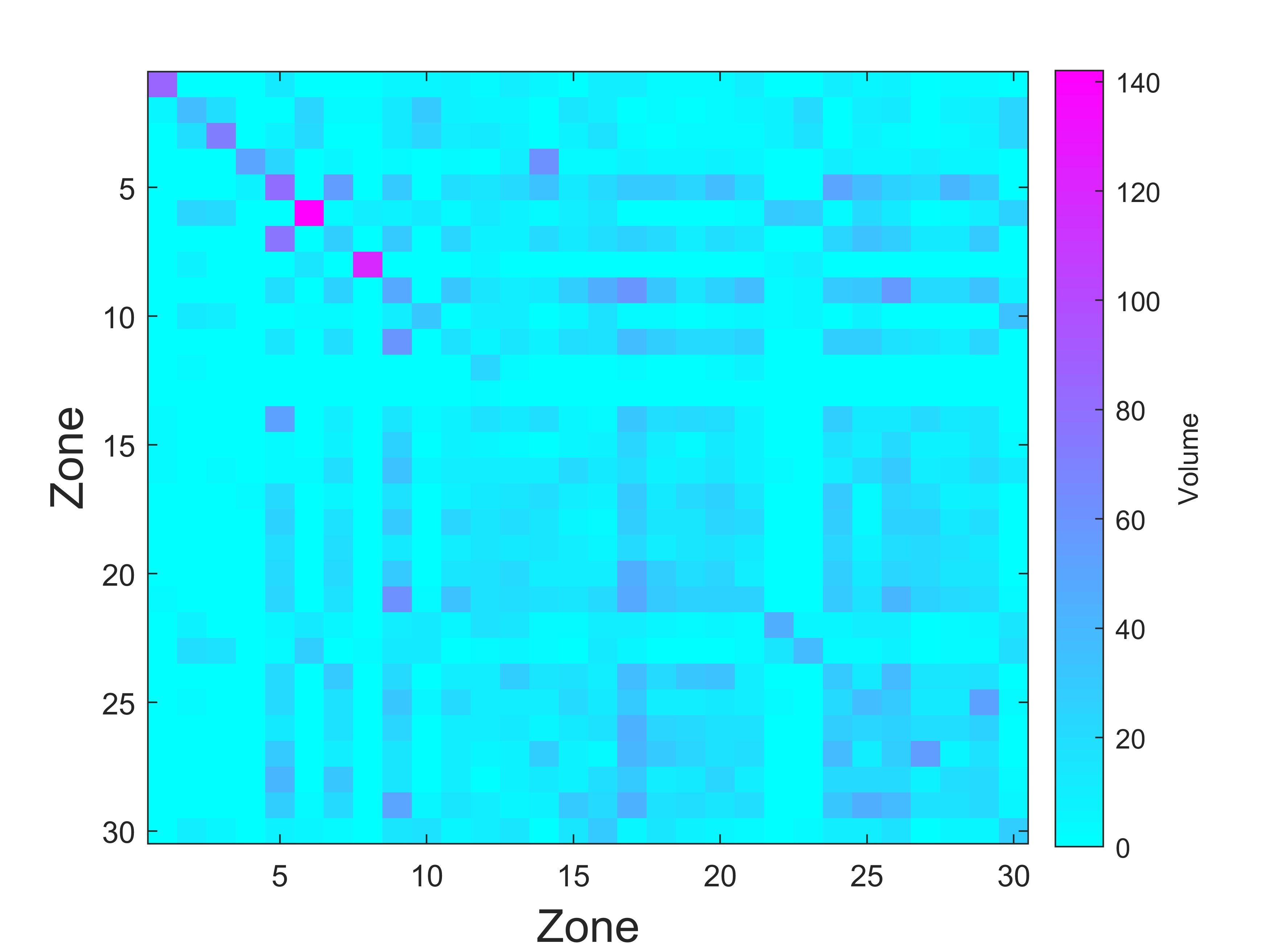}&
\includegraphics[width=39.3mm, height = 35.3mm]{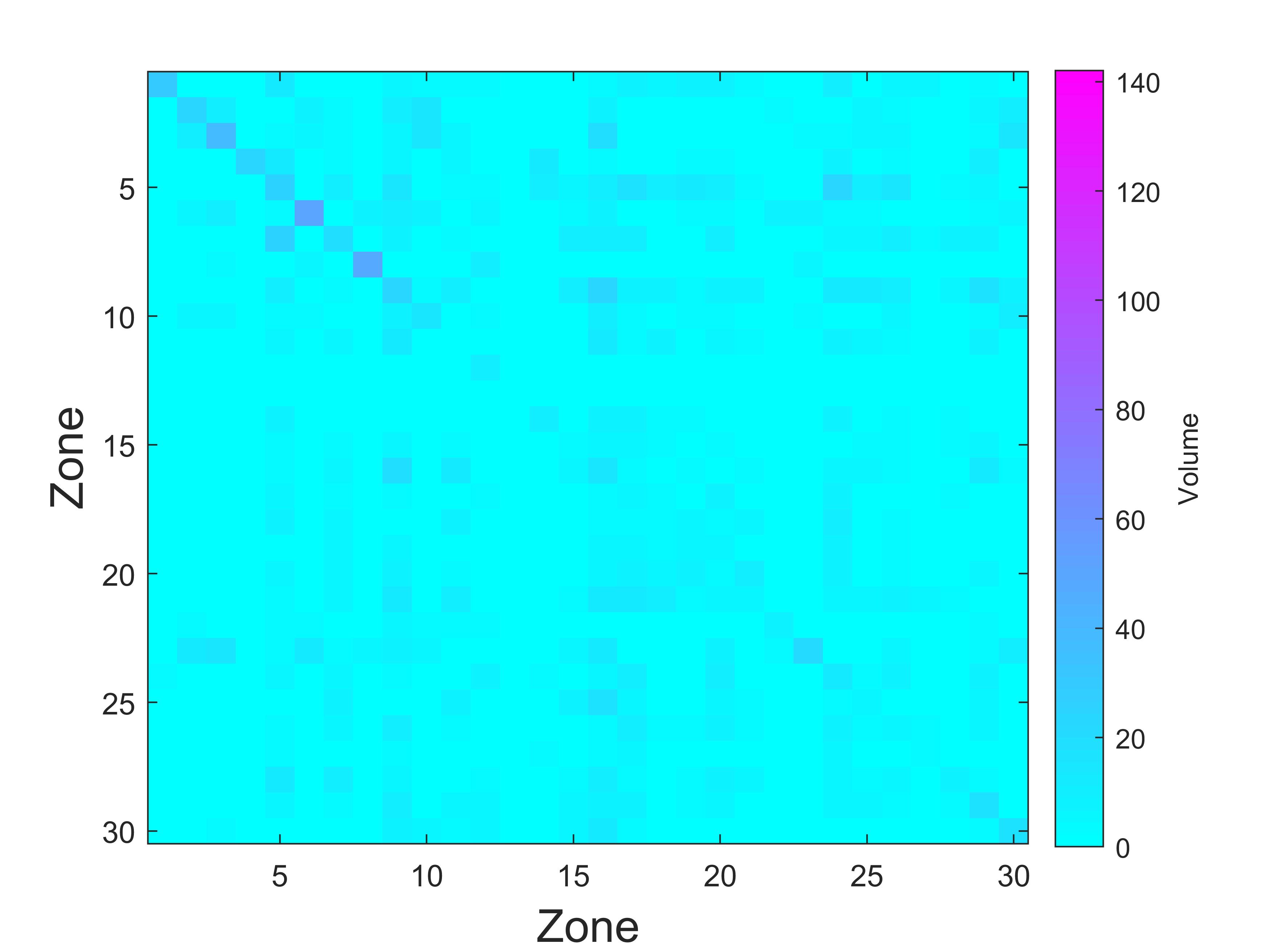}\\

\scriptsize \textbf{Ground truth 1}& \scriptsize \textbf{Ground truth 2}\\

\includegraphics[width=39.3mm, height = 35.3mm]{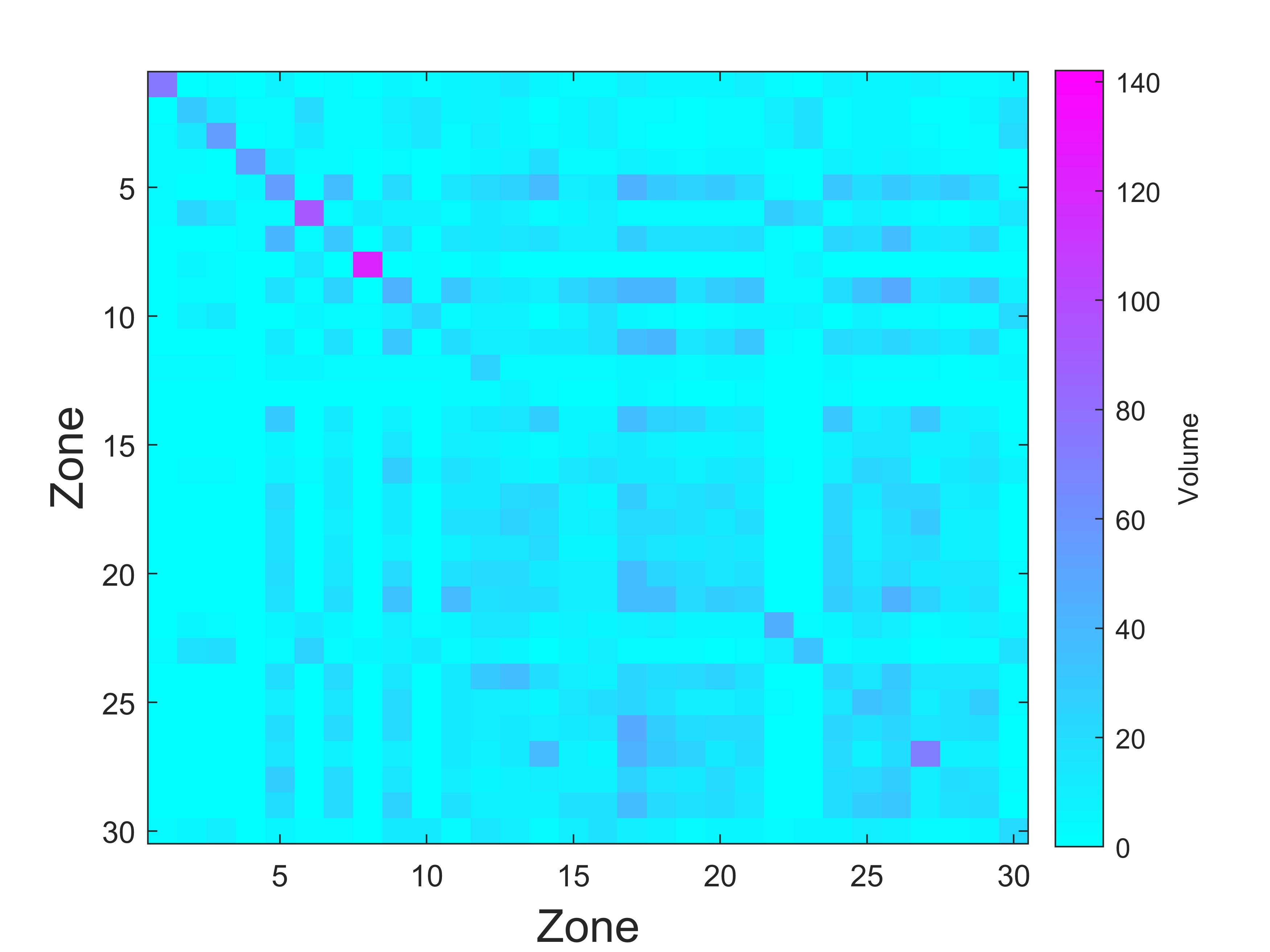}&
\includegraphics[width=39.3mm, height = 35.3mm]{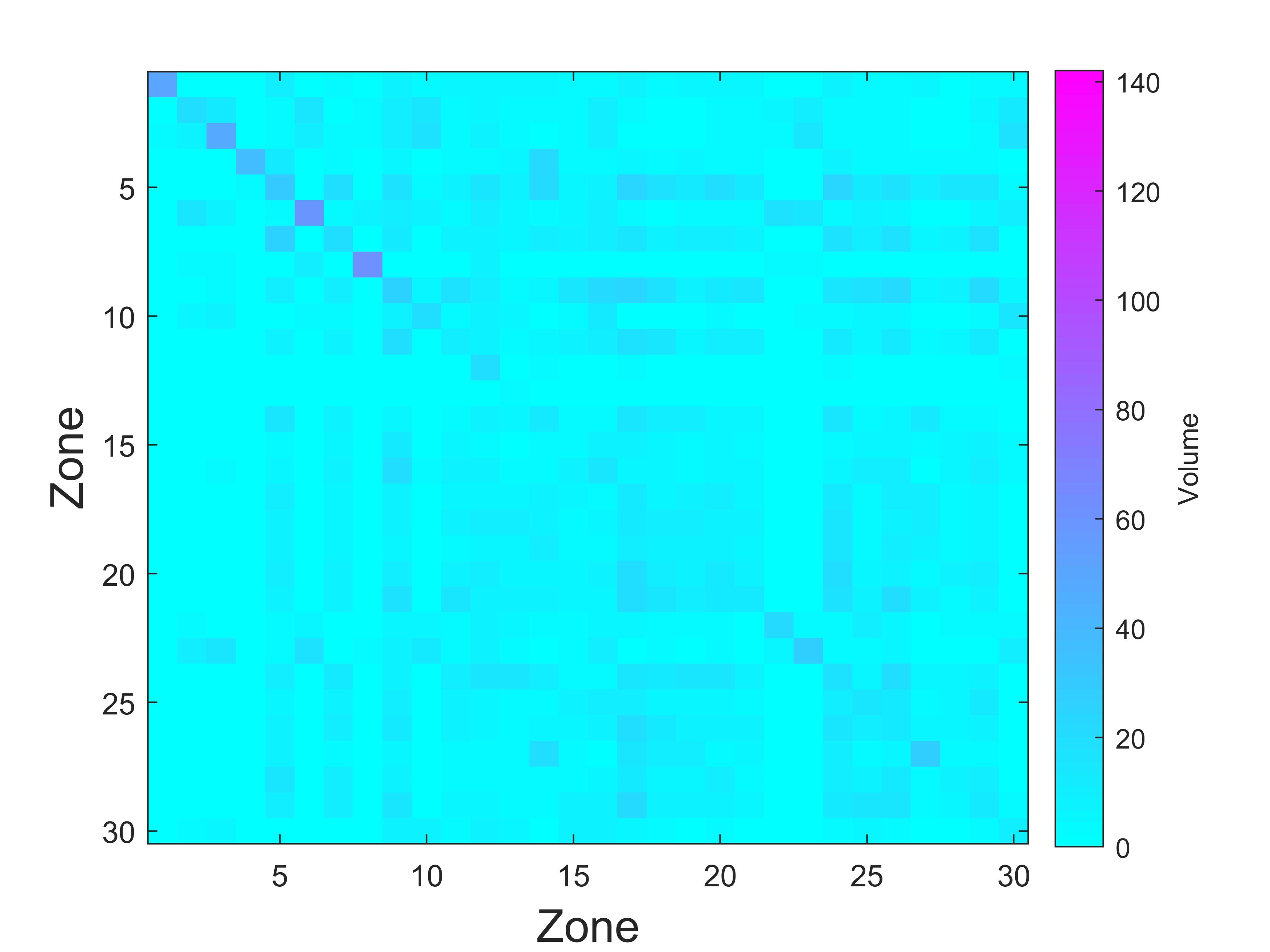}\\

\scriptsize \textbf{TCTNN 1}& \scriptsize \textbf{TCTNN 2}\\
\end{tabular}
\vspace{-0.2cm}
\caption{The predicted results of the TCTNN model and the corresponding true values under forecast horizon 4 on the NYC taxi dataset. Ground truth 1 and Ground truth 2 represent the true values at two distinct time points, while TCTNN 1 and TCTNN 2 denote the predicted values at the corresponding time points. }\label{fig.NYTAXI}
\vspace{-0.5cm}
\end{figure}

\begin{figure}[!htbp]
\renewcommand{\arraystretch}{0.5}
\setlength\tabcolsep{0.5pt}
\centering
\begin{tabular}{ccccccc}
\centering

\includegraphics[width=39.3mm, height = 35.3mm]{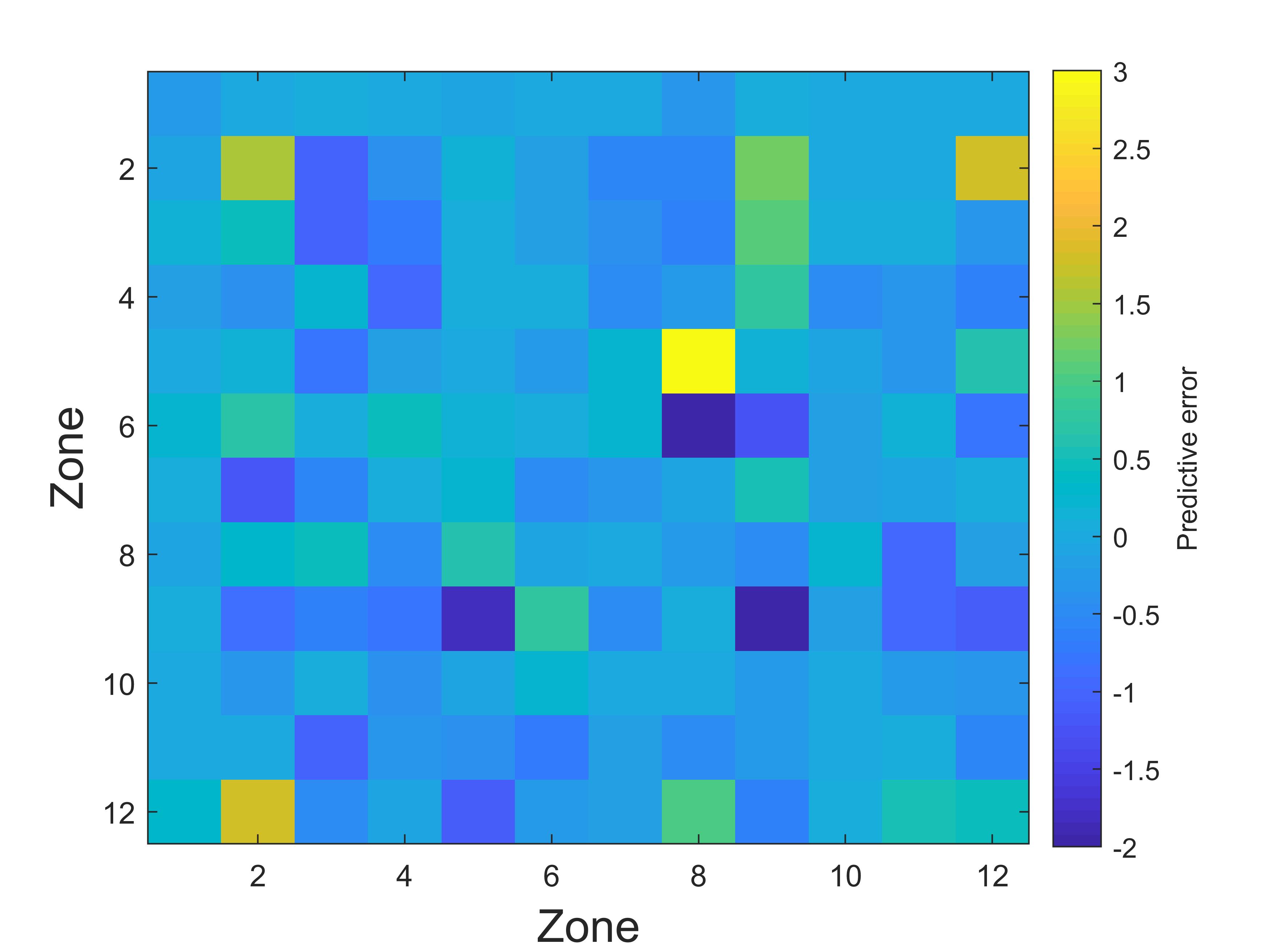}&
\includegraphics[width=39.3mm, height = 35.3mm]{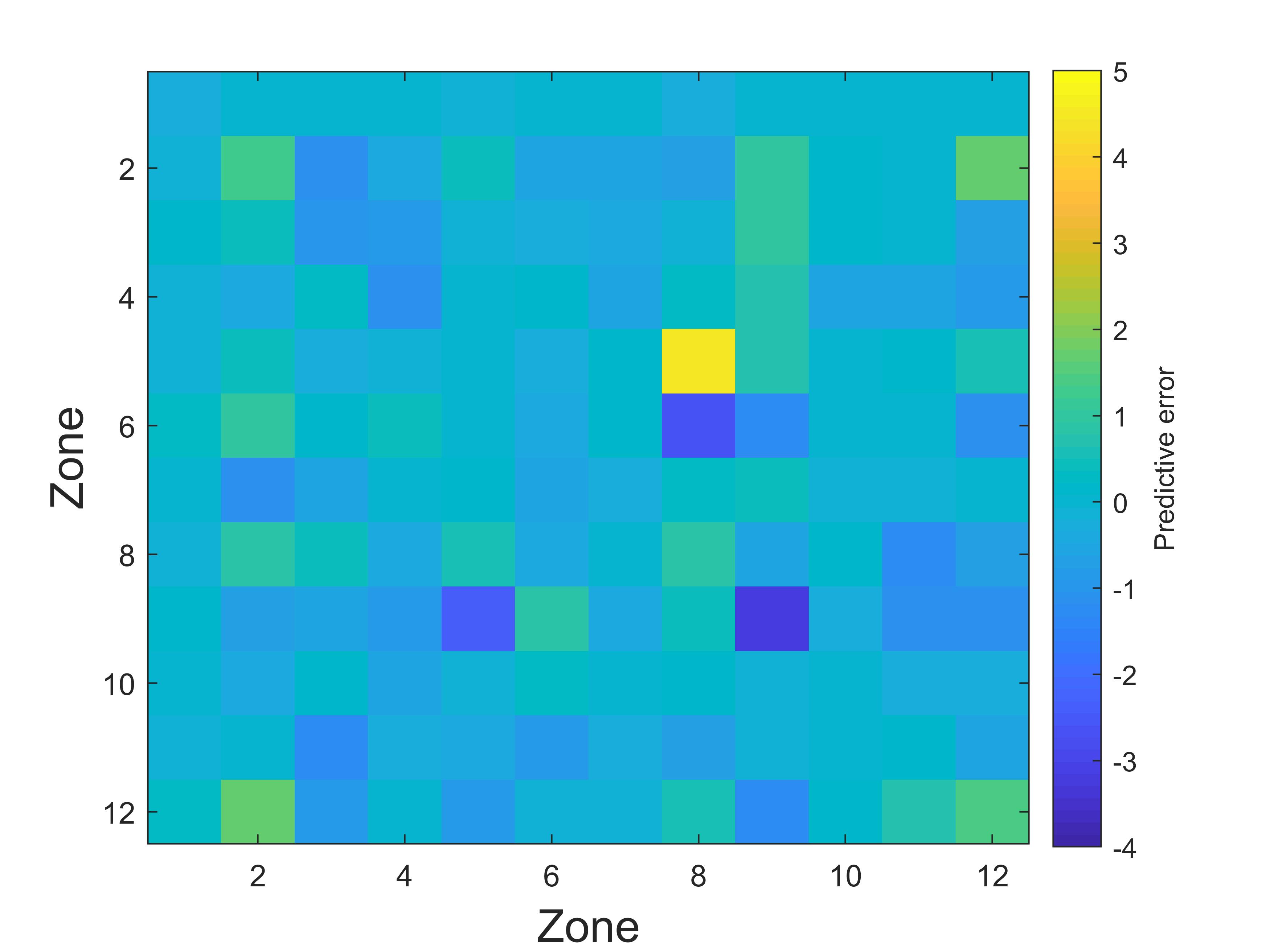}\\

\scriptsize \textbf{h=4}& \scriptsize \textbf{h=6}\\

\includegraphics[width=39.3mm, height = 35.3mm]{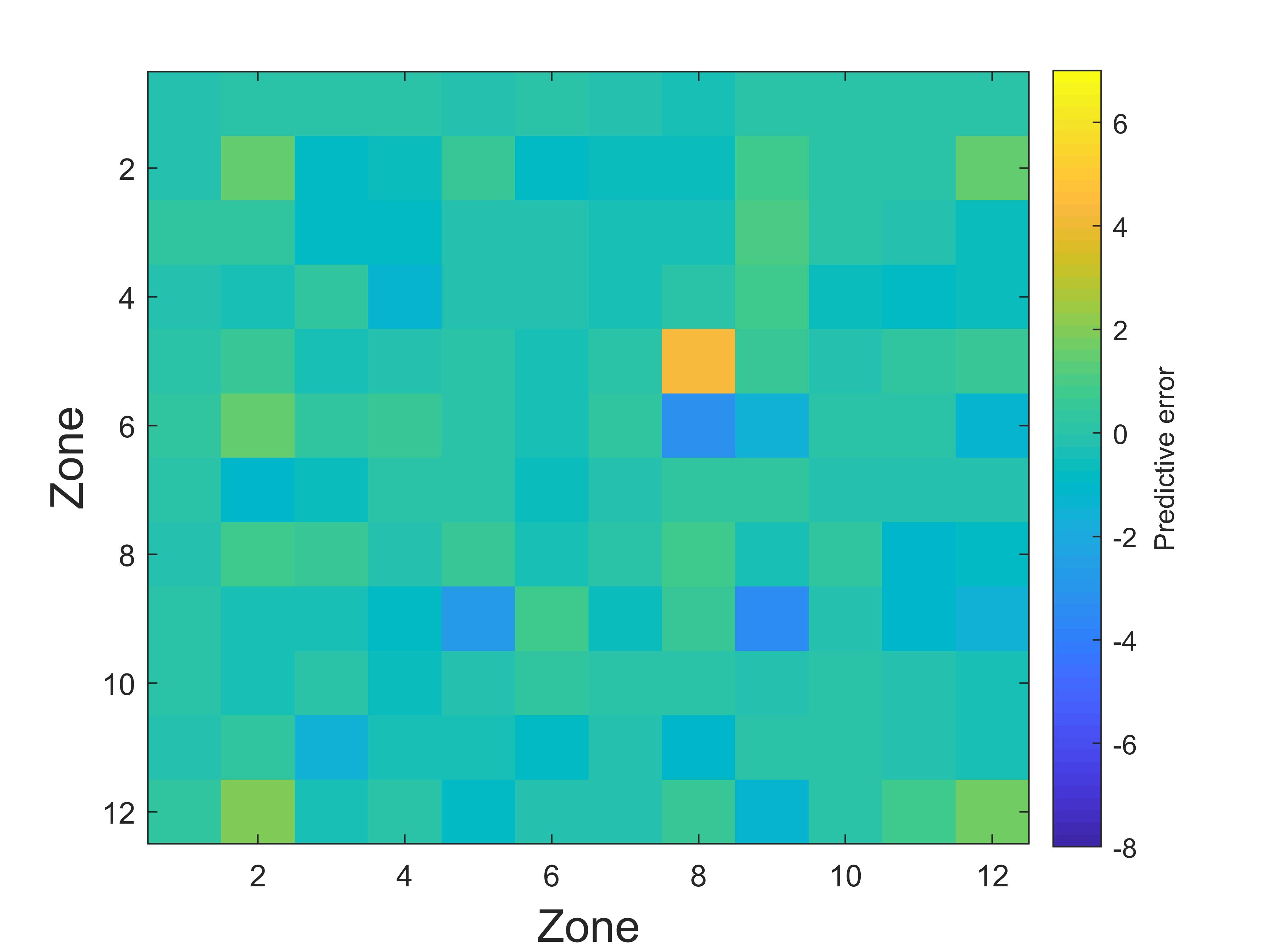}&
\includegraphics[width=39.3mm, height = 35.3mm]{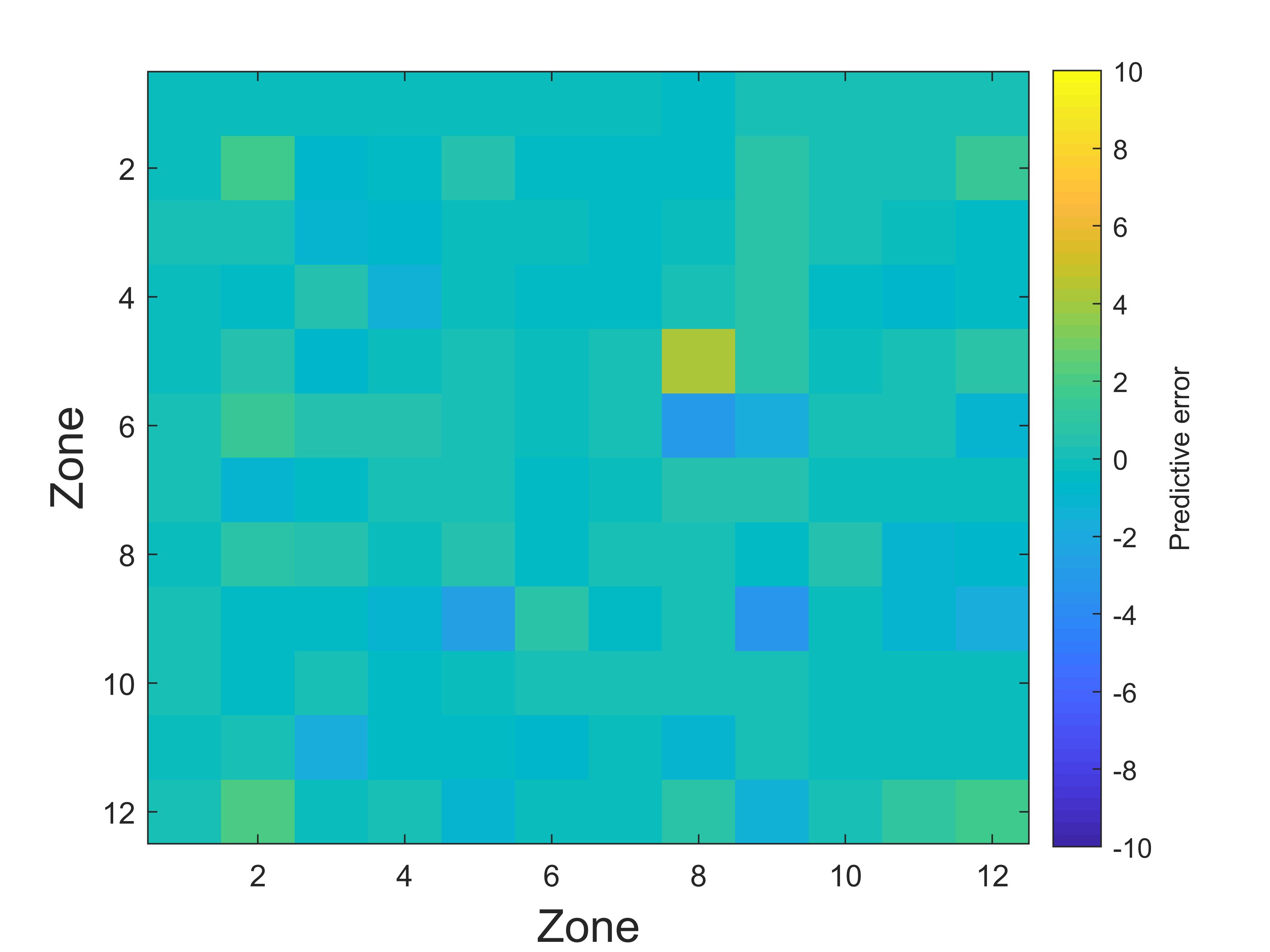}\\

\scriptsize \textbf{h=8}& \scriptsize \textbf{h=10}\\
\end{tabular}
\vspace{-0.2cm}
\caption{The gap between the predictive values obtained by the TCTNN  model and the true values 
at 5:00 PM on March 1 under different forecast horizon on the Abilene data.}\label{fig.Abilene}
\vspace{-0.5cm}
\end{figure}

\begin{figure}[!htbp]
\renewcommand{\arraystretch}{0.5}
\setlength\tabcolsep{0.5pt}
\centering
\begin{tabular}{ccccccc}
\centering

\includegraphics[width=29.3mm, height = 28.3mm]{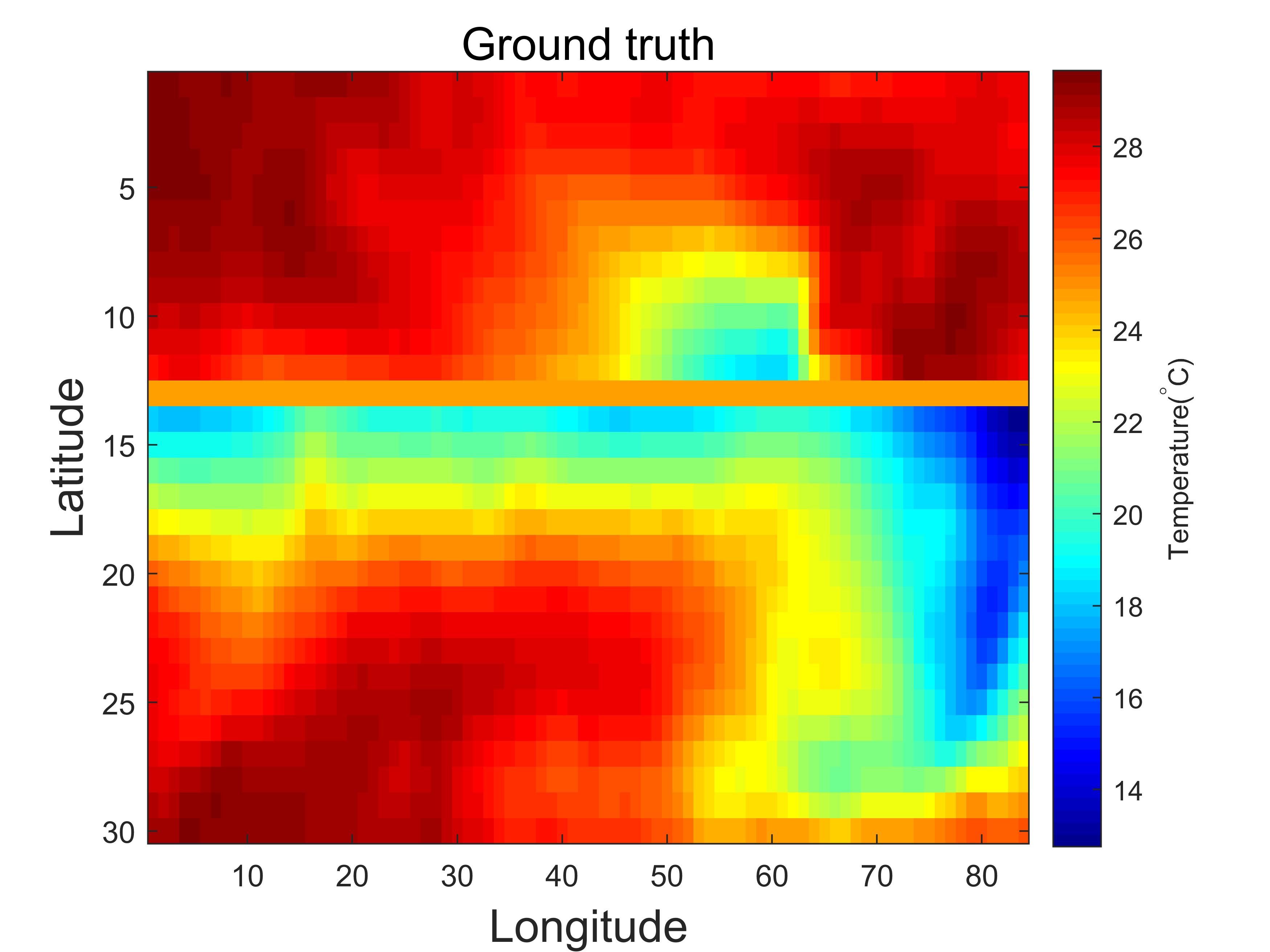}&
\includegraphics[width=29.3mm, height = 28.3mm]{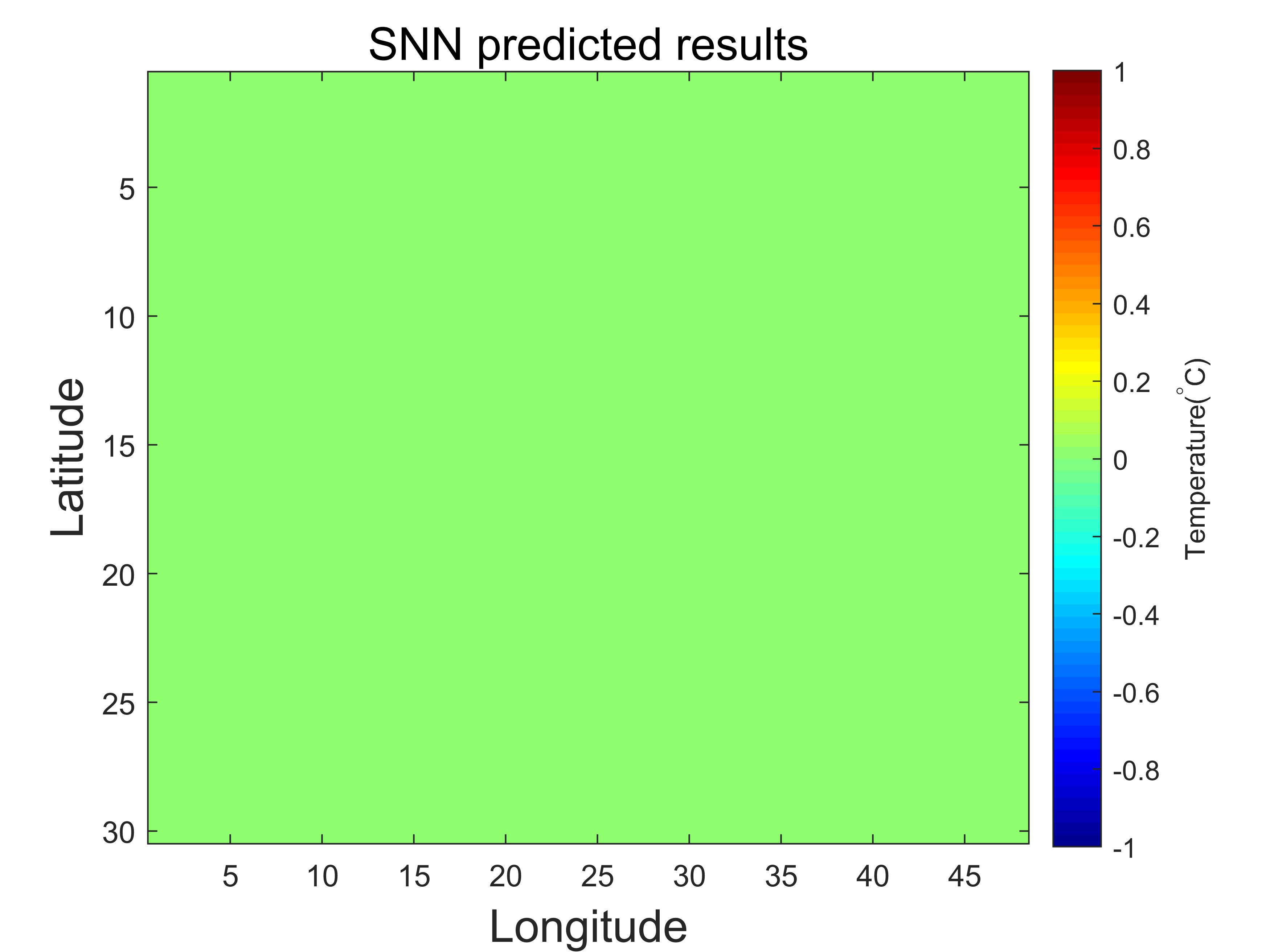}&
\includegraphics[width=29.3mm, height = 28.3mm]{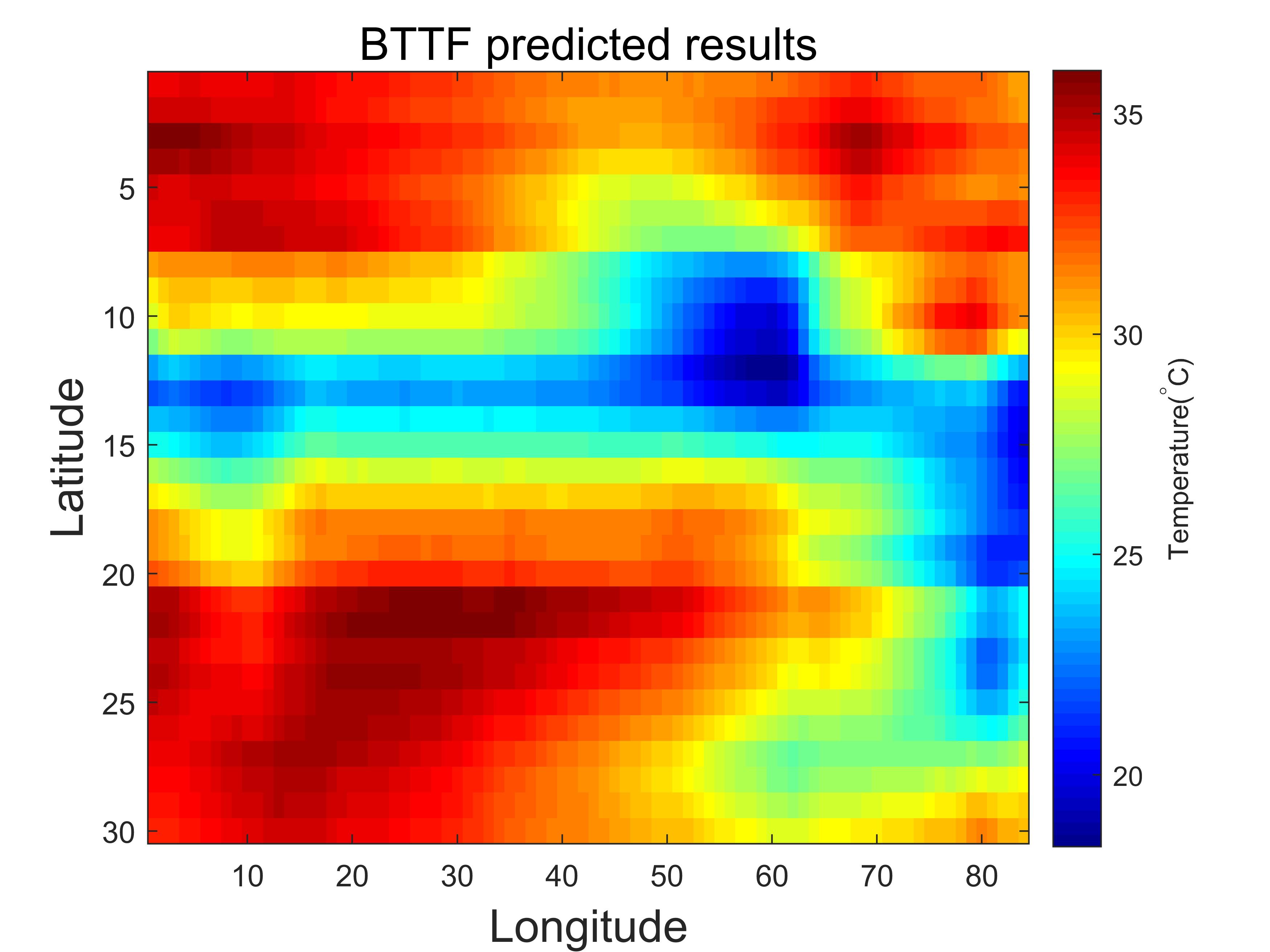}\\

\scriptsize \textbf{Ground truth}& \scriptsize \textbf{SNN\&TNN}  & \scriptsize \textbf{BTTF}\\

\includegraphics[width=29.3mm, height = 28.3mm]{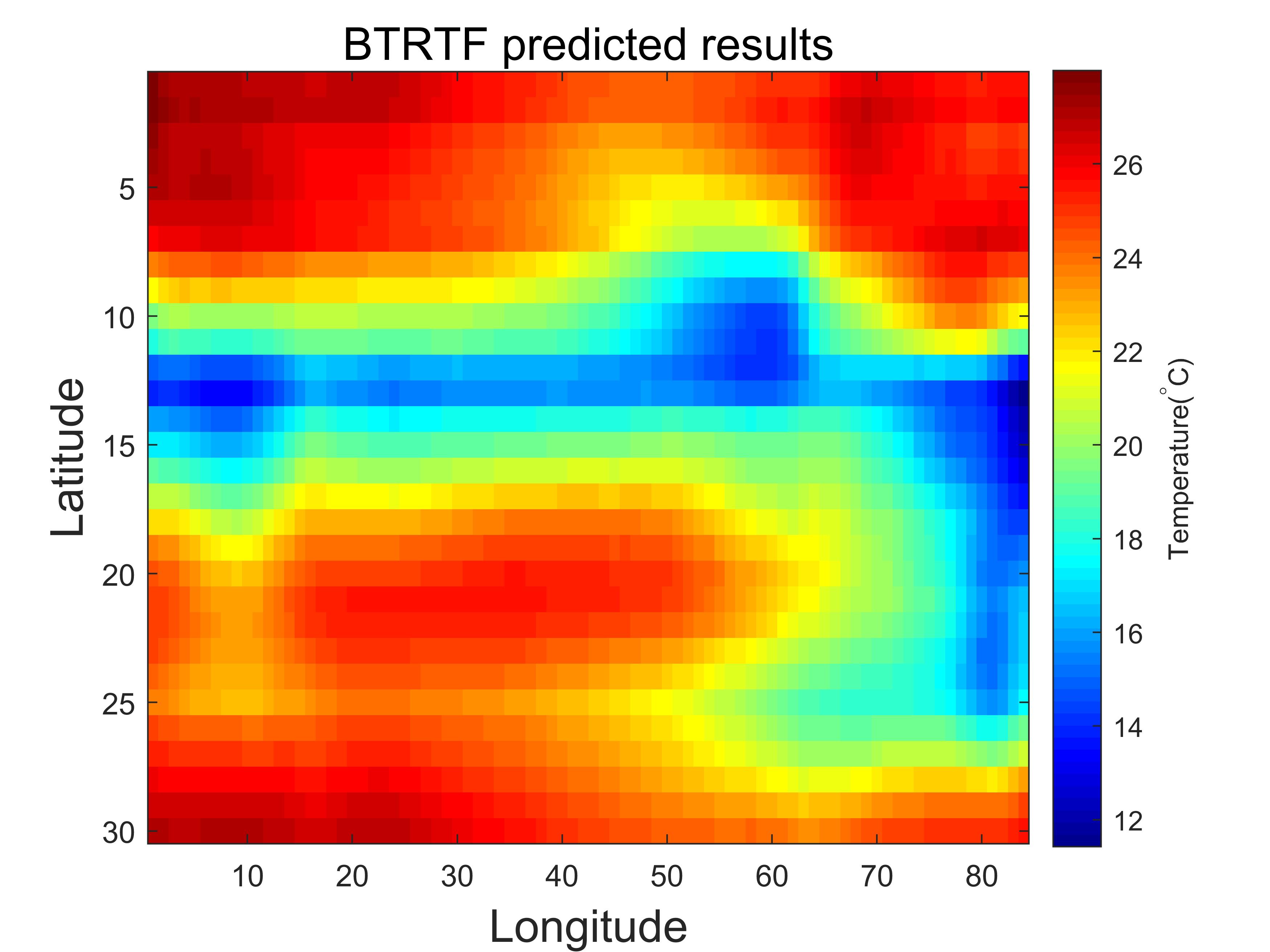}&
\includegraphics[width=29.3mm, height = 28.3mm]{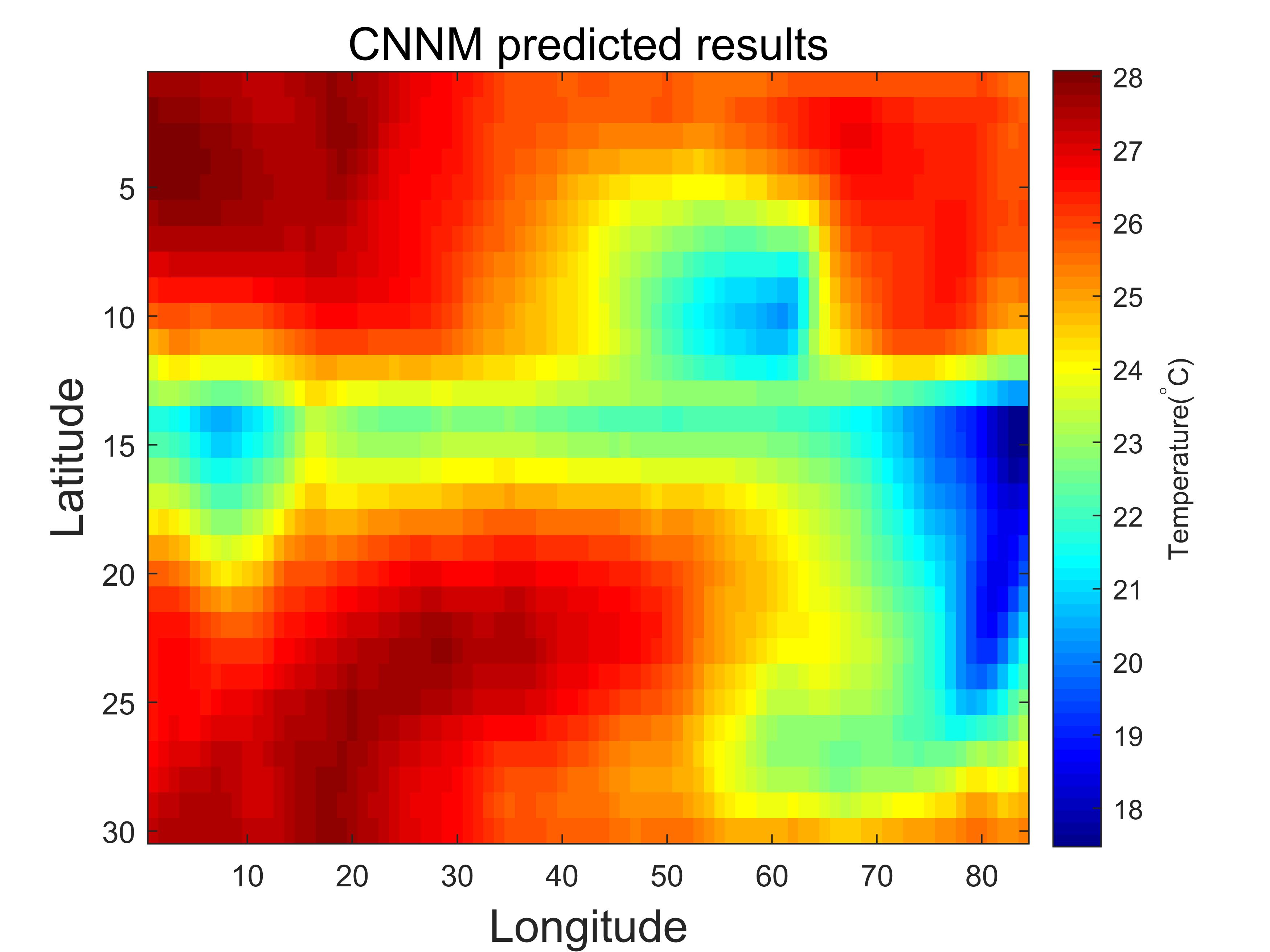}&
\includegraphics[width=29.3mm, height = 28.3mm]{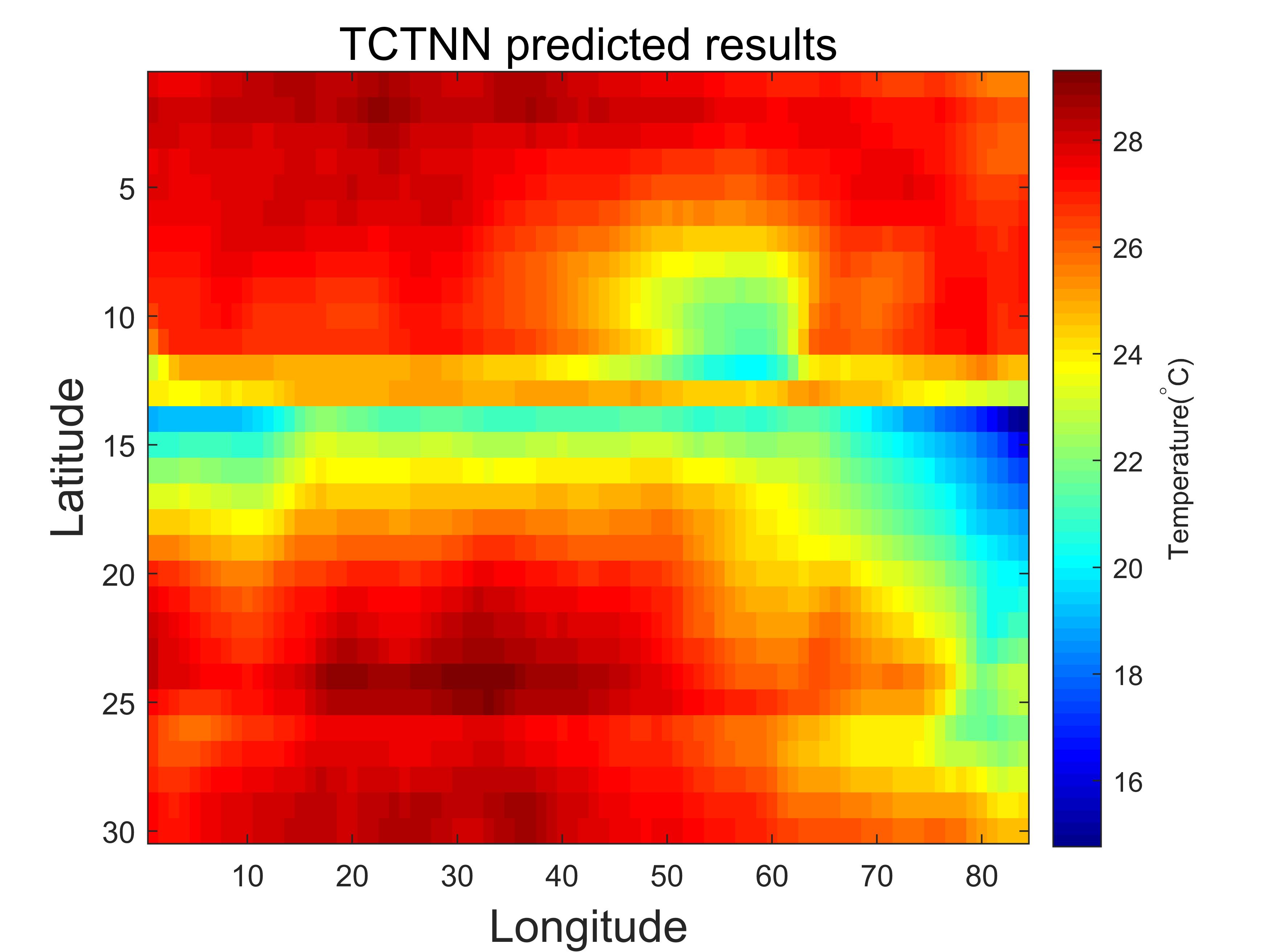}\\

  \scriptsize \textbf{BTRTF}& \scriptsize \textbf{CNNM} &\scriptsize \textbf{TCTNN} \\

\end{tabular}
\caption{The Prediction results of the Pacific surface temperature in September 1973 using the TCTNN model and other tensor based models.}\label{fig.temperature_all_method}
\vspace{-0.7cm}
\end{figure}

\begin{figure*}[!htbp]
\renewcommand{\arraystretch}{0.5}
\setlength\tabcolsep{0.5pt}
\centering
\begin{tabular}{ccccccc}
\centering

\scriptsize \textbf{Sep. 1973}& \scriptsize \textbf{Oct. 1973} & \scriptsize \textbf{Nov. 1973} & \scriptsize \textbf{Dec. 1973} & \scriptsize \textbf{Jan. 1974} & \scriptsize \textbf{Feb. 1974}\\

\includegraphics[width=29.3mm, height = 28.3mm]{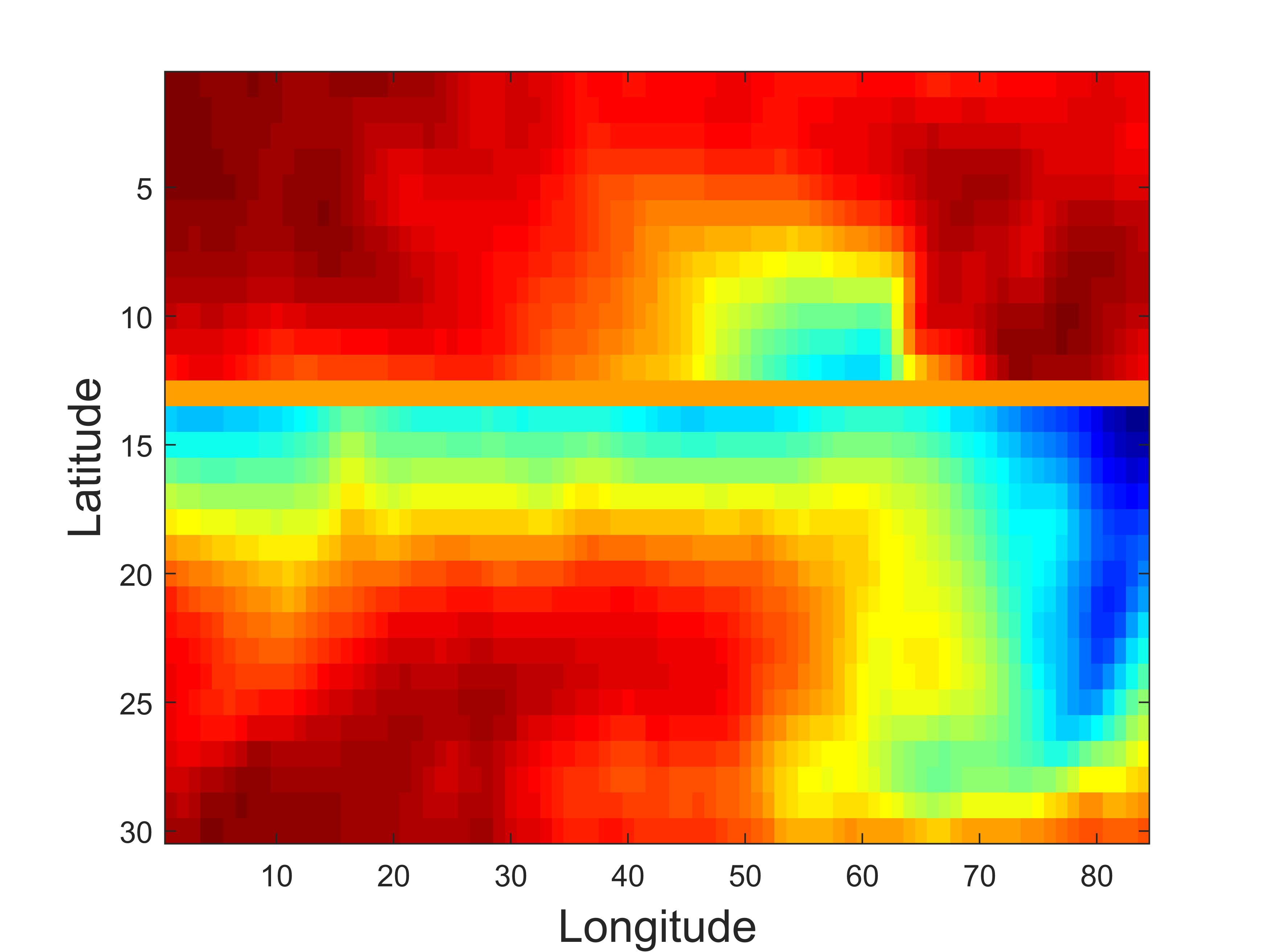}&
\includegraphics[width=29.3mm, height = 28.3mm]{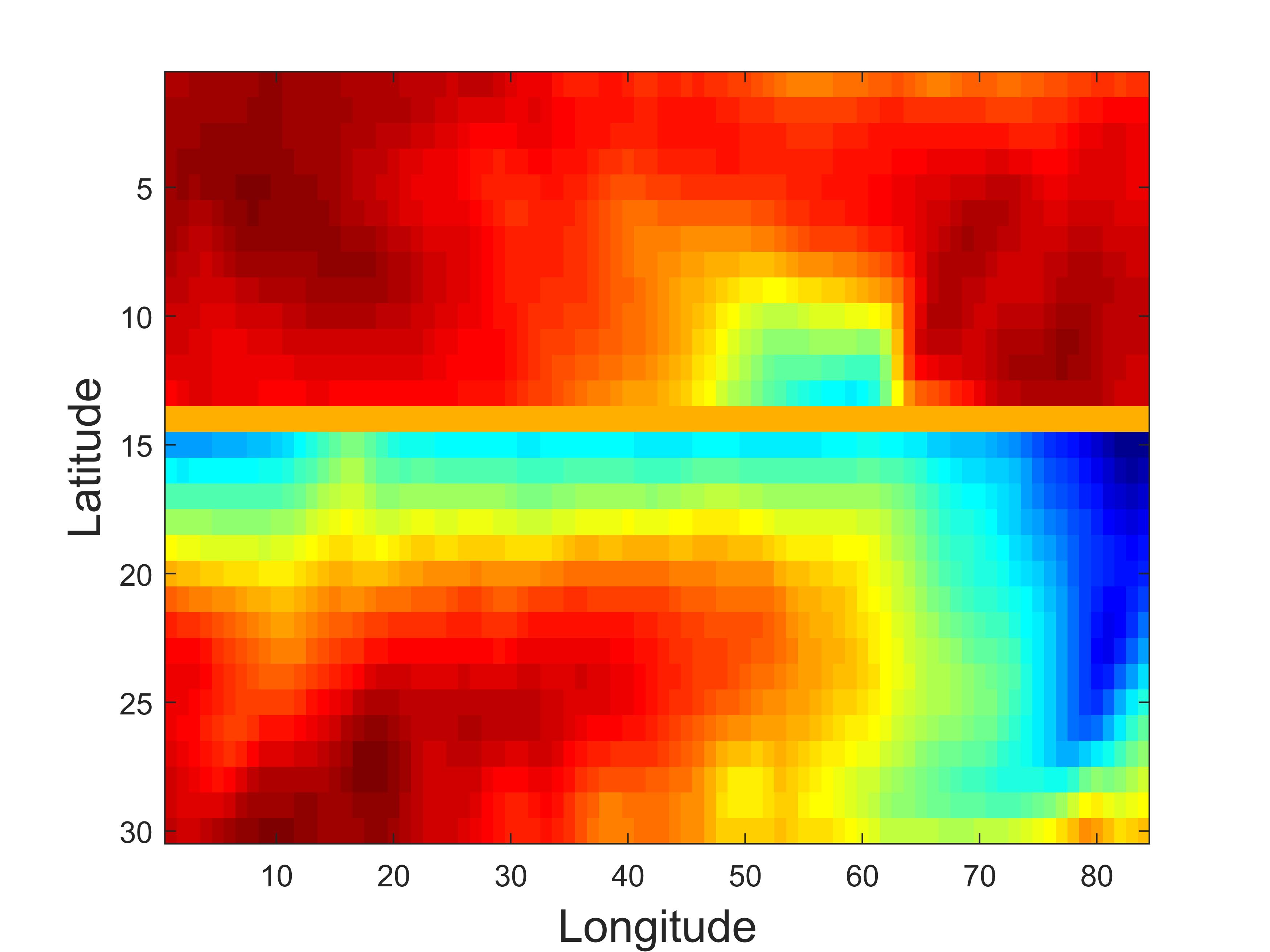}&
\includegraphics[width=29.3mm, height = 28.3mm]{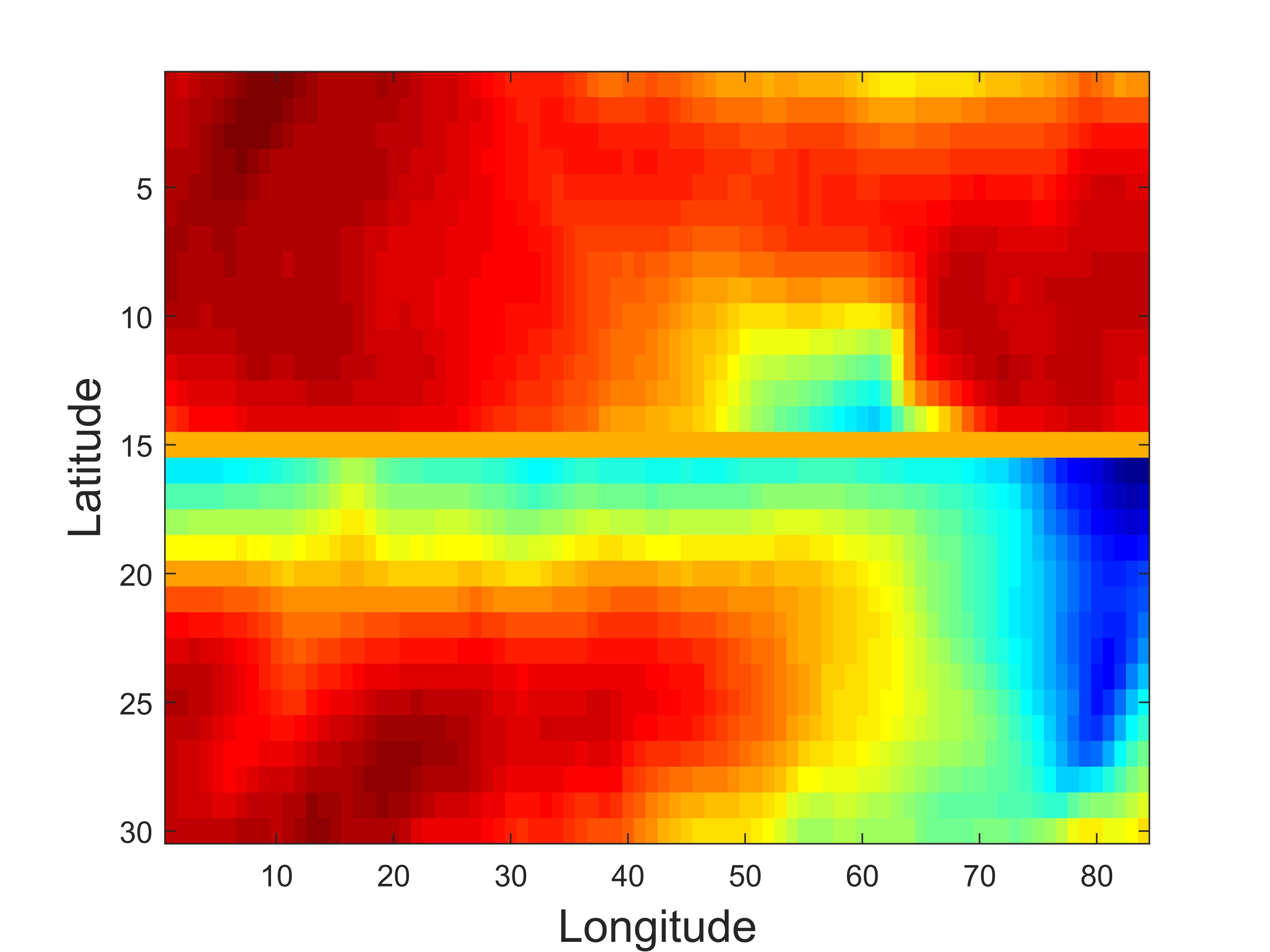}&
\includegraphics[width=29.3mm, height = 28.3mm]{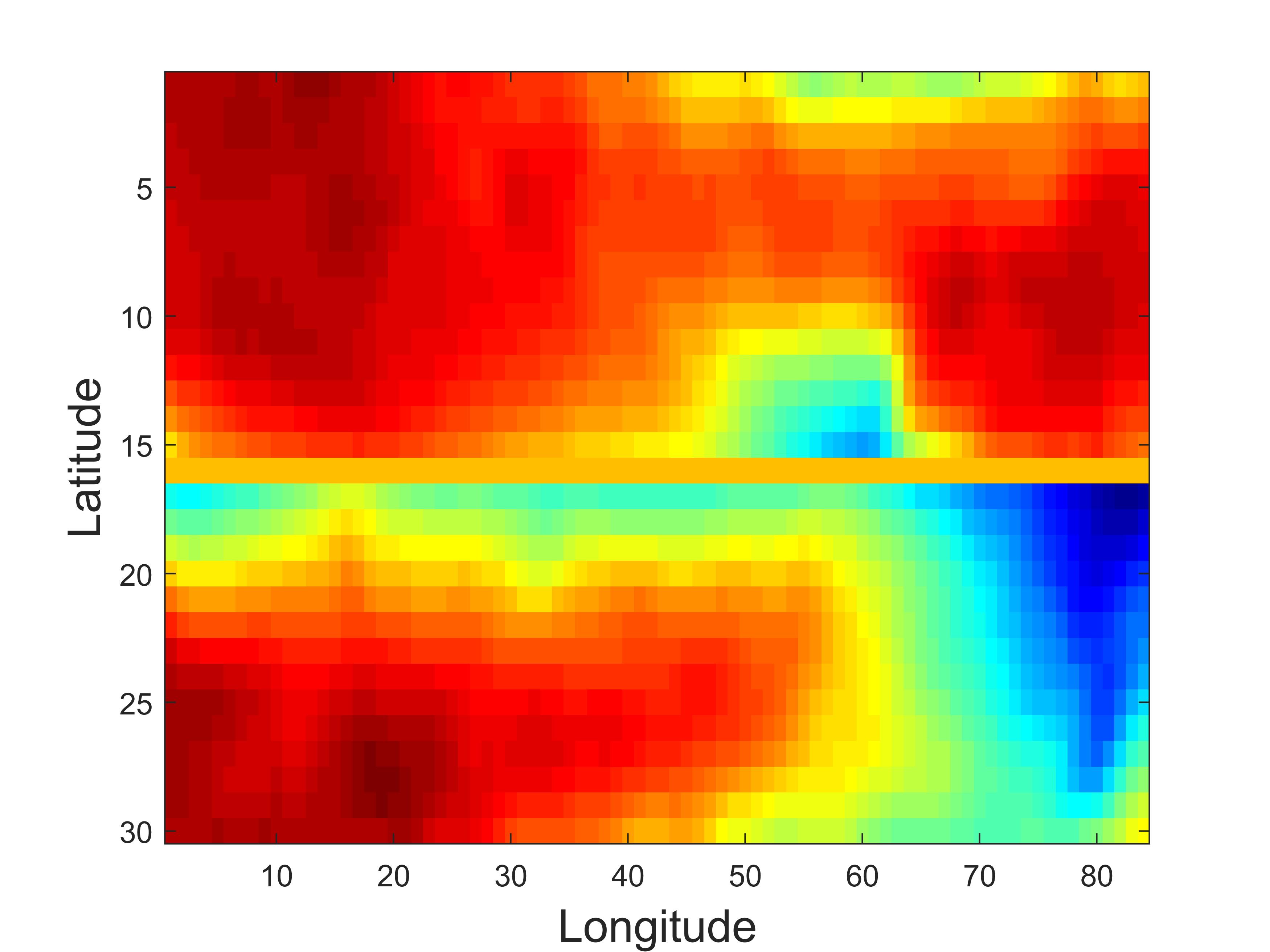}&
\includegraphics[width=29.3mm, height = 28.3mm]{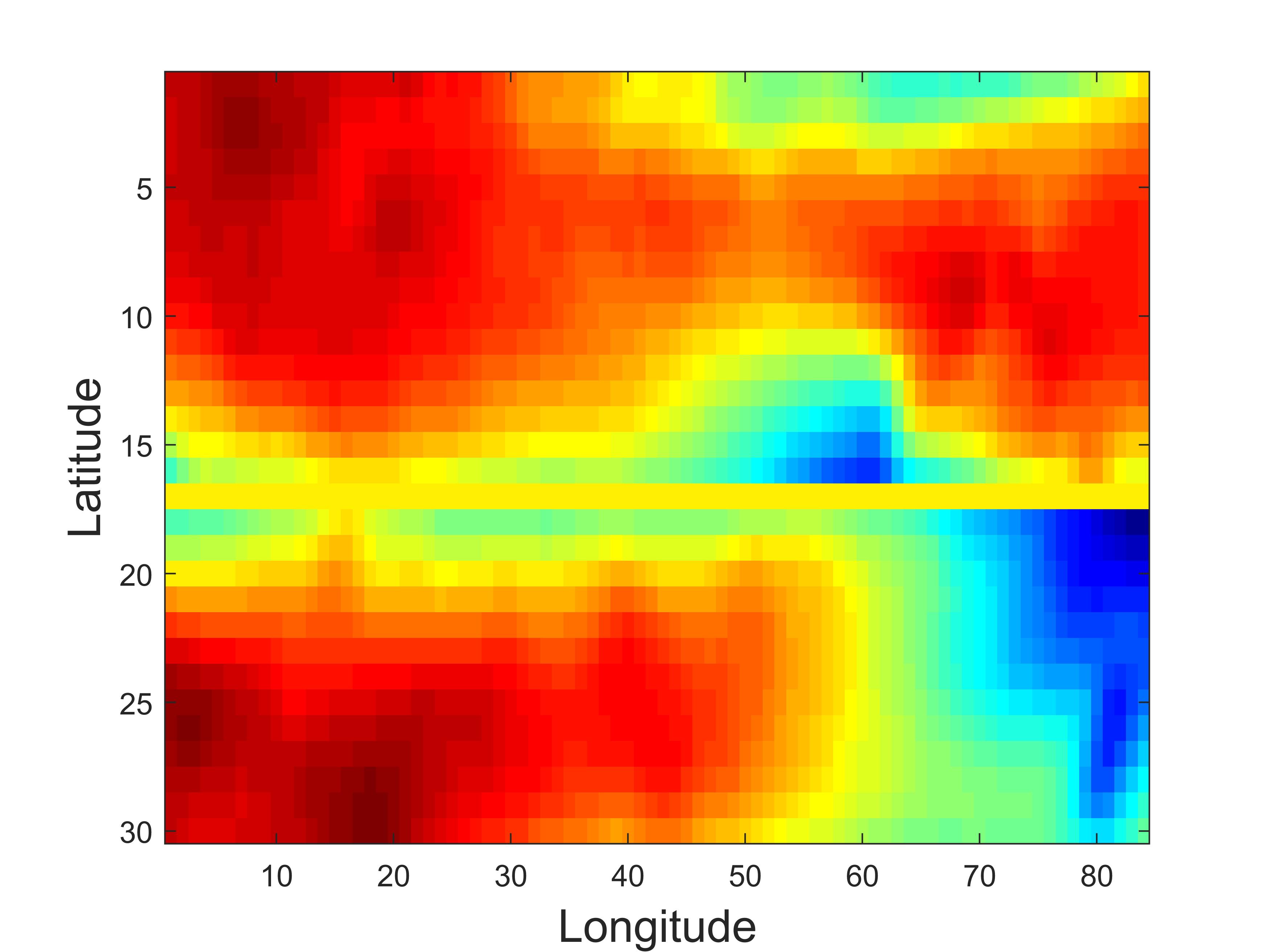}&
\includegraphics[width=29.3mm, height = 28.3mm]{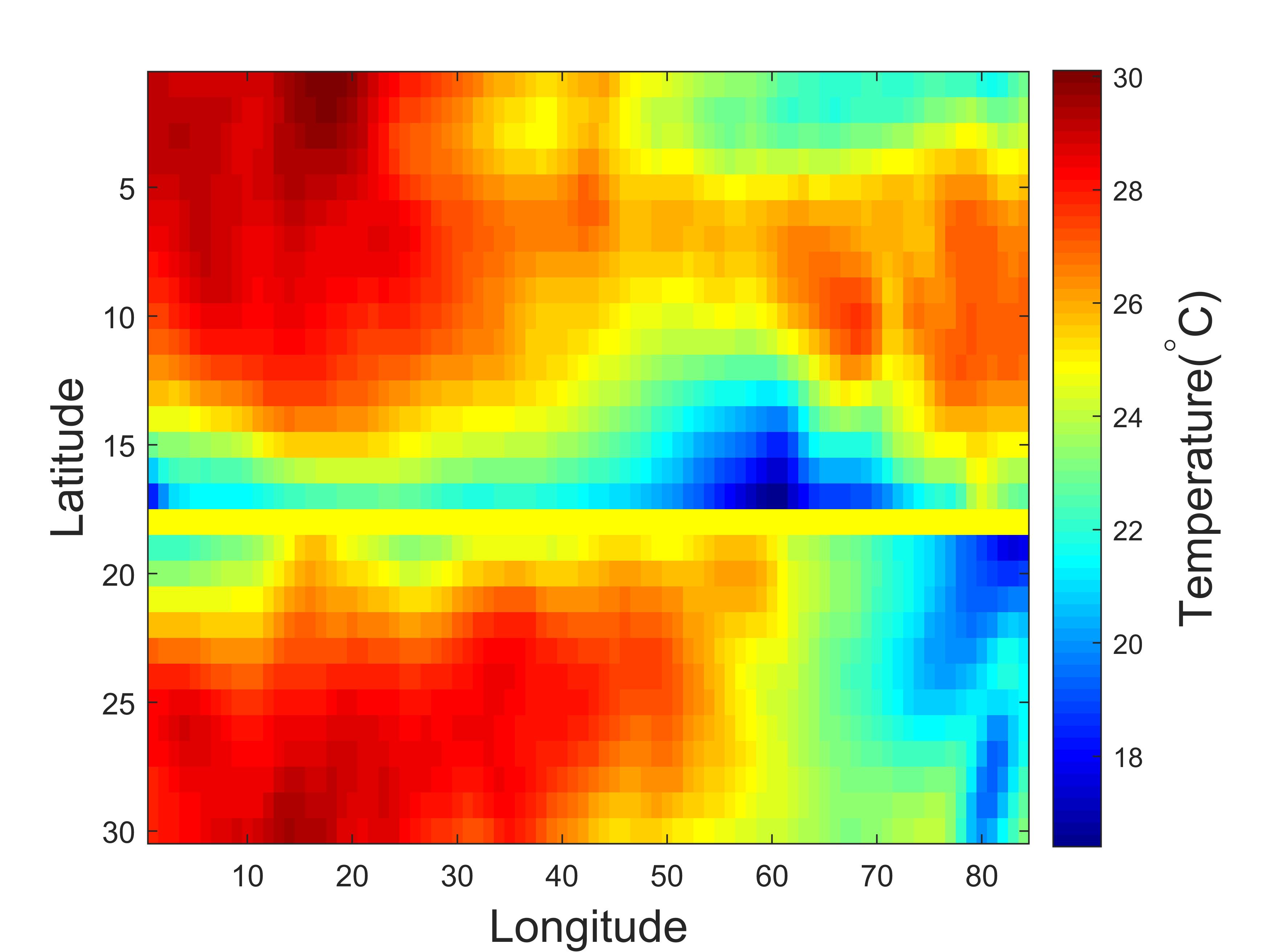}\\

\scriptsize \textbf{Ground truth 1}& \scriptsize \textbf{Ground truth 2} & \scriptsize \textbf{Ground truth 3} & \scriptsize \textbf{Ground truth 4} & \scriptsize \textbf{Ground truth 5} & \scriptsize \textbf{Ground truth 6}\\

\includegraphics[width=29.3mm, height = 28.3mm]{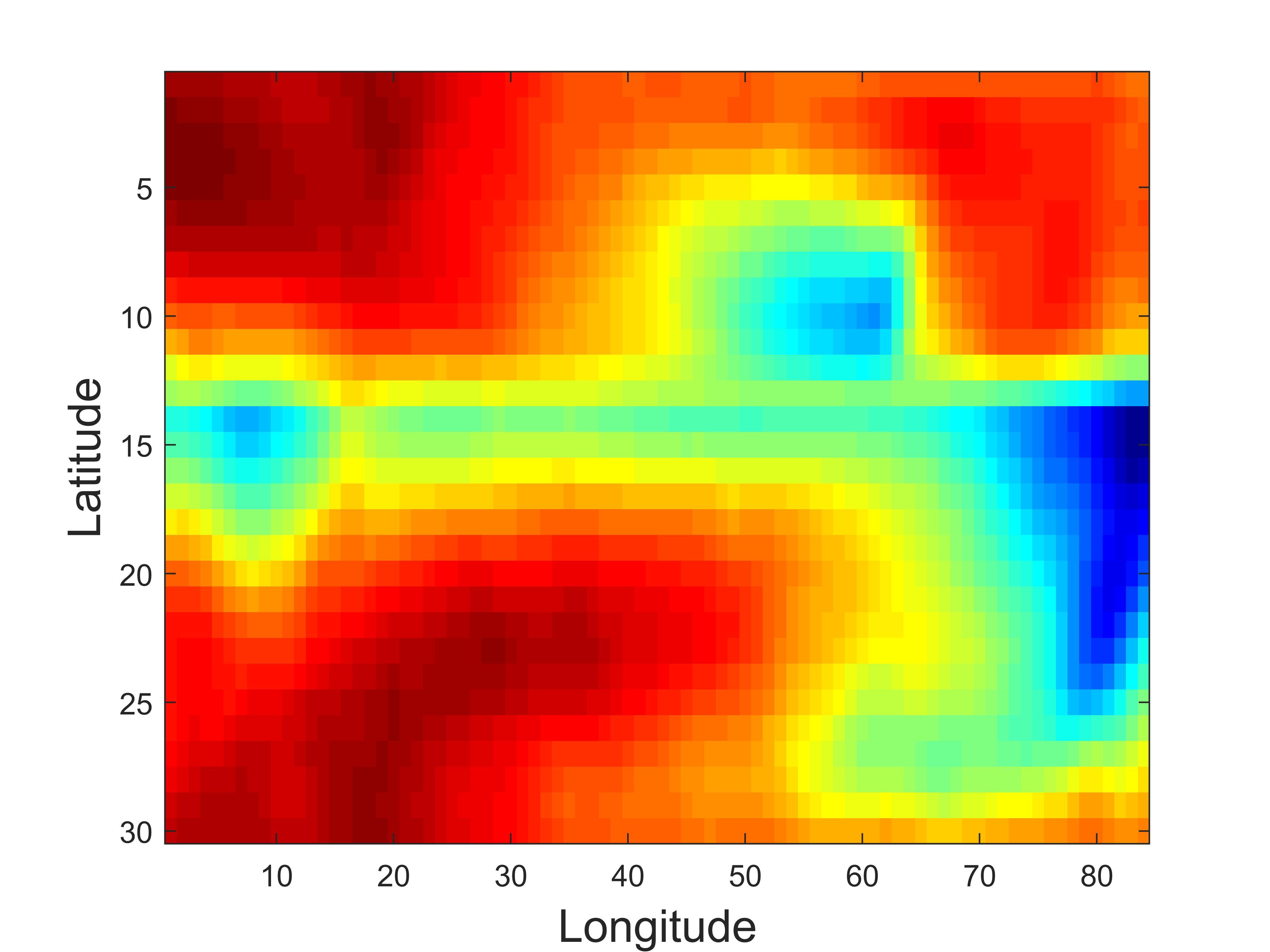}&
\includegraphics[width=29.3mm, height = 28.3mm]{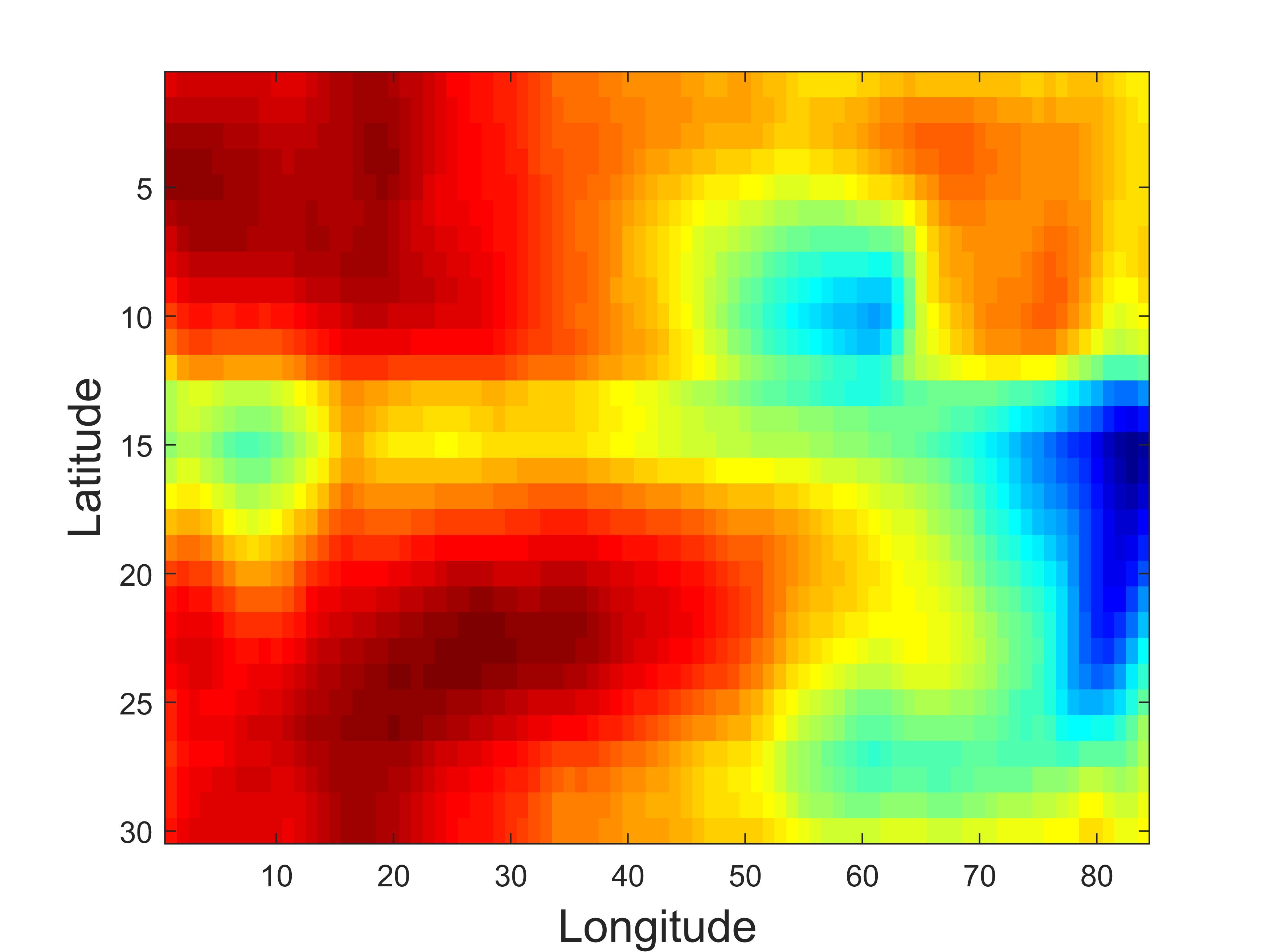}&
\includegraphics[width=29.3mm, height = 28.3mm]{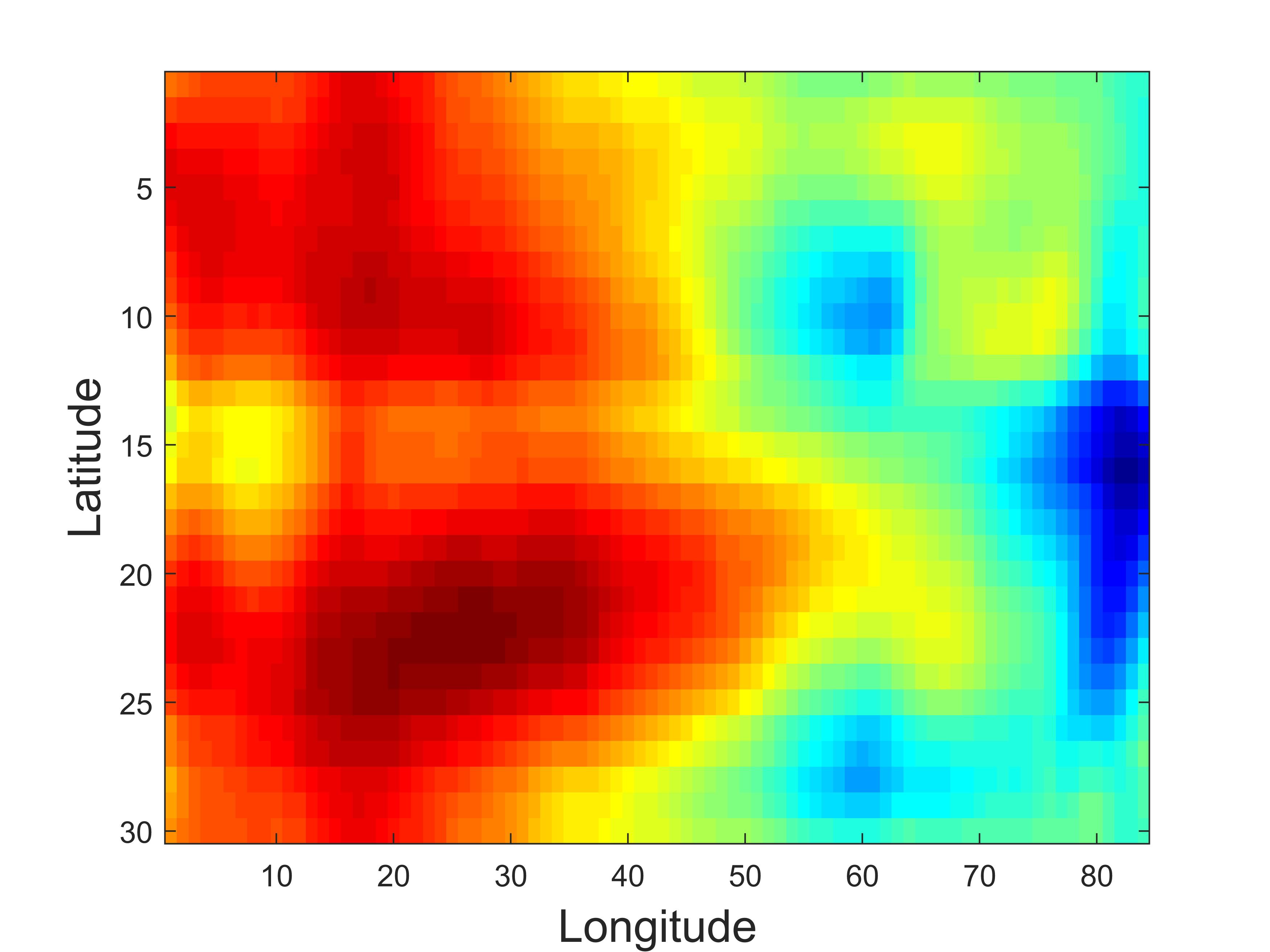}&
\includegraphics[width=29.3mm, height = 28.3mm]{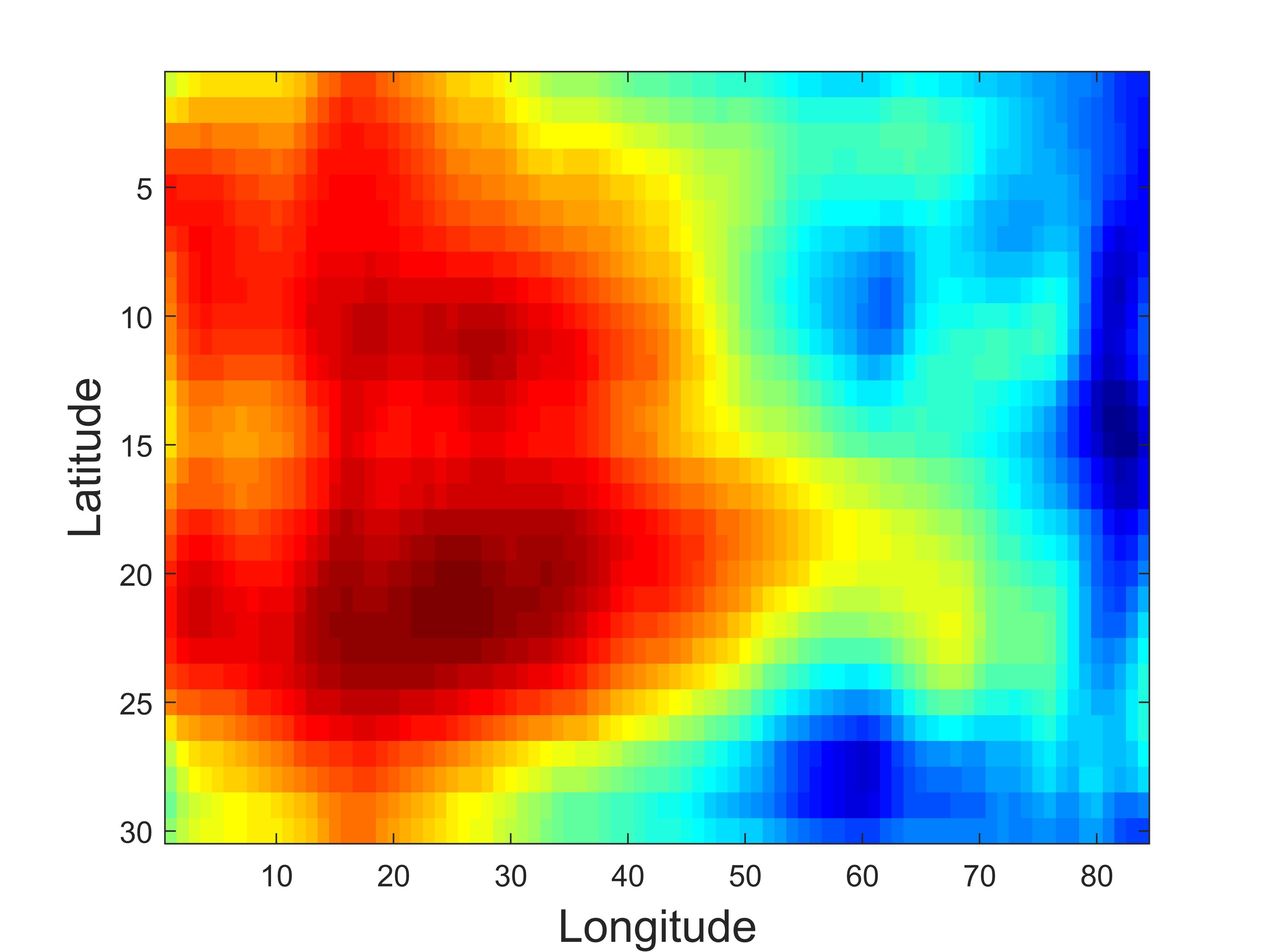}&
\includegraphics[width=29.3mm, height = 28.3mm]{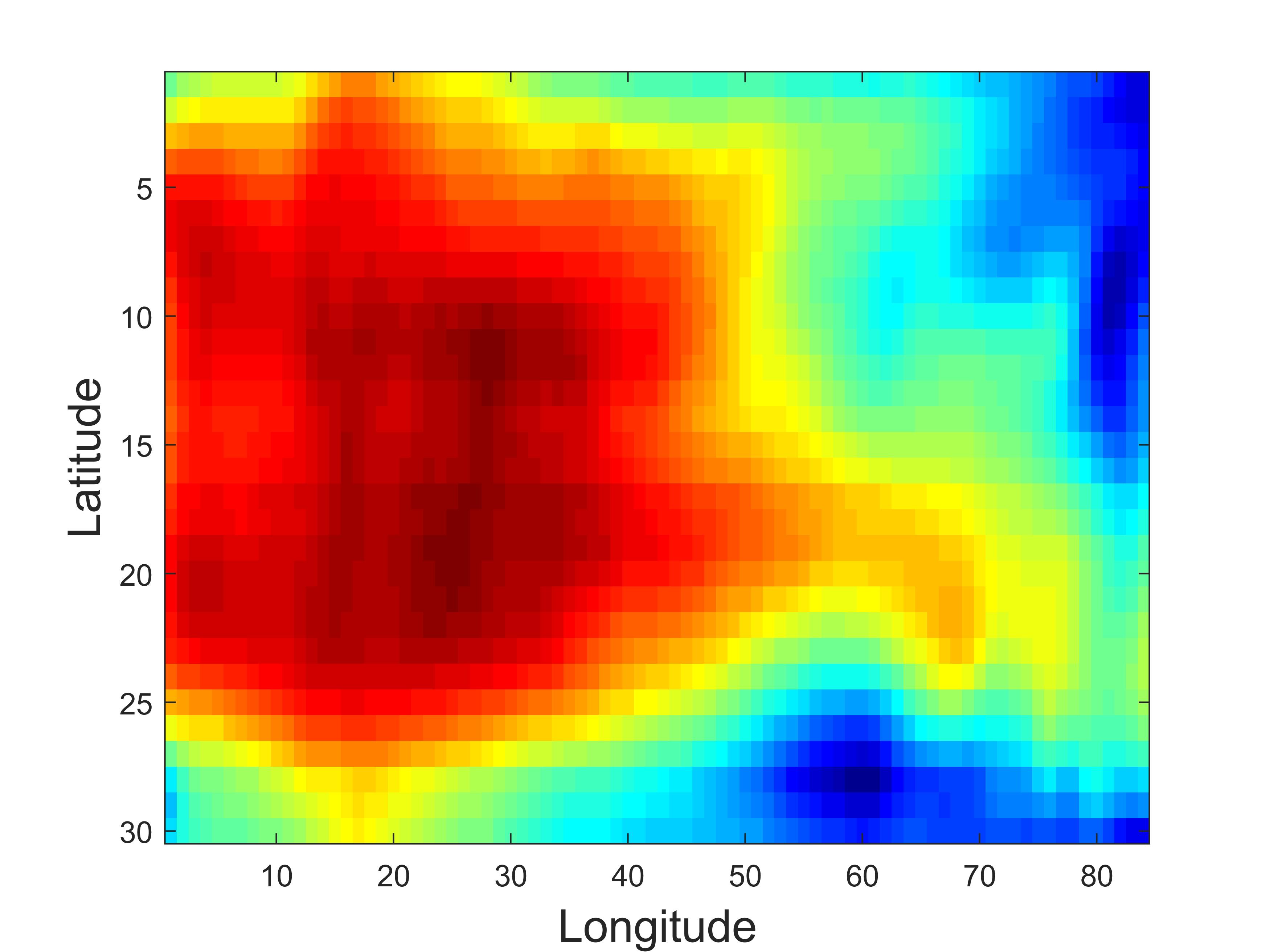}&
\includegraphics[width=29.3mm, height = 28.3mm]{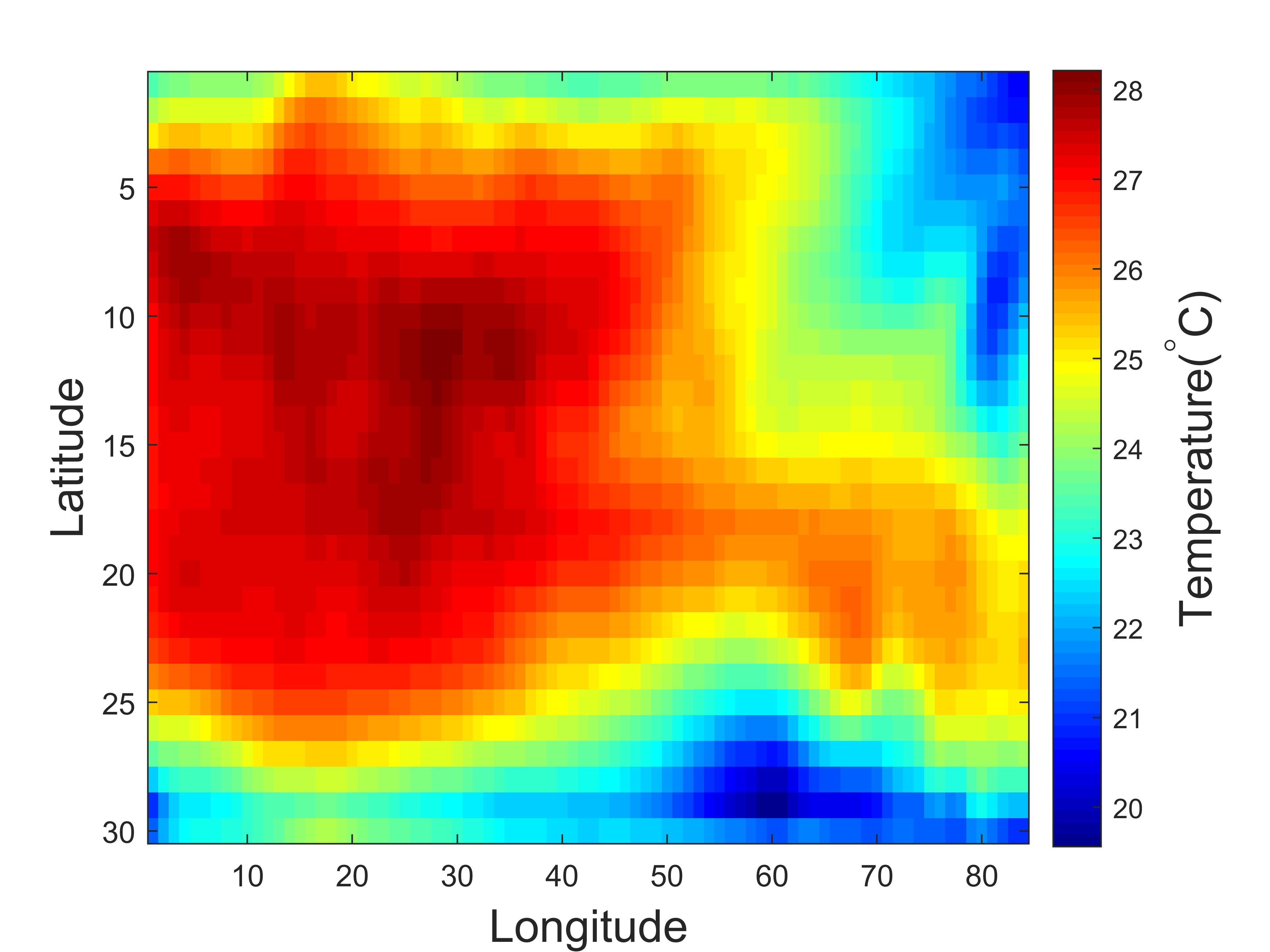}\\

\scriptsize \textbf{CNNM 1}& \scriptsize \textbf{CNNM 2} & \scriptsize \textbf{CNNM 3} & \scriptsize \textbf{CNNM 4} & \scriptsize \textbf{CNNM 5} & \scriptsize \textbf{CNNM 6}\\

\includegraphics[width=29.3mm, height = 28.3mm]{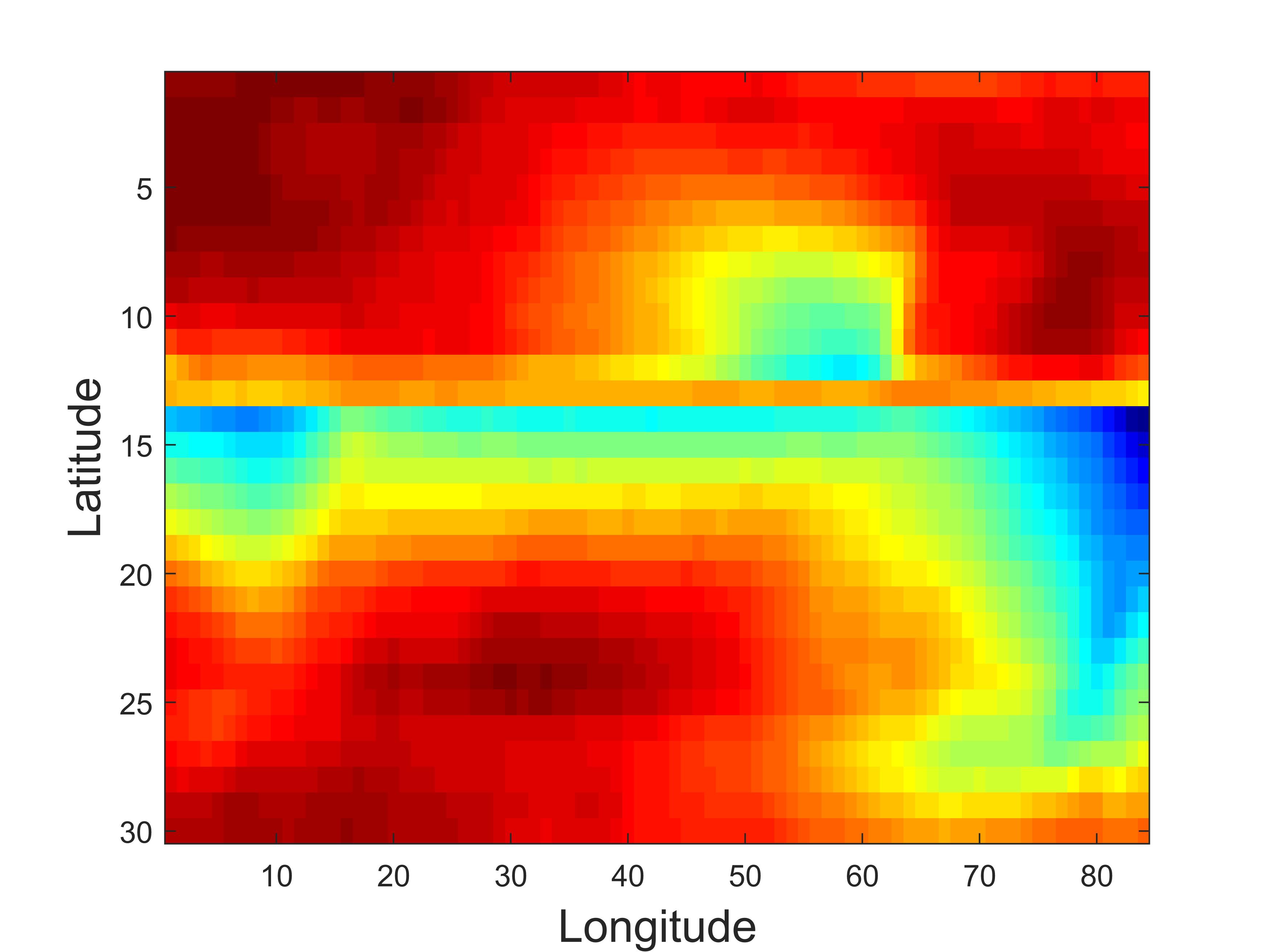}&
\includegraphics[width=29.3mm, height = 28.3mm]{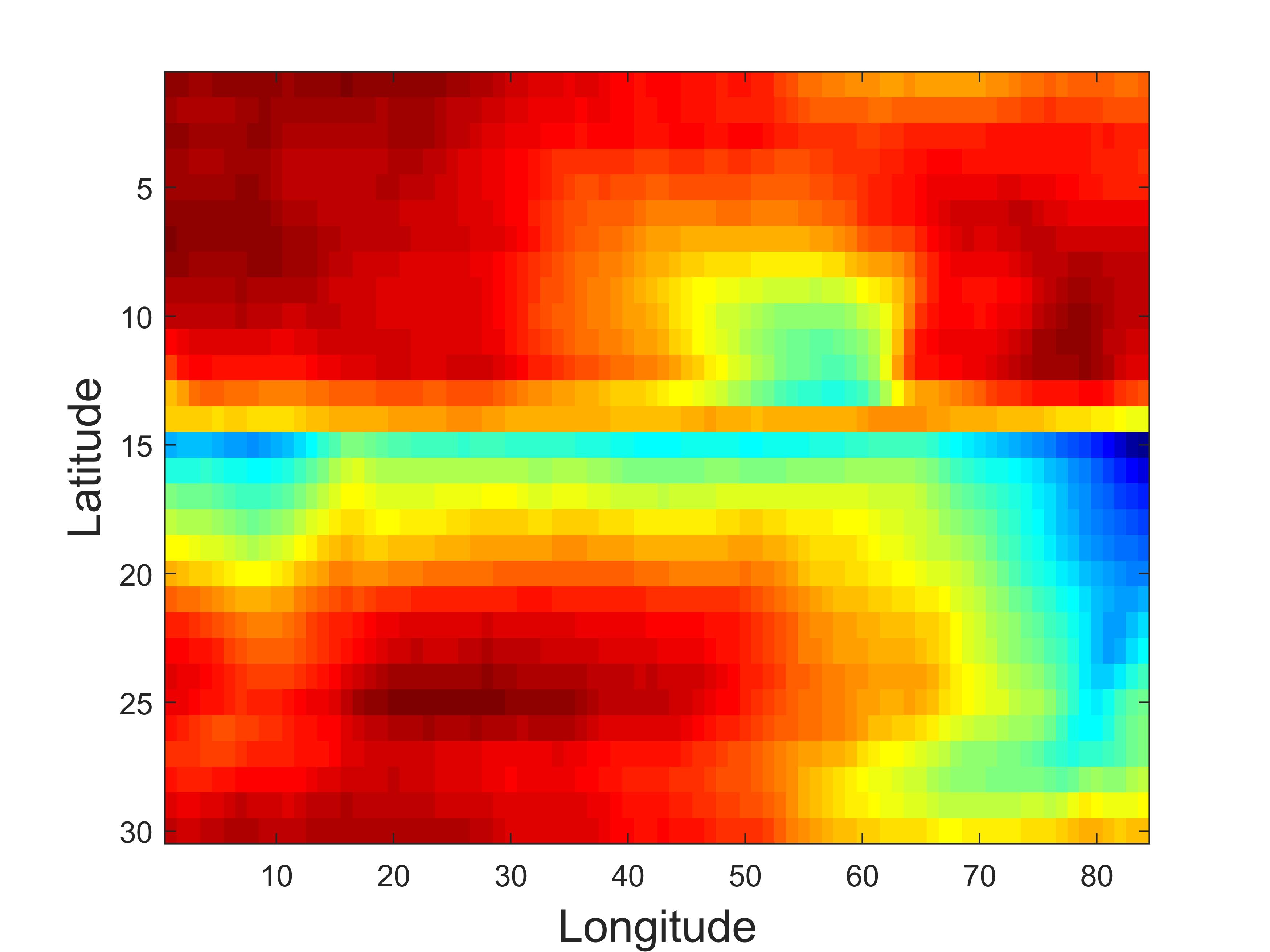}&
\includegraphics[width=29.3mm, height = 28.3mm]{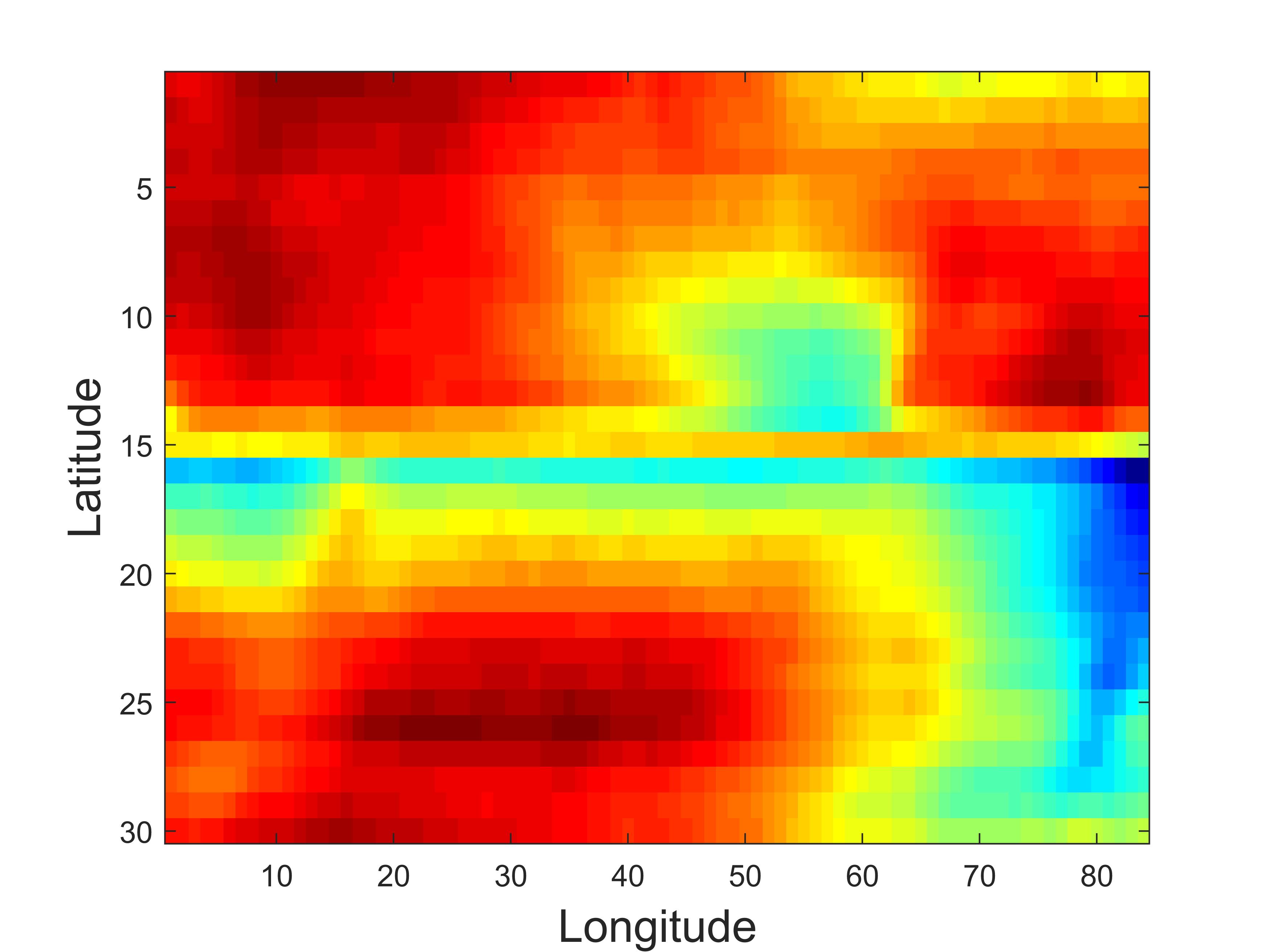}&
\includegraphics[width=29.3mm, height = 28.3mm]{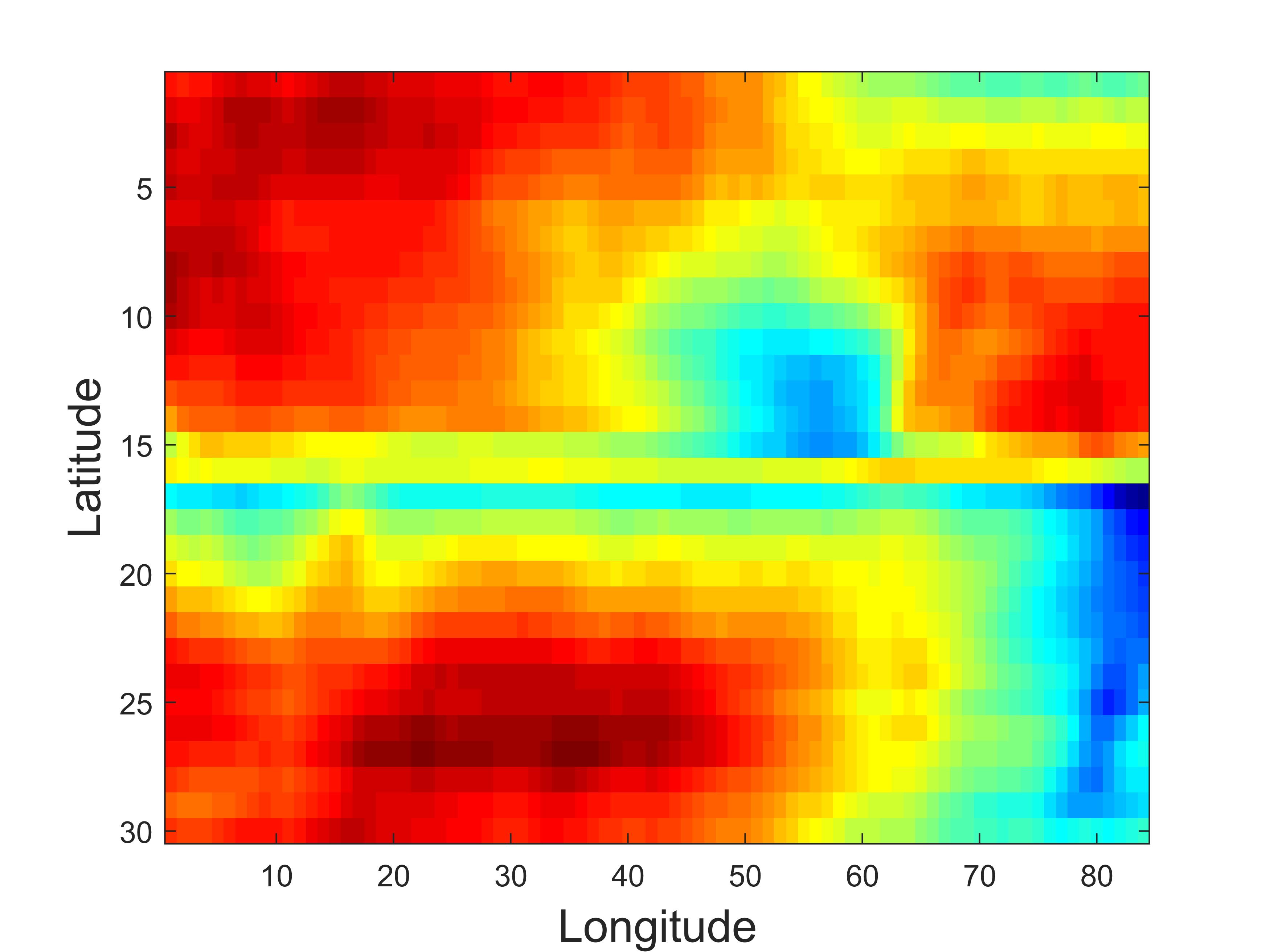}&
\includegraphics[width=29.3mm, height = 28.3mm]{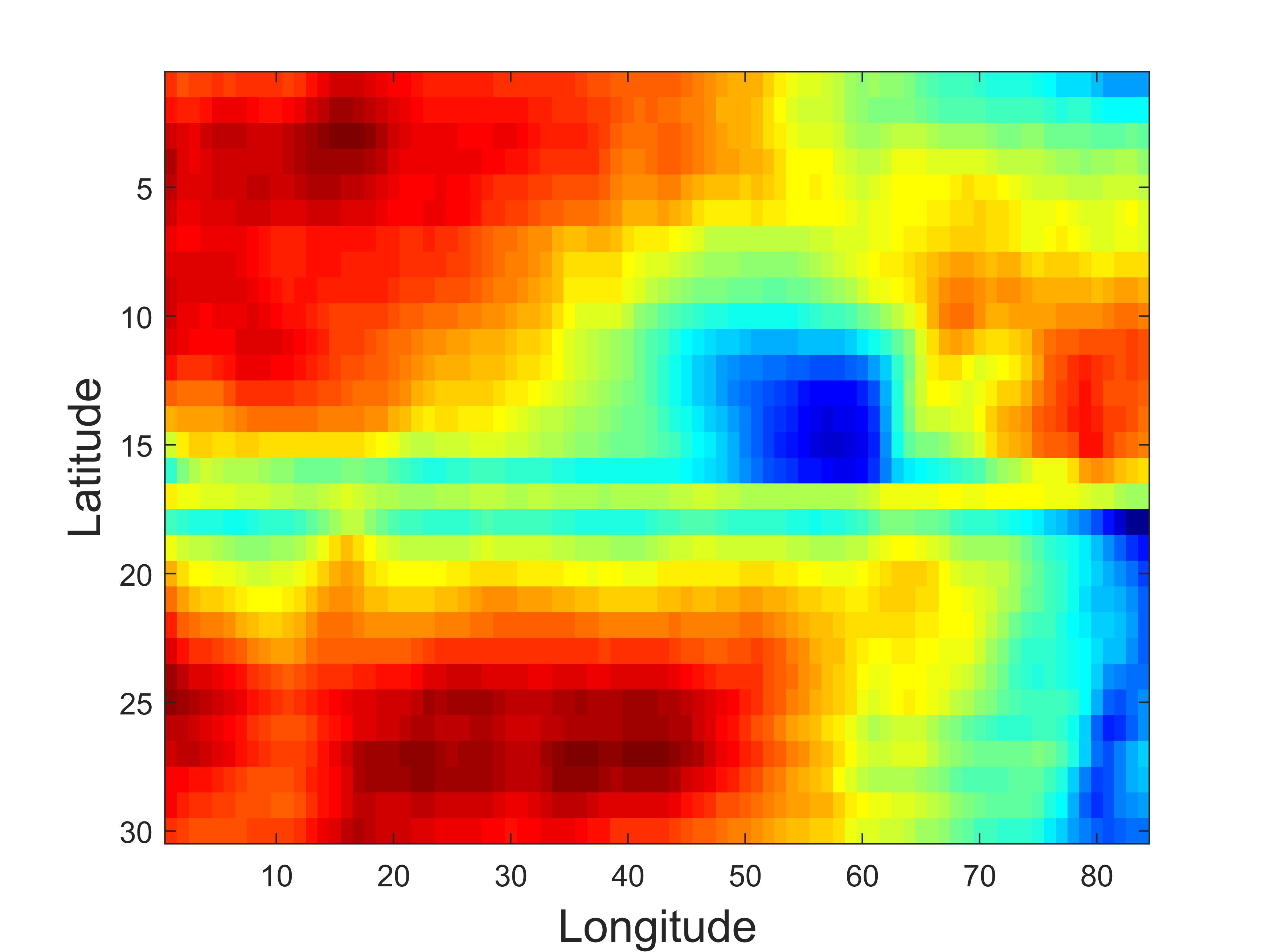}&
\includegraphics[width=29.3mm, height = 28.3mm]{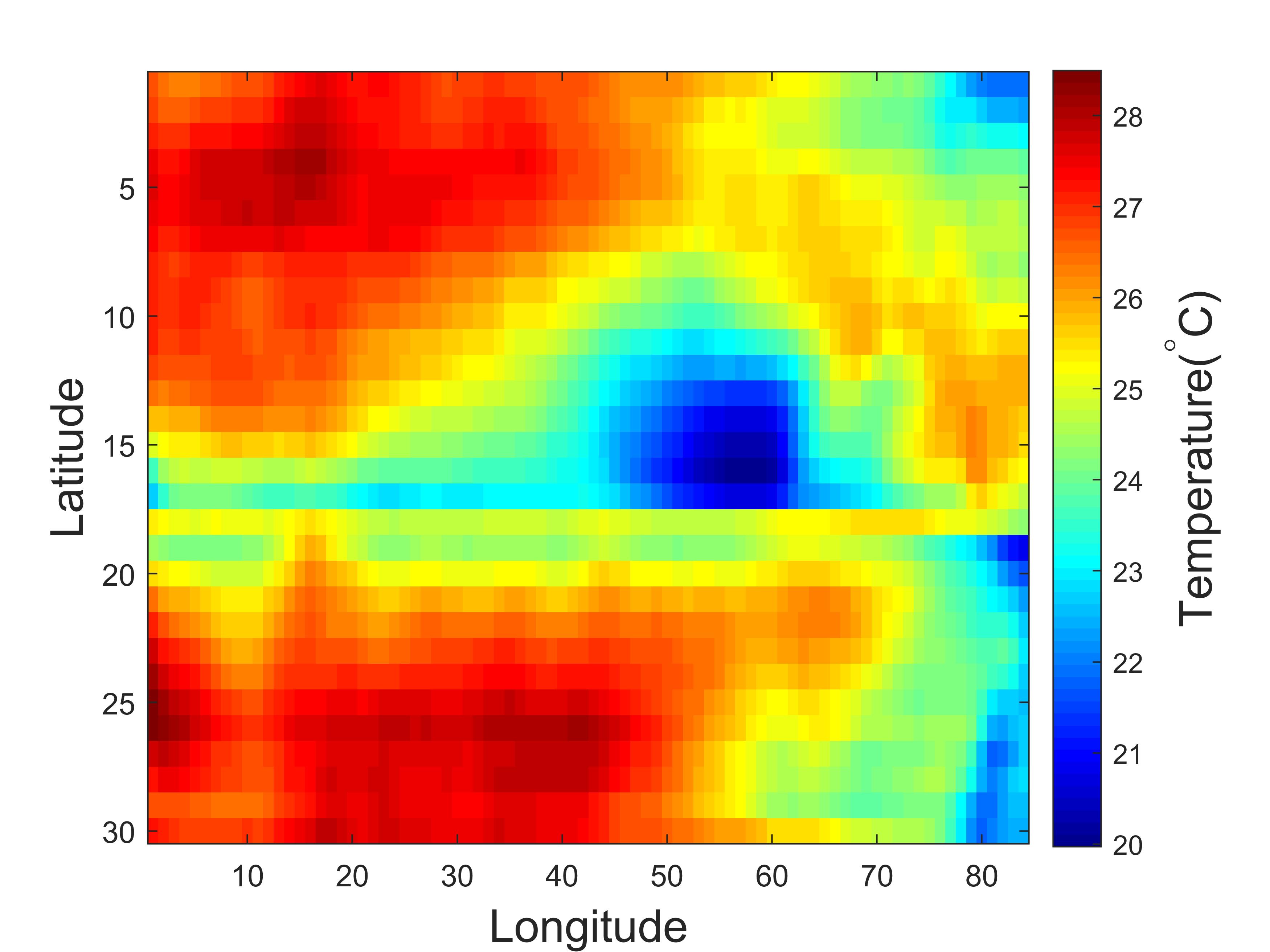}\\

\scriptsize \textbf{TCTNN 1}& \scriptsize \textbf{TCTNN 2} & \scriptsize \textbf{TCTNN 3} & \scriptsize \textbf{TCTNN 4} & \scriptsize \textbf{TCTNN 5} & \scriptsize \textbf{TCTNN 6}\\
\end{tabular}
\vspace{-0.2cm}
\caption{The predictive results of the TCTNN model and the CNNM 
model on the  Pacific surface temperature dataset.}\label{fig.temperature}
\vspace{-0.5cm}
\end{figure*}

To intuitively evaluate the prediction data obtained by the proposed  TCTNN method, we use the NYC taxi traffic dataset for testing and take the forecast horizon as 4, then plot the comparison of the prediction results and the true value in Fig.\ref{fig.NYTAXI}. We can see that our prediction results are very consistent with the actual situation. Besides, we calculate the difference between the predicted value and the true value under different  forecast horizon on the Abilene data, and the results are shown in Fig.\ref{fig.Abilene}. 
The figure illustrates that the difference between the predicted values and the true values across each prediction domain is minimal, with most differences approaching zero.
In order to further demonstrate the advantages of the proposed TCTNN method over other baseline methods, we test these models on the the Pacific temperature dataset with a prediction domain of 6 and obtain the prediction results of each method, as shown in Figure \ref{fig.temperature_all_method}.
We observe that the prediction results of SNN and TNN are all 0, the excellent prediction performance of CNNM and TCTNN shows the advantages of convolution nulclear norm type methods in small-sample time series prediction.
To ulteriorly  clarify that our proposed TCTNN method has a stronger ability to capture temporal patterns in multidimensional time series compared to the CNNM method, we apply the TCTNN  model and the CNNM model to the Pacific temperature dataset for predictions at h=6. The prediction results are displayed alongside the Ground truth over time in Figure \ref{fig.temperature}. It is unequivocal that while CNNM's predictions closely align with the Ground truth during the first three months, its performance deteriorates significantly in the following three months compared to TCTNN. This effectively demonstrates that TCTNN possesses a stronger capability for capturing the characteristics of temporal data.

\subsection{Discussions}
\subsubsection{Convergence testing}
\begin{figure}[!htbp]
\centering
\vspace{-0.1cm}
\includegraphics[width=0.5\linewidth]{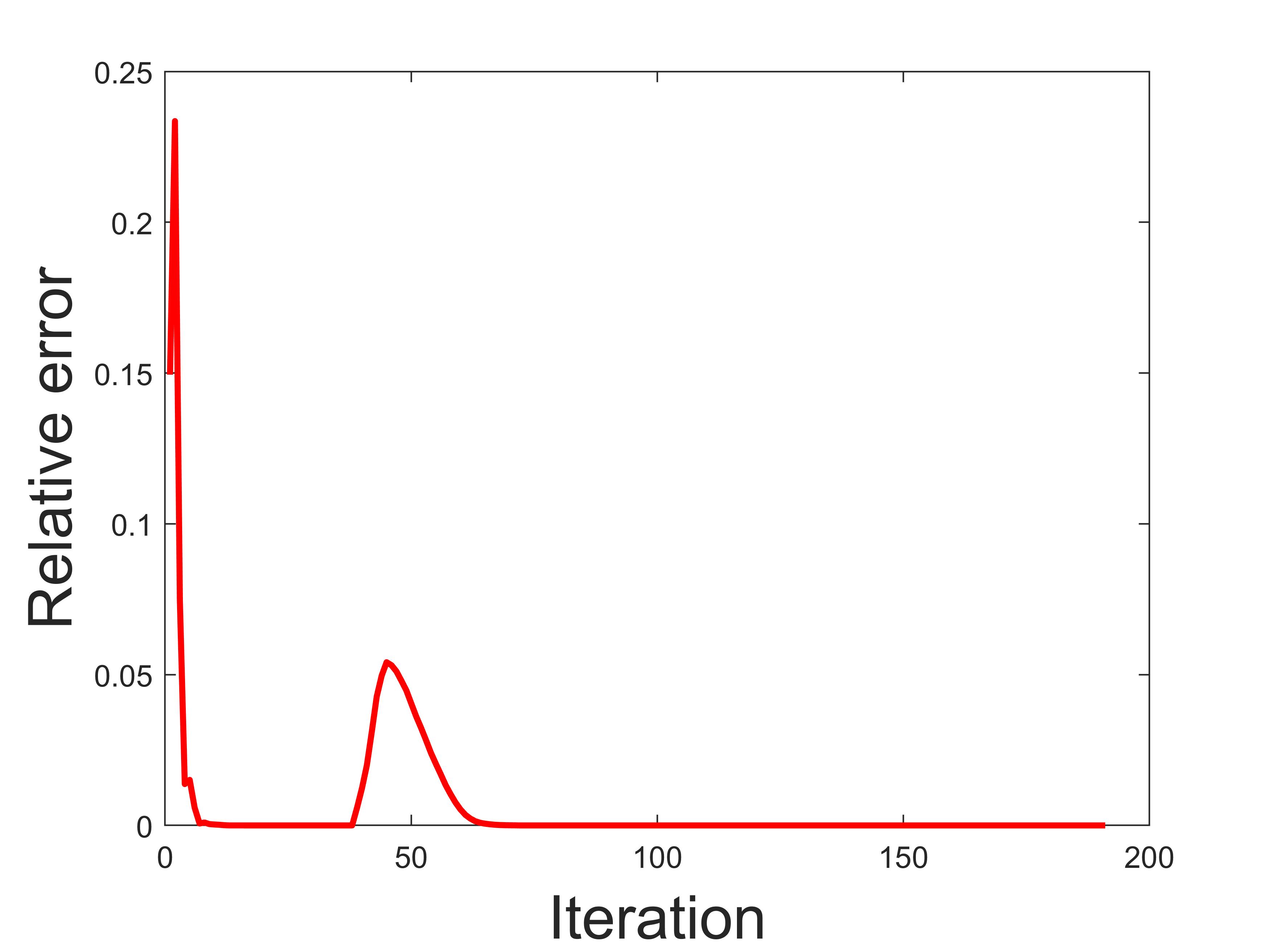}
\vspace{-0.2cm}
\caption{The convergence curves of Algorithm 1. }\label{fig:RelErr}
\vspace{-0.2cm}
\end{figure}

To verify the convergence of our proposed Algorithm 1, we employ   a randomly generated tensor of size 50×50×50 for testing. We calculate the relative error at each iteration step using the formula 
 $\|{\X_{k+1}}-{\X_k}\|_{F}/\|{\X_k}\|_{F}$, and present the corresponding error plots in Figure \ref{fig:RelErr}. From the error curves, it is obvious that the tensor changes minimally after 100 iterations, demonstrating the convergence of the algorithm.

\subsubsection{Kernel size selection}
\begin{figure}[!htbp]
\centering
\vspace{-0.1cm}
\includegraphics[width=0.88\linewidth]{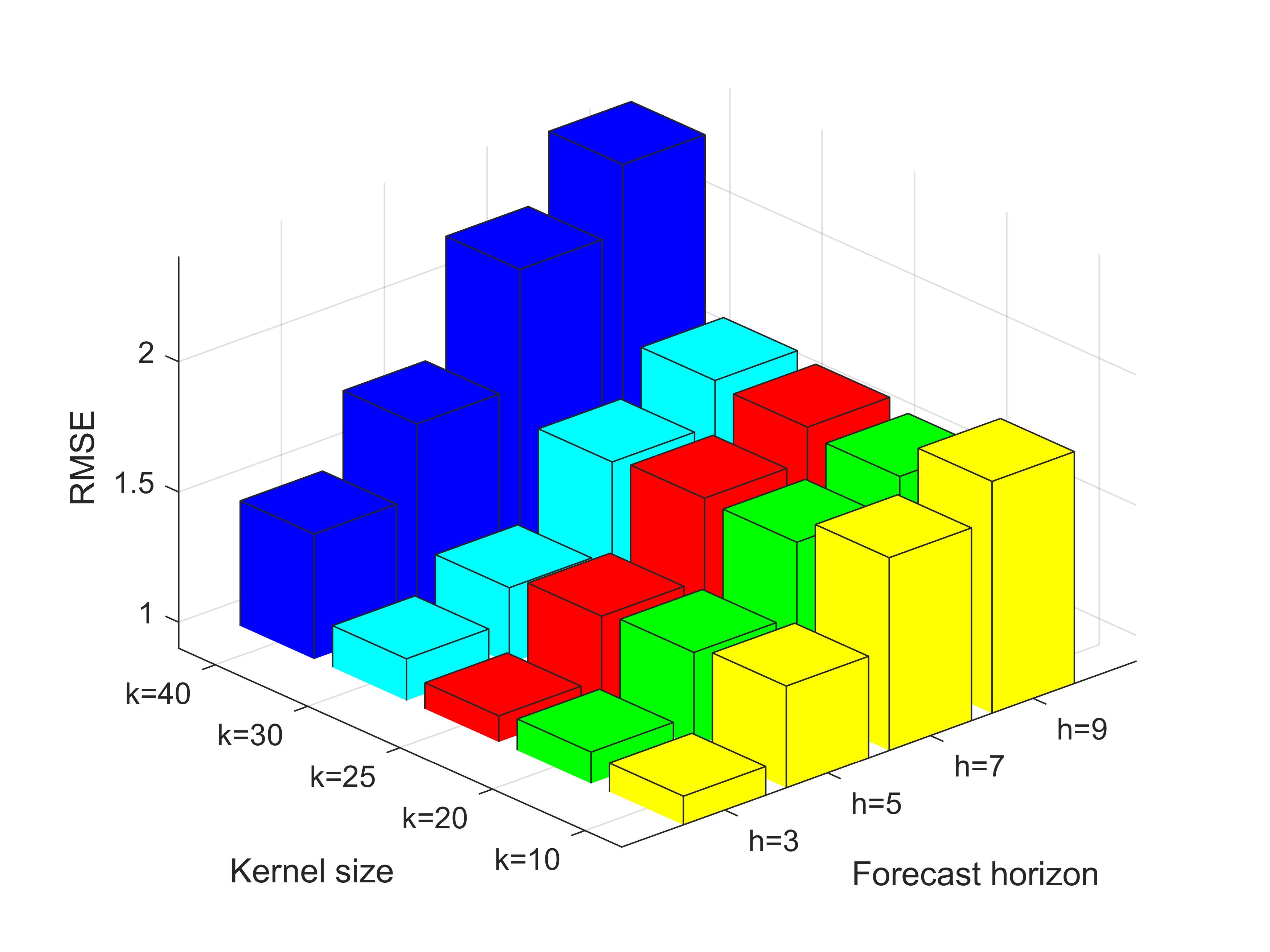}
\vspace{-0.2cm}
\caption{Histogram of the predicted RMSE results caused by various kernel size selections under the FFT transform matrix on the Pacific dataset.}\label{fig:Kernel size}
\vspace{-0.2cm}
\end{figure}

The selection of convolution kernel size is a crucial issue for TCTNN models, as different kernel sizes lead to varying prediction results. After conducting tests on multiple types of data, we find that setting the kernel size $k$ to half the time  dimension scale $t$, i.e., $k=t/2$, consistently produces favorable prediction results, which aligns with findings from previous study \cite{liu2022recovery}. As illustrated in Figure \ref{fig:Kernel size}, the test results on the Pacific dataset  also demonstrate the advantage of using $k=t/2$.

\subsubsection{Abalation study}
To further highlight the benefits of the proposed TCTNN method in ensuring the tensor structure, we convolve  time series along the temporal dimension to obtain the convolution matrix, upon which the nuclear norm is imposed
to get the following  \textit{Temporal Convolution Matrix Nulclear Norm}    (TCMNN) model:
$$
\min_{\tensor{X}\in \mathbb{R}^{t \times n_1 \times\cdots\times n_p}} ~\norm{\mathcal{V}_k(\X)}_{*},\ \text{s.t.} \ \Pomega(\X)= \Pomega(\M).
$$
where $\|\cdot\|_*$ denotes the nuclear norm of a matrix,  
$Y=reshape(\X,[n_1...n_p,t])=[\mathbf{y_1}',\cdots,\mathbf{y_{n_1...n_p}}']'$, $\mathcal{V}_k(\X)=[\mathcal{A}_k(\mathbf{y_1})',\cdots,\mathcal{A}_k(\mathbf{y_{n_1...n_p}})']'$, and $\mathcal{A}_k(\cdot)$ is a vector convolution transform that obeys the following form
$$
\mathcal{A}_k(\mathbf{a})=\left[\begin{array}{cccc}
a_1 & a_t & \cdots & a_{t-k+2} \\
a_2 & a_1 & \cdots & a_{t-k+3} \\
\vdots & \vdots & \vdots & \vdots \\
a_t & a_{t-1} & \dot & a_{t-k+1}
\end{array}\right], \mathbf{a}=[a_1,\cdots, a_t]'.
$$
Obviously, the TCMNN model is a special case of CNNM with a convolution kernel size of $k\times 1 \times\cdots\times 1$. For both the TCTNN model and the TCMNN model, we use $k=t/2$ to test on the  Pacific dataset. We select the forecast horizon $h=3, 5, 7, 9$ and summarize the prediction error MAE and RMSE of TCTNN and TCMNN in Table \ref{tab:TCMNN}. Table \ref{tab:TCMNN} shows that the prediction accuracy of TCTNN model is much higher than that of TCMNN, which reflects the importance of preserving the tensor structure of multi-dimensional time series in the prediction task. For other discussions of the experiment, such as multi-sample time series analysis and applications to multivariate time series, please refer to  supplementary material.
\begin{table}[!htbp]
	\centering
\renewcommand{\arraystretch}{1}
\setlength{\tabcolsep}{3pt}
\fontsize{9pt}{10pt}\selectfont
	\caption{Performance comparison (in MAE/RMSE) ) of TCTNN and TCMNN for  time series prediction on the Pacific dataset. }\label{tab:TCMNN}
\begin{tabular}{l|ccccccc}
      \hline
      Method  & h=3 & h=5  &h=7 &h=9 \\    
     \hline
     TCMNN & 1.77/2.18 & 2.32/2.75& 2.79/3.29& 2.93/3.49 \\
     TCTNN  & \textbf{0.75}/\textbf{1.01} & \textbf{0.94}/\textbf{1.25} & \textbf{1.20}/\textbf{1.56} & \textbf{1.29}/\textbf{1.68} \\
     \hline
\end{tabular}
\end{table}

\section{Conclusion and future directions } \label{sec:conclusion}
This study introduces an efficient method for multidimensional time series prediction by leveraging a novel deterministic tensor completion theory. Initially, we illustrate the limitations of applying tensor nuclear norm minimization directly to the prediction problem within the deterministic tensor completion framework. To address this challenge, we propose structural modifications to the original multidimensional time series and employ tensor nuclear norm minimization on the temporal convolution tensor, resulting in the TCTNN model. This model achieves exceptional performance in terms of accuracy and computational efficiency across various real-world scenarios.

However, as discussed, the assumption of low-rankness in the temporal convolution tensor may not hold when the multidimensional time series lack periodicity or smoothness along the time dimension. This observation suggests the need for further investigation into learning-based low-rank temporal convolution methods, similar to learning-based convolution norm minimization approaches \cite{liu2022time}. Additionally, the deterministic tensor completion theory developed in this study has broader applicability to structured missing data completion problems, presenting exciting opportunities for extending its use to other structured data recovery tasks.

\ifCLASSOPTIONcaptionsoff
  \newpage
\fi



\bibliographystyle{IEEEtranN}
\bibliography{ref}





\end{document}